\documentclass[journal]{IEEEtran}

\usepackage{graphicx}
\usepackage{comment}
\usepackage{amsmath,amssymb}
\usepackage{color}
\usepackage{subcaption}
\usepackage{booktabs}
\usepackage{multirow}
\usepackage{array}
\usepackage{cite}

\definecolor{ground}{RGB}{127, 201, 127}
\definecolor{pred}{RGB}{190, 174, 212}

\usepackage{fixltx2e}

\usepackage{stfloats}

\ifCLASSOPTIONcaptionsoff
  \usepackage[nomarkers]{endfloat}
 \let\MYoriglatexcaption\caption
 \renewcommand{\caption}[2][\relax]{\MYoriglatexcaption[#2]{#2}}
\fi

\begin{document}

\title{Rethinking Semantic Segmentation Evaluation for Explainability and Model Selection}

\author{Yuxiang~Zhang,
        Sachin~Mehta,
        and~Anat~Caspi 
\thanks{This work has been submitted to the IEEE for possible publication. Copyright may be transferred without notice, after which this version may no longer be accessible}
\thanks{Y. Zhang, S. Mehta, and A. Caspi are with the University of Washington, Seattle,
 WA, 98115 USA. E-mail: \{yz325, sacmehta, caspian\}@cs.washington.edu.}
}

\maketitle

\begin{abstract}
Semantic segmentation aims to robustly predict coherent class labels for entire regions of an image. It is a scene understanding task that powers real-world applications (e.g., autonomous navigation). One important application, the use of imagery for \textit{automated semantic understanding of pedestrian environments}, provides remote mapping of accessibility features in street environments. This application (and others like it) require detailed geometric information of geographical objects.  Semantic segmentation is a prerequisite for this task since it maps contiguous regions of the same class as single entities. Importantly, semantic segmentation uses like ours are not pixel-wise outcomes; however, most of their \textit{quantitative} evaluation metrics (e.g., mean Intersection Over Union) are based on pixel-wise similarities to a ground-truth, which fails to emphasize over- and under-segmentation properties of a segmentation model. Here, we introduce a \textit{new metric to assess region-based over- and under-segmentation. }We analyze and compare it to other metrics, demonstrating that the use of our metric lends greater explainability to semantic segmentation model performance in real-world applications.
\end{abstract}


\section{Introduction}
\label{sec:introduction}

\IEEEPARstart{S}{emantic} segmentation is an important computer vision task that clusters together parts of images belonging to the same object class. It is used in many real-world visual scene understanding applications (e.g., autonomous driving, robotics, medical diagnostics). Substantial improvements in semantic segmentation performance were driven by recent advances in machine learning (ML), especially convolutional neural networks (CNNs) \cite{lecun1995convolutional,krizhevsky2012imagenet}, coupled with advancements in hardware technology
\cite{long2015fully,chen2017deeplab,zhao2017pyramid,chen2017rethinking}. Moreover, these developments have made some time-sensitive domains (e.g., autonomous driving \cite{bojarski2016end}) practically feasible. 

\vspace{1mm}
\noindent \textbf{Motivation.} In the area of autonomous driving, extensive research and mapping efforts have produced detailed maps of automobile roads. However, street-side environments that serve pedestrians have not received merited attention. Real-time, efficient inference about street-side infrastructure (i.e., mapping the pedestrian environment) is essential for path planning in delivery robots and autonomous wheelchairs. Importantly, similarly detailed maps of path networks can provide data feeds to mobility apps, closing informational gaps experienced by pedestrians with disabilities that answer \textit{where} paths are, \textit{how} paths are connected, and \textit{whether paths have the amenities and landmarks} necessary to traverse them efficiently and safely.  Providing street-side maps will promote sustainable, equal access for people of all abilities to navigate urban spaces. A recent study of semantic segmentation in imagery data of pedestrian environments, conducted to enable real-time mapping of accessibility features in street environments ~\cite{zhang2019stereo}, identifies challenges of deploying CNNs in real-time, compute-sparse environments. In this application space, an \textit{ideal model} should predict connected regions of pixels that represent: (1) pedestrian pathways, e.g., sidewalks, footways, and road crossings in a scene, and (2) common object classes in the pedestrian environments, e.g., poles, fire hydrants, and benches. For these applications, removing under- or over-segmentation is often more important than achieving high pixel-wise accuracy. Questions like, "How many paths are there in the current scene, and how are they connected?" can be answered only through segmentation results with region-wise fidelity.
 
\vspace{1mm}
\noindent \textbf{Model evaluation for explainability.} Semantic segmentation is performed through pixel-wise classification; however, as an important subtask of image understanding, it contributes to the discovery of groups of objects and identification of semantically meaningful distributions and patterns in input image data. In ~\cite{shi2000normalized}, segmentation is framed as a graph partitioning problem, and normalized cut criteria is proposed for measuring the total similarity and dissimilarity of partitions. In ~\cite{achanta2010slic}, superpixel (effectively regions consisting of multiple pixels) is used to advocate for evaluating segmentation in practical applications. Semantic segmentation decomposes a surrounding scene into meaningful semantic regions that are useful for decision-making in downstream applications. For instance, when mapping in the pedestrian environment, it is important to identify which connected regions are correctly mapped to real pathways. In sum, uses of semantic segmentation are not pixel- but \textit{region-based}, as evidenced in their application to aggregate semantically rich information about the surroundings used to assess navigation options for people with visual disabilities (e.g., Mehta et al. \cite{mehta2017identifying} and Yang et al. \cite{yang2018unifying}). 

\begin{figure*}[t!]
    \centering
    \begin{subfigure}[b]{1.85\columnwidth}
        \centering
        \begin{subfigure}[b]{0.23\columnwidth}
            \centering
            \includegraphics[width=\columnwidth]{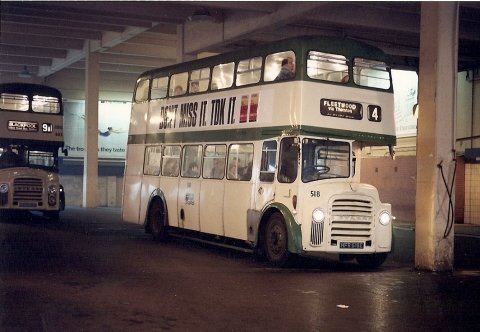}
        \end{subfigure}
        \hfill
        \begin{subfigure}[b]{0.23\columnwidth}
            \centering
            \includegraphics[width=\columnwidth]{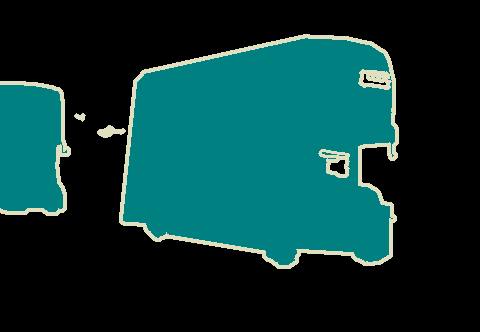}
        \end{subfigure}
        \hfill
        \begin{subfigure}[b]{0.23\columnwidth}
            \centering
            \includegraphics[width=\columnwidth]{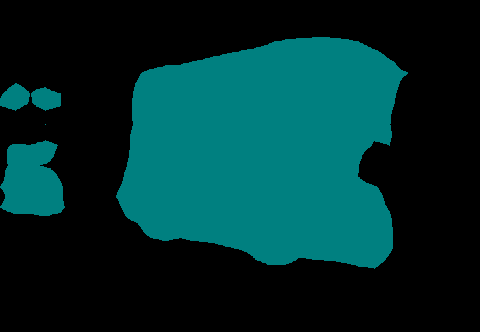}
        \end{subfigure}
        \hfill
        \begin{subfigure}[b]{0.25\columnwidth}
            \centering
            \includegraphics[width=\columnwidth]{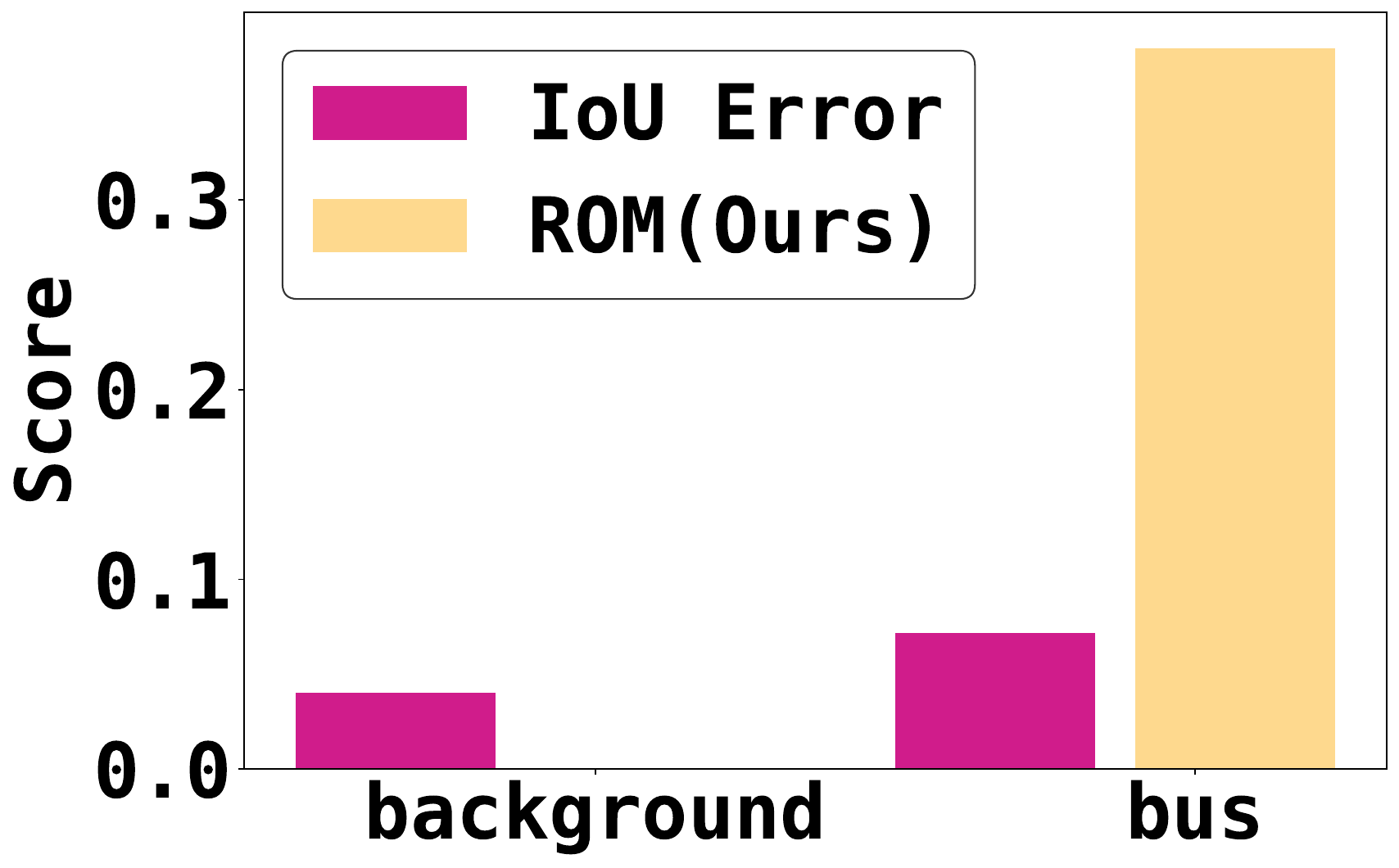}
        \end{subfigure}
        \caption{Over-segmentation}
        \label{fig:oversegexample}
    \end{subfigure}
    \vfill
    \begin{subfigure}[b]{1.85\columnwidth}
        \centering
        \begin{subfigure}[b]{0.23\columnwidth}
            \centering
            \includegraphics[width=\columnwidth, height=75px]{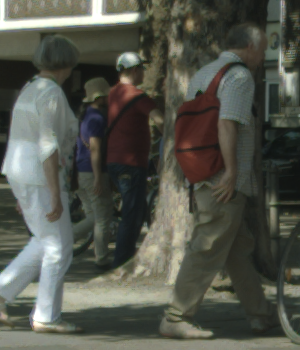}
        \end{subfigure}
        \hfill
        \begin{subfigure}[b]{0.23\columnwidth}
            \centering
            \includegraphics[width=\columnwidth, height=75px]{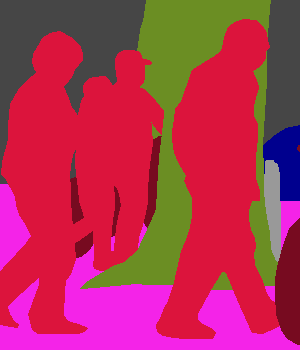}
        \end{subfigure}
        \hfill
        \begin{subfigure}[b]{0.23\columnwidth}
            \centering
            \includegraphics[width=\columnwidth, height=75px]{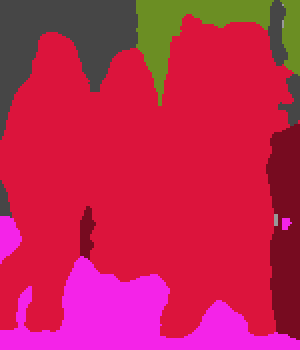}
        \end{subfigure}
        \hfill
        \begin{subfigure}[b]{0.25\columnwidth}
            \centering
            \includegraphics[width=\columnwidth]{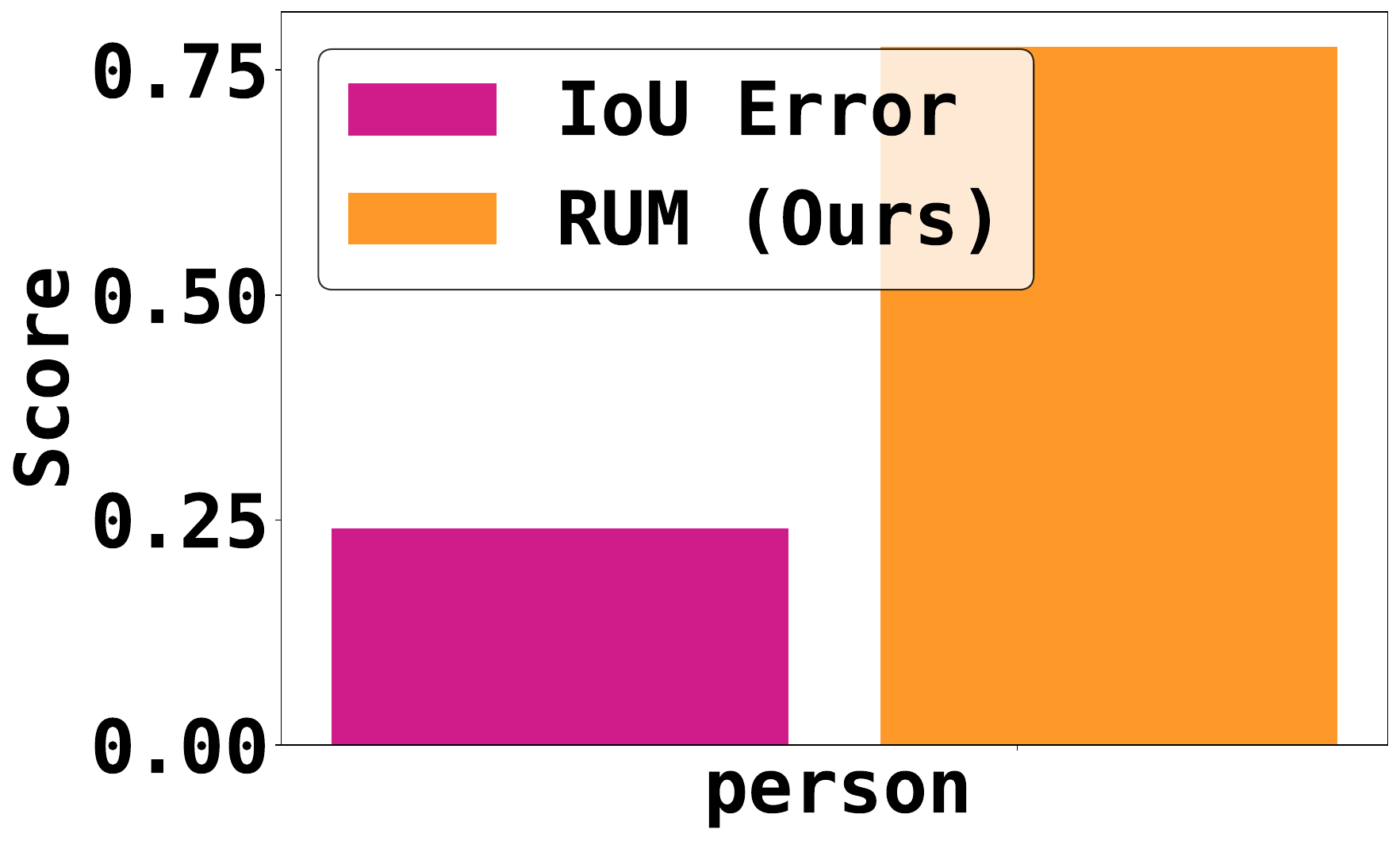}
        \end{subfigure}
        \caption{Under-segmentation}
        \label{fig:undersegexample}
    \end{subfigure}
    \caption{
    Examples that illustrate over- and under-segmentation issues in semantic segmentation.(\textbf{Left} to \textbf{right}: RGB image, ground-truth, prediction, and error metrics)   In (a), the back of the bus is over-segmented (one region is segmented into three). In (b), people are under-segmented (three groups are segmented as one). Though the IOU error (1 - IOU score) is low in both cases, the IOU metric does not reflect these distinctions. However, these issues can be explained using the proposed metrics for over- (ROM) and under-segmentation (RUM).
    }
    \label{fig:overunderseg}
\end{figure*}

Most quantitative evaluation metrics (e.g., mean Intersection Over Union, mIOU) currently used to evaluate performance or error in semantic segmentation algorithms are based on pixel-wise similarity between ground-truth and predicted segmentation masks. Though these metrics effectively capture certain properties about clustering or classification performance, they fail to provide a proper meter of region-based agreement with a given ground-truth or provide an expectation about broken regions that model segmentation may yield. We assert that quantitative measures of \textit{region segmentation similarity} (rather than pixel-wise or boundary similarity) will prove useful to perform model selection and to measure the consistency of predicted regions with a ground-truth in a manner that is invariant to region refinement or boundary ambiguity. Such measures will prove practically useful in applications of semantic segmentation where performance is affected by over- and under-segmentation. For clarity, a predicted region is said to contribute to \textit{over-segmentation }when the prediction overlaps a ground-truth region that another predicted region also overlaps. A predicted region is \textit{under-segmented} when the prediction overlaps two or more ground-truth regions. Examples of over- and under-segmentation are shown in Figure \ref{fig:overunderseg}.

\vspace{1mm}
\noindent \textbf{Contributions.} This work introduces quantitative evaluation metrics for semantic segmentation that provide granular accounting for over- and under-segmentation. We express the empirical probability that a predicted segmentation mask is over- or under-segmenting and also   penalize model-segmentations that repeatedly over-segment the same region, causing large variations between model-prediction and ground-truth. We demonstrate how these issues affect currently used segmentation algorithms across a variety of segmentation datasets and models and that currently available metrics do not differentiate between models since they do not measure these issues directly. 

\section{Related Work}
\label{sec:related_work}
Most segmentation models deployed today are CNN-based models that adopt encoder-decoder architectures (e.g., \cite{long2015fully,badrinarayanan2017segnet,chen2017deeplab,zhao2017pyramid,ronneberger2015u}).
To enable real-world, low-latency deployments on resource-constrained devices, recent efforts in the field are proposing light-weight segmentation models (e.g., \cite{mehta2018espnet,romera2017erfnet,mehta2019espnetv2,sandler2018mobilenetv2}). Most of these models are trained with strong pixel-level supervision, using error metrics such as IOU (Intersection Over Union) error to assess model consistency with annotated image segments \cite{Shimoda2018}. However, in practice, these metrics are ill-suited for discriminating between effective and ineffective region-based segmentation consistency.

Here, we survey measures of similarity popular in
segmentation literature and discuss their relevance and associated disadvantages as performance metrics for semantic segmentation. In particular, we highlight two questions that help us characterize and differentiate metrics that are used to quantify segmentation performance: (1) Does the metric correlate with the level of agreement between model segmentation and ground-truth?, and (2) Is the metric robust to region refinements or small ambiguities in region boundaries that naturally arise in images?

\begin{figure*}[t!]
    \centering
    \begin{subfigure}[b]{0.24\columnwidth}
        \centering
        \caption{}\vspace{-7pt}
        \includegraphics[width=\columnwidth]{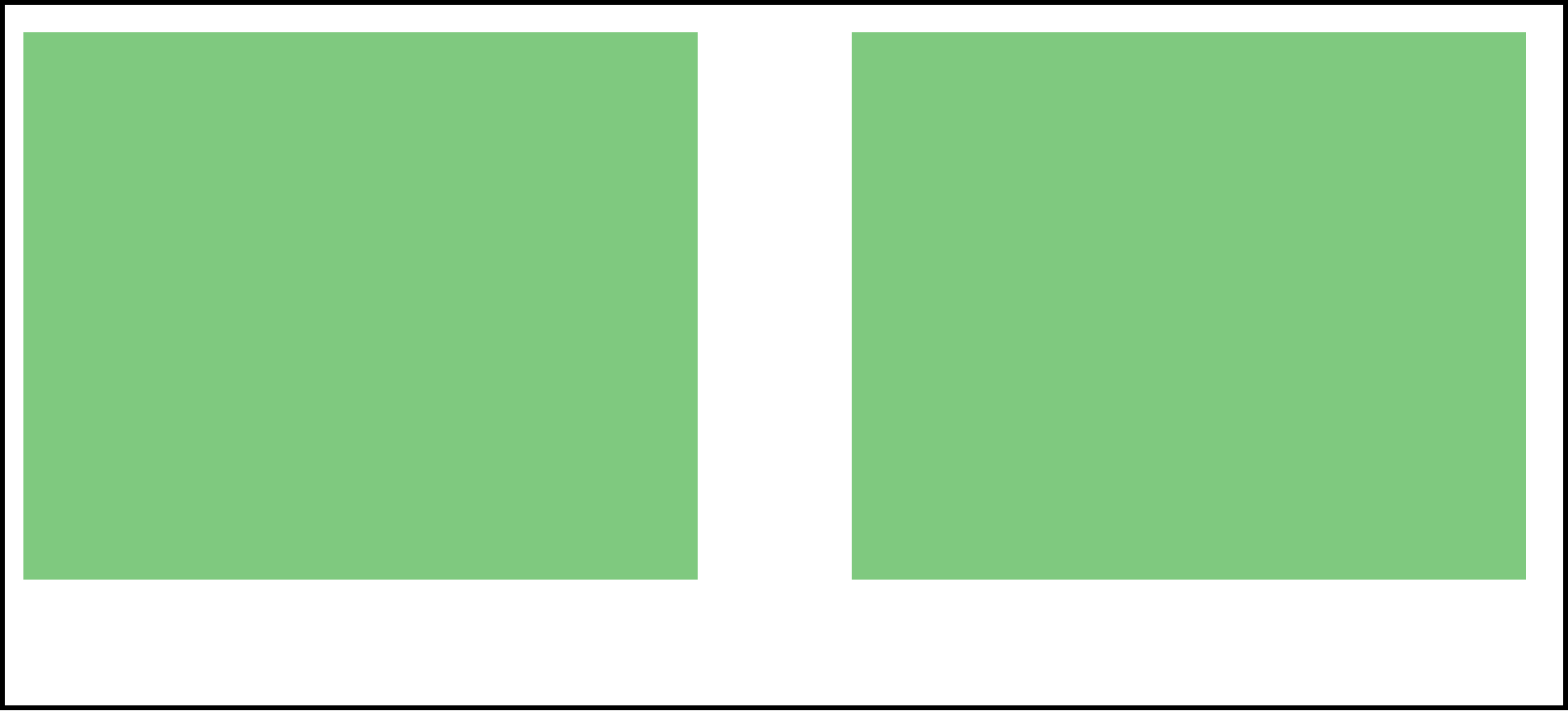}
        \label{fig:toy_a}
    \end{subfigure}
    \begin{subfigure}[b]{0.24\columnwidth}
        \centering
        \caption{}\vspace{-7pt}
        \includegraphics[width=\columnwidth]{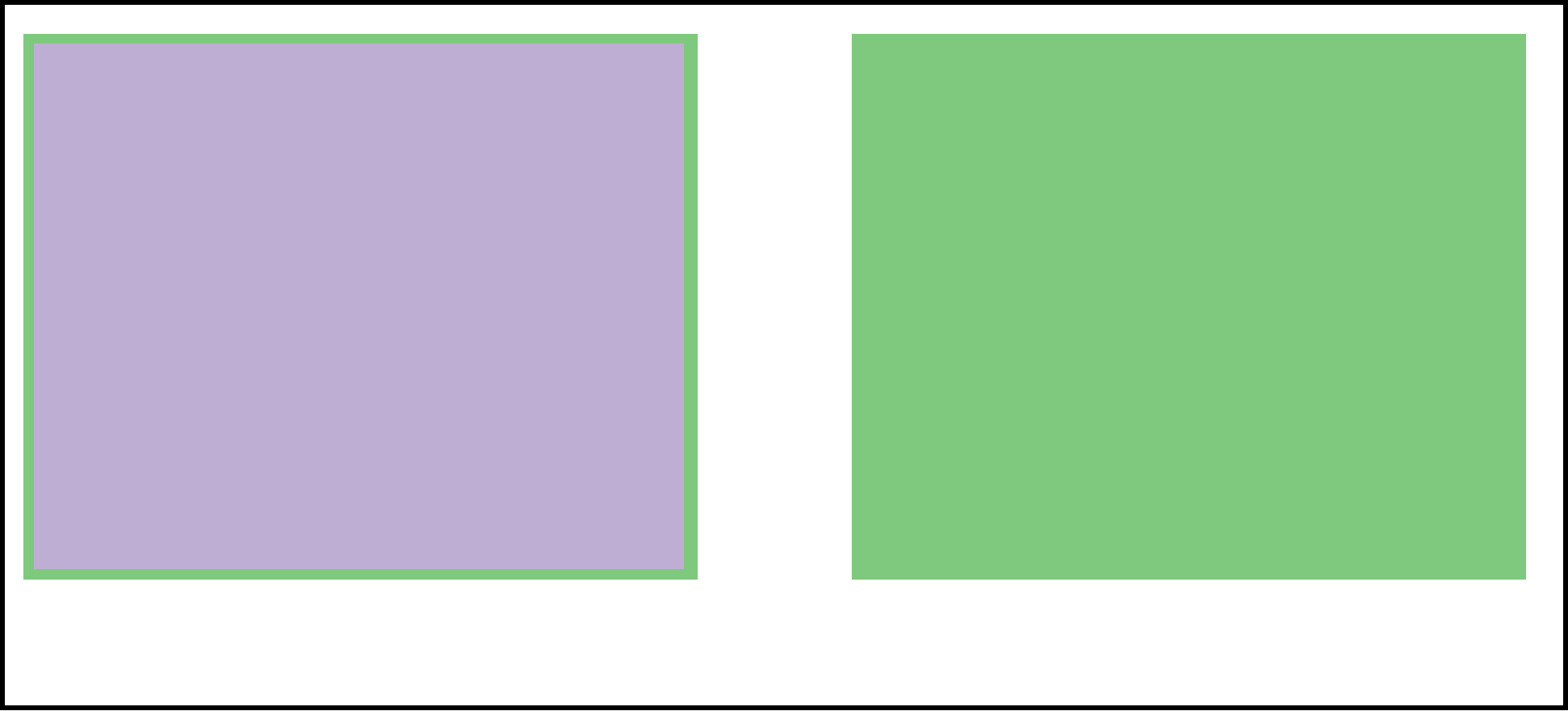}
        \label{fig:toy_b}
    \end{subfigure}
    \begin{subfigure}[b]{0.24\columnwidth}
        \centering
        \caption{}\vspace{-7pt}
        \includegraphics[width=\columnwidth]{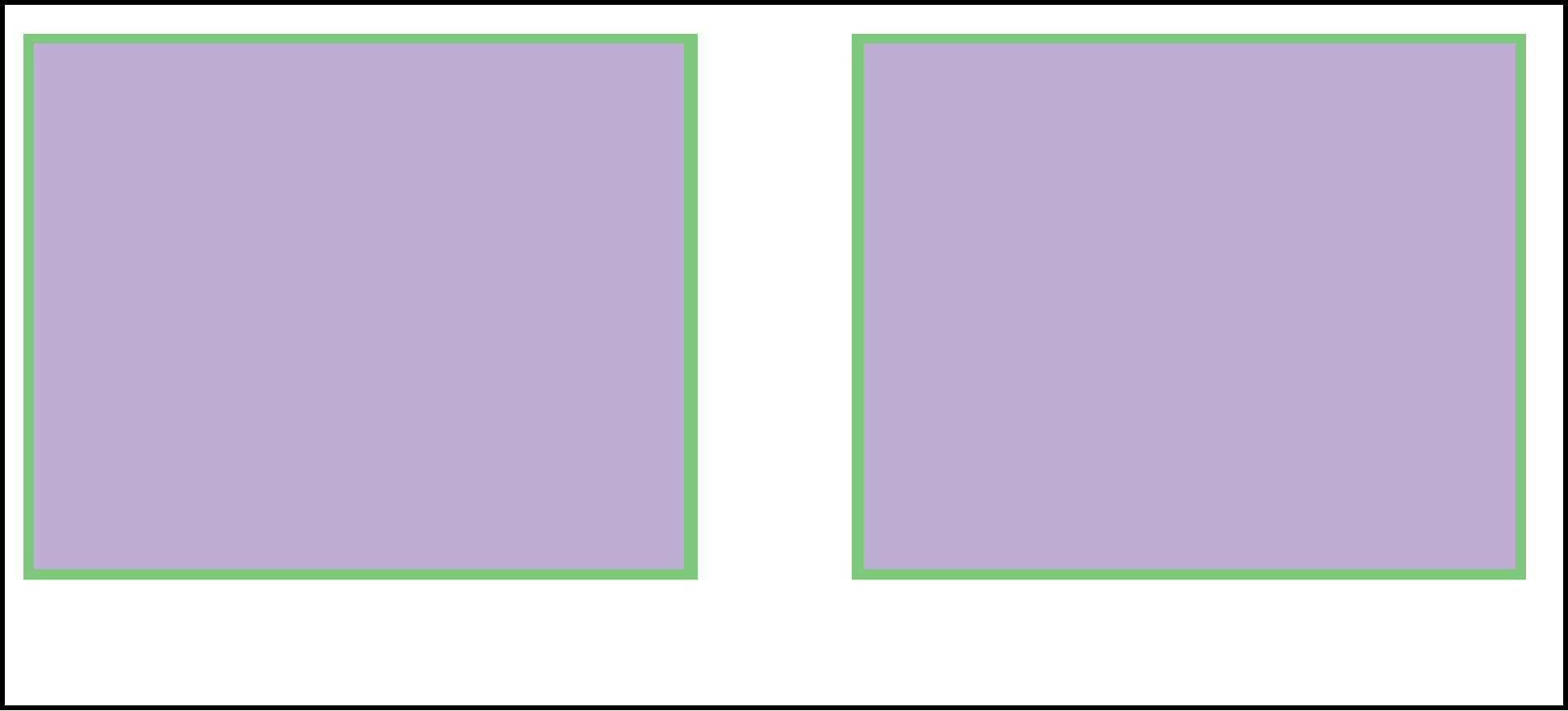}
        \label{fig:toy_c}
    \end{subfigure}
    \begin{subfigure}[b]{0.24\columnwidth}
        \centering
        \caption{}\vspace{-7pt}
        \includegraphics[width=\columnwidth]{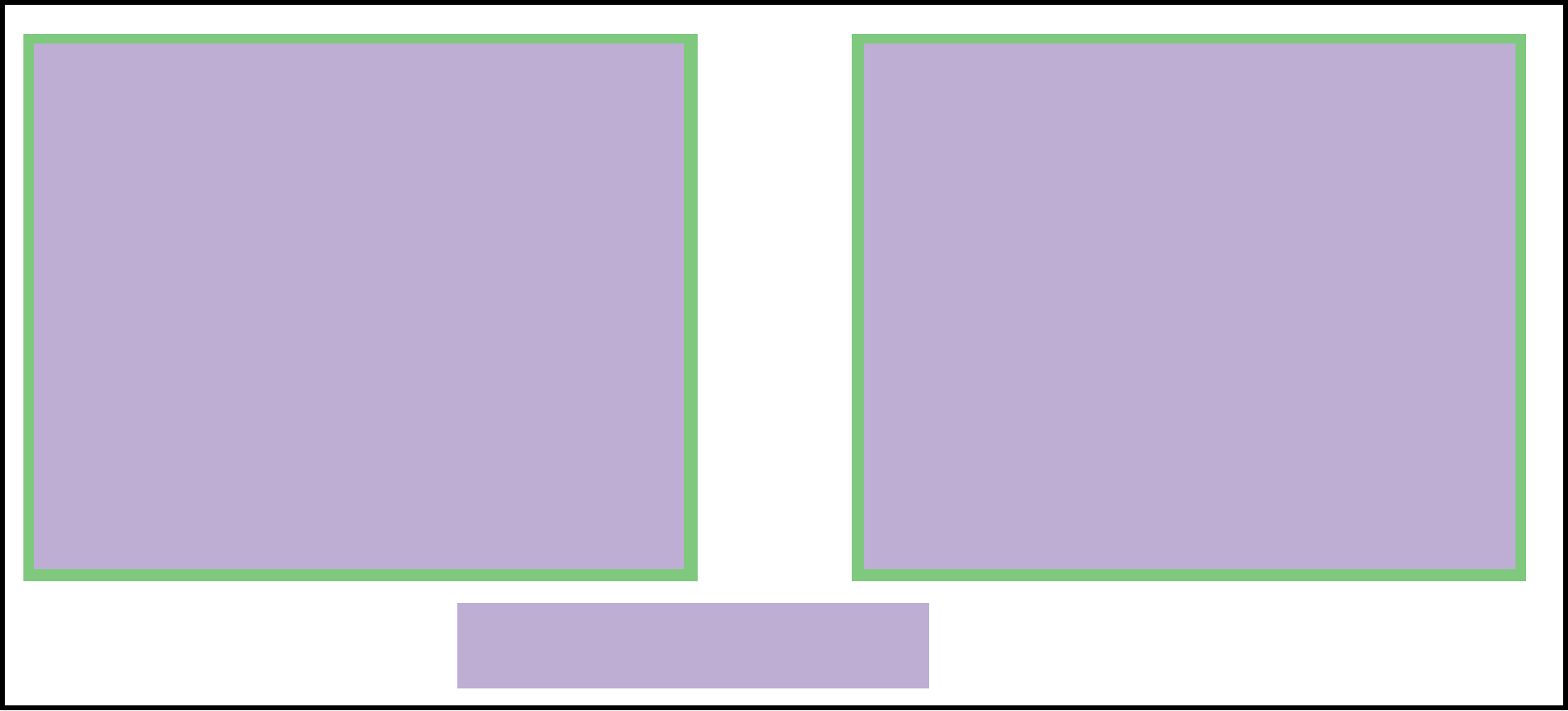}
        \label{fig:toy_d}
    \end{subfigure}
    \begin{subfigure}[b]{0.24\columnwidth}
        \centering
        \caption{}\vspace{-7pt}
        \includegraphics[width=\columnwidth]{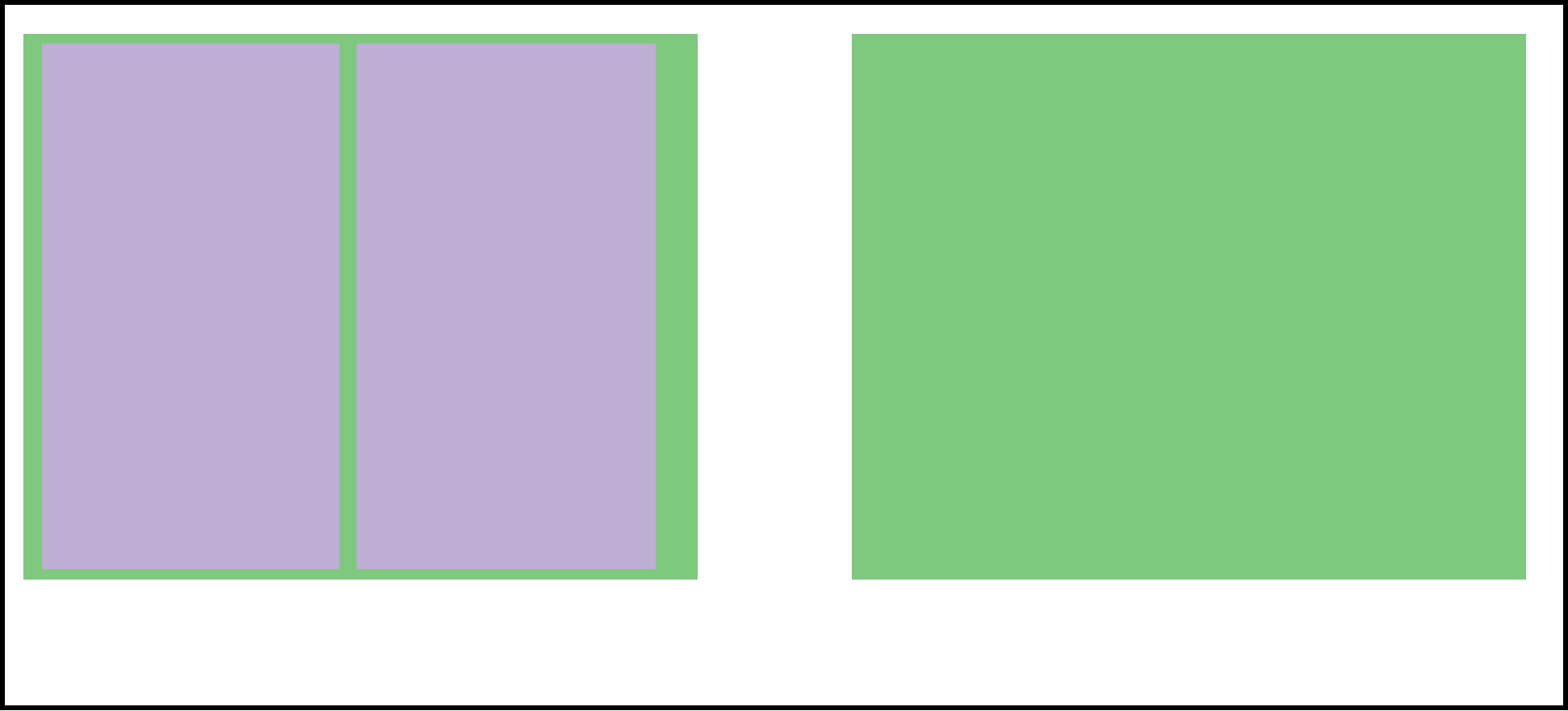}
        \label{fig:toy_e}
    \end{subfigure}
    \begin{subfigure}[b]{0.24\columnwidth}
        \centering
        \caption{}\vspace{-7pt}
        \includegraphics[width=\columnwidth]{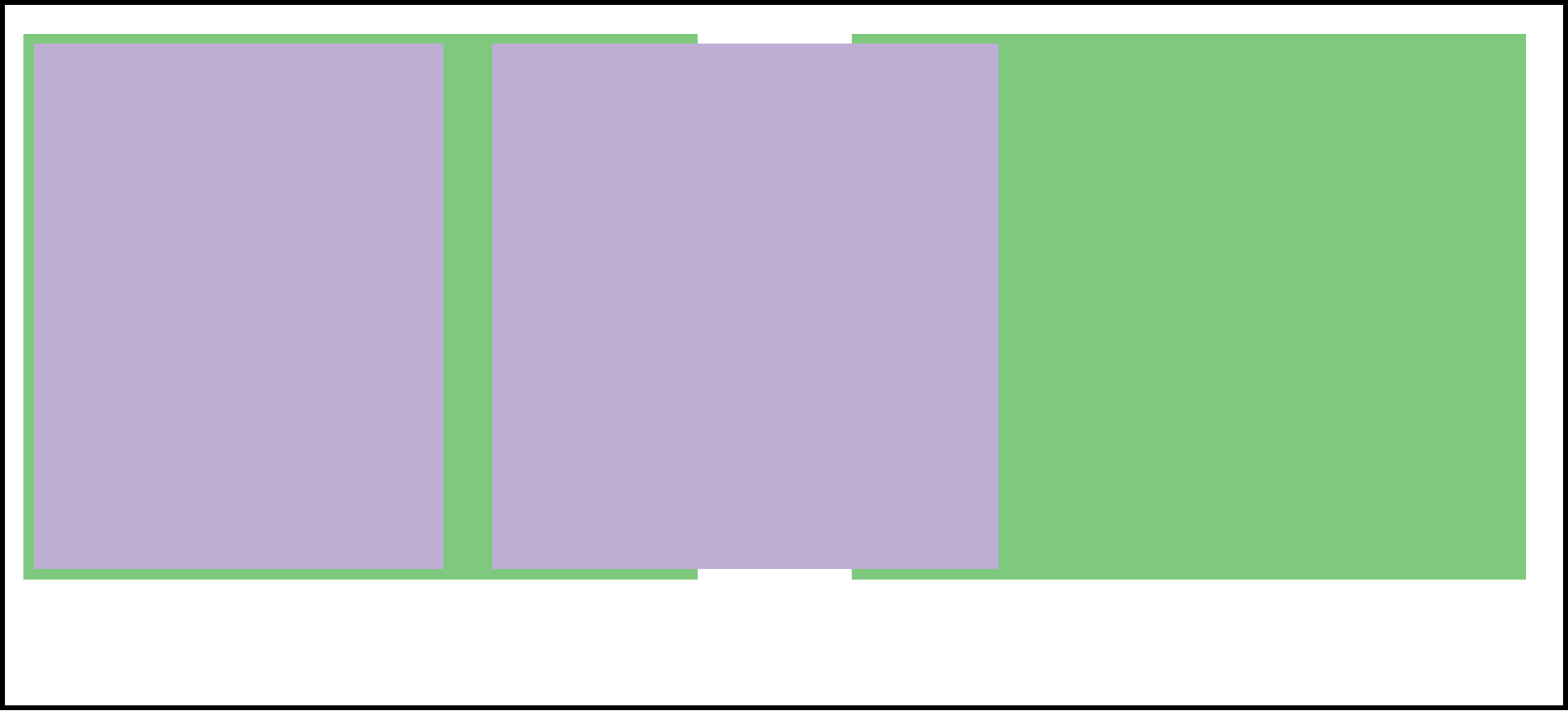}
        \label{fig:toy_f}
    \end{subfigure}
    \begin{subfigure}[b]{0.24\columnwidth}
        \centering
        \caption{}\vspace{-7pt}
        \includegraphics[width=\columnwidth]{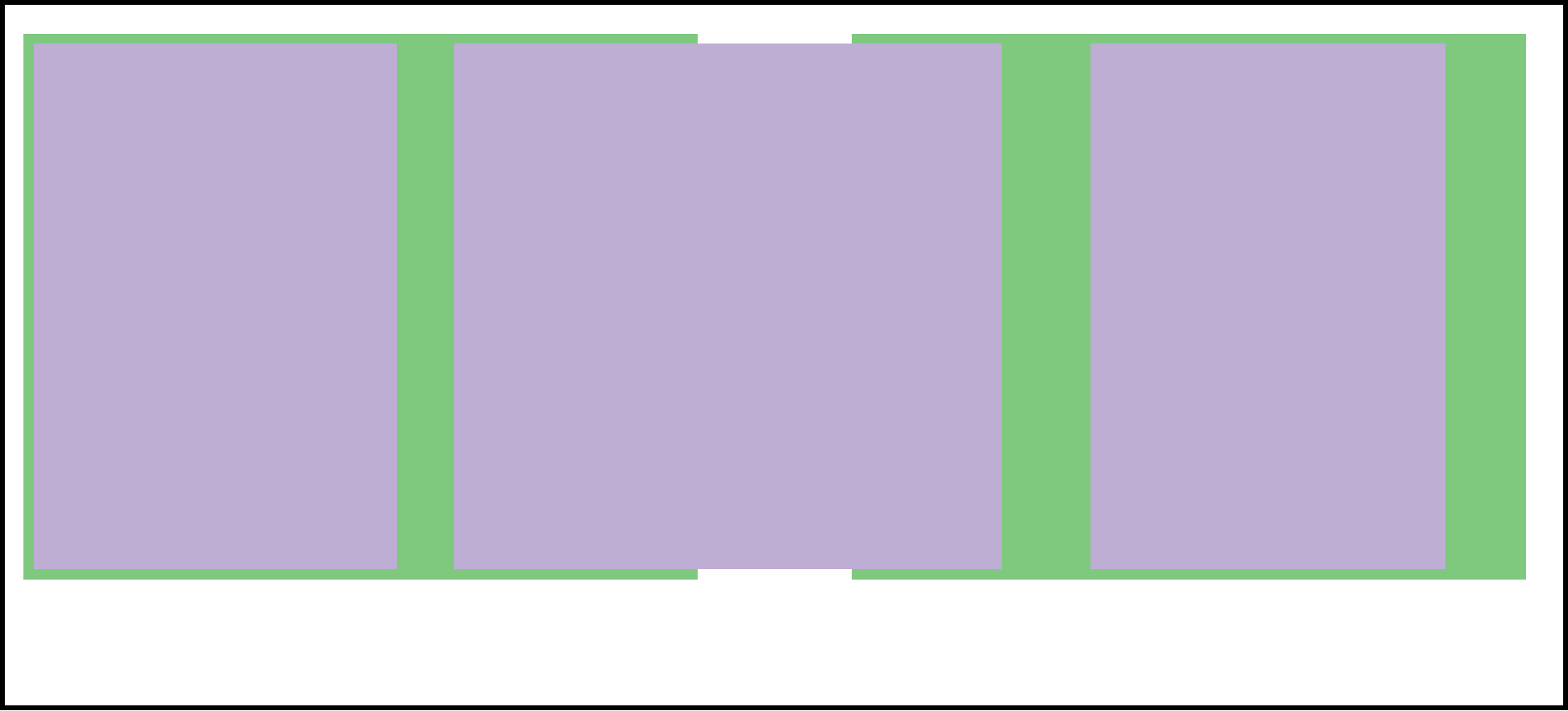}
        \label{fig:toy_g}
    \end{subfigure}
    \begin{subfigure}[b]{0.24\columnwidth}
        \centering
        \caption{}\vspace{-7pt}
        \includegraphics[width=\columnwidth]{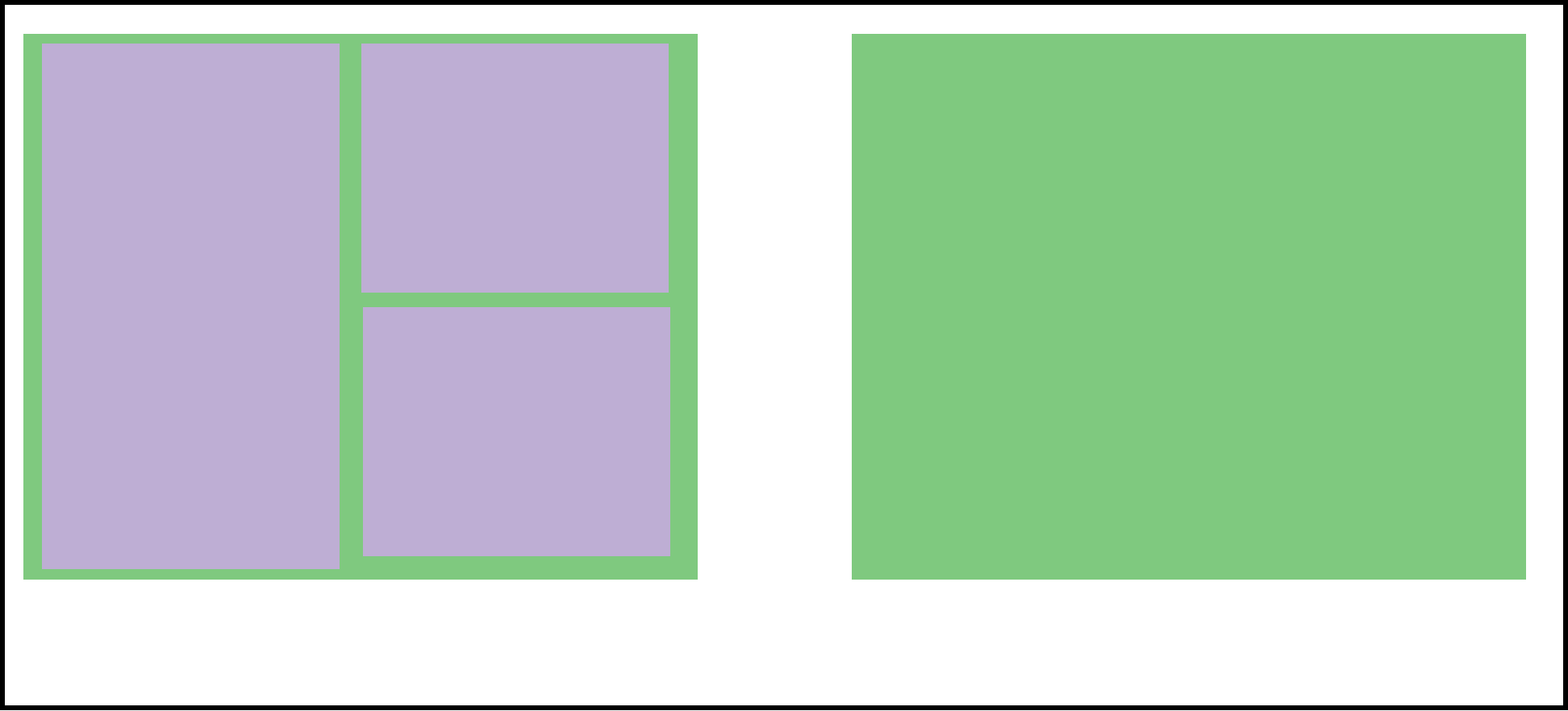}
        \label{fig:toy_h}
    \end{subfigure}
    \begin{subfigure}[b]{0.24\columnwidth}
        \centering
        \caption{}\vspace{-7pt}
        \includegraphics[width=\columnwidth]{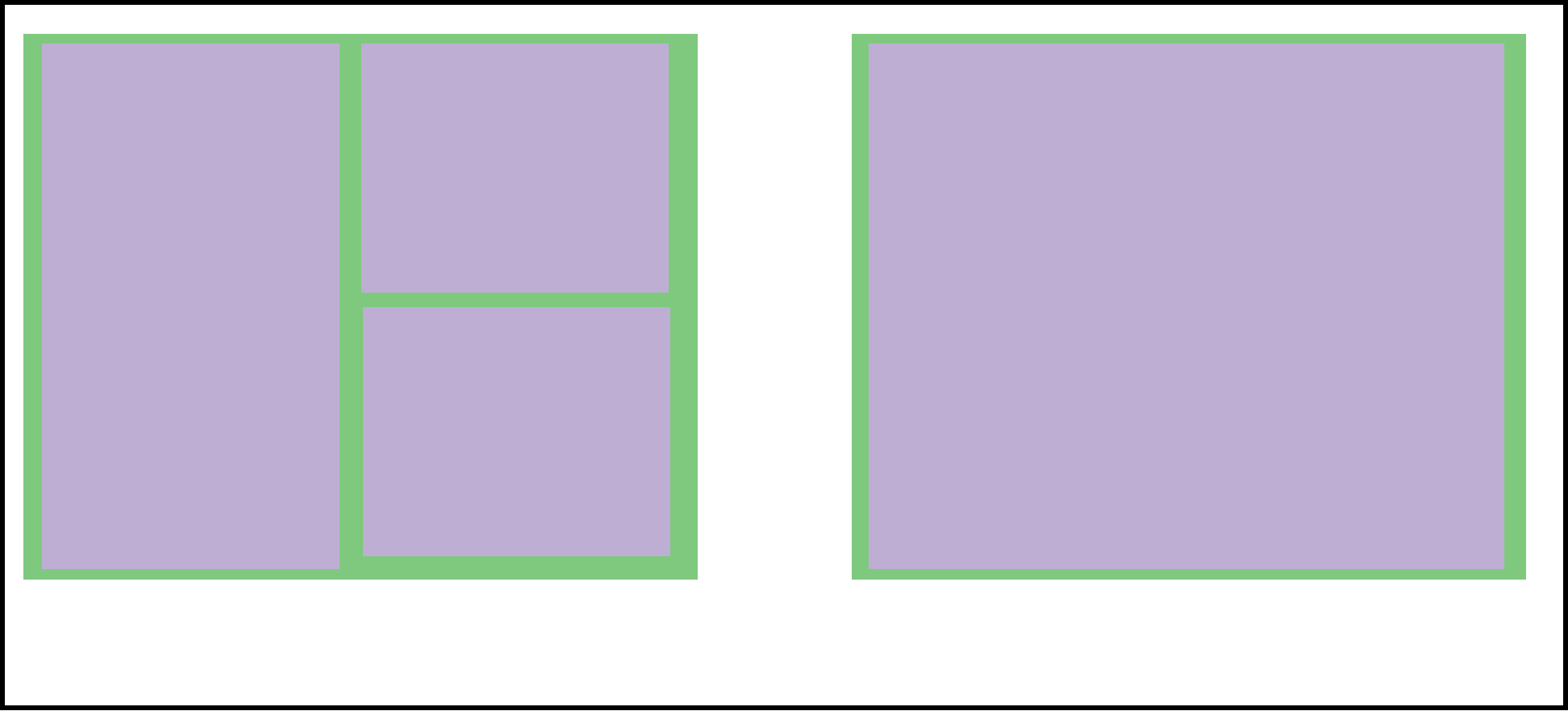}
        \label{fig:toy_i}
    \end{subfigure}
    \begin{subfigure}[b]{0.24\columnwidth}
        \centering
        \caption{}\vspace{-7pt}
        \includegraphics[width=\columnwidth]{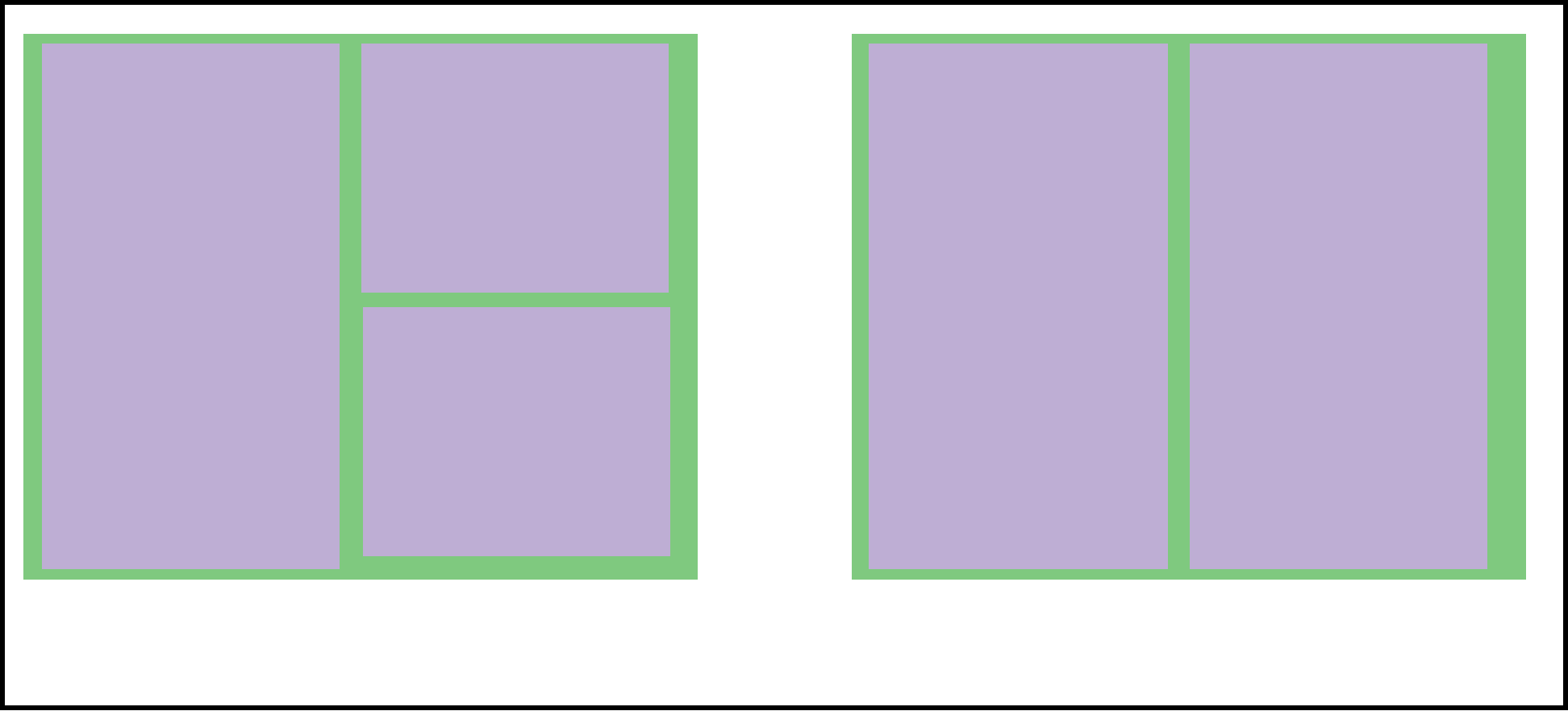}
        \label{fig:toy_j}
    \end{subfigure}
    \begin{subfigure}[b]{0.24\columnwidth}
        \centering
        \caption{}\vspace{-7pt}
        \includegraphics[width=\columnwidth]{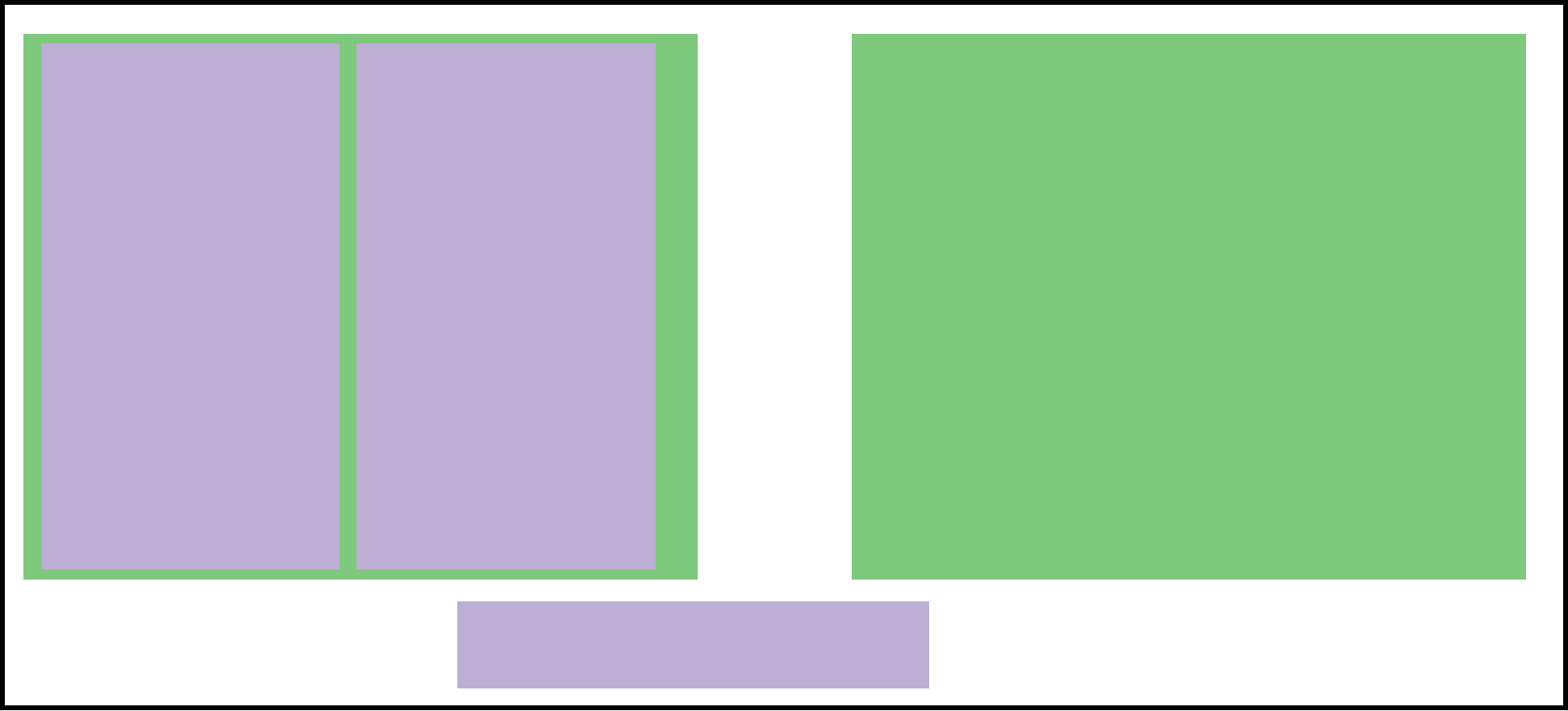}
        \label{fig:toy_k}
    \end{subfigure}
    \begin{subfigure}[b]{0.24\columnwidth}
        \centering
        \caption{}\vspace{-7pt}
        \includegraphics[width=\columnwidth]{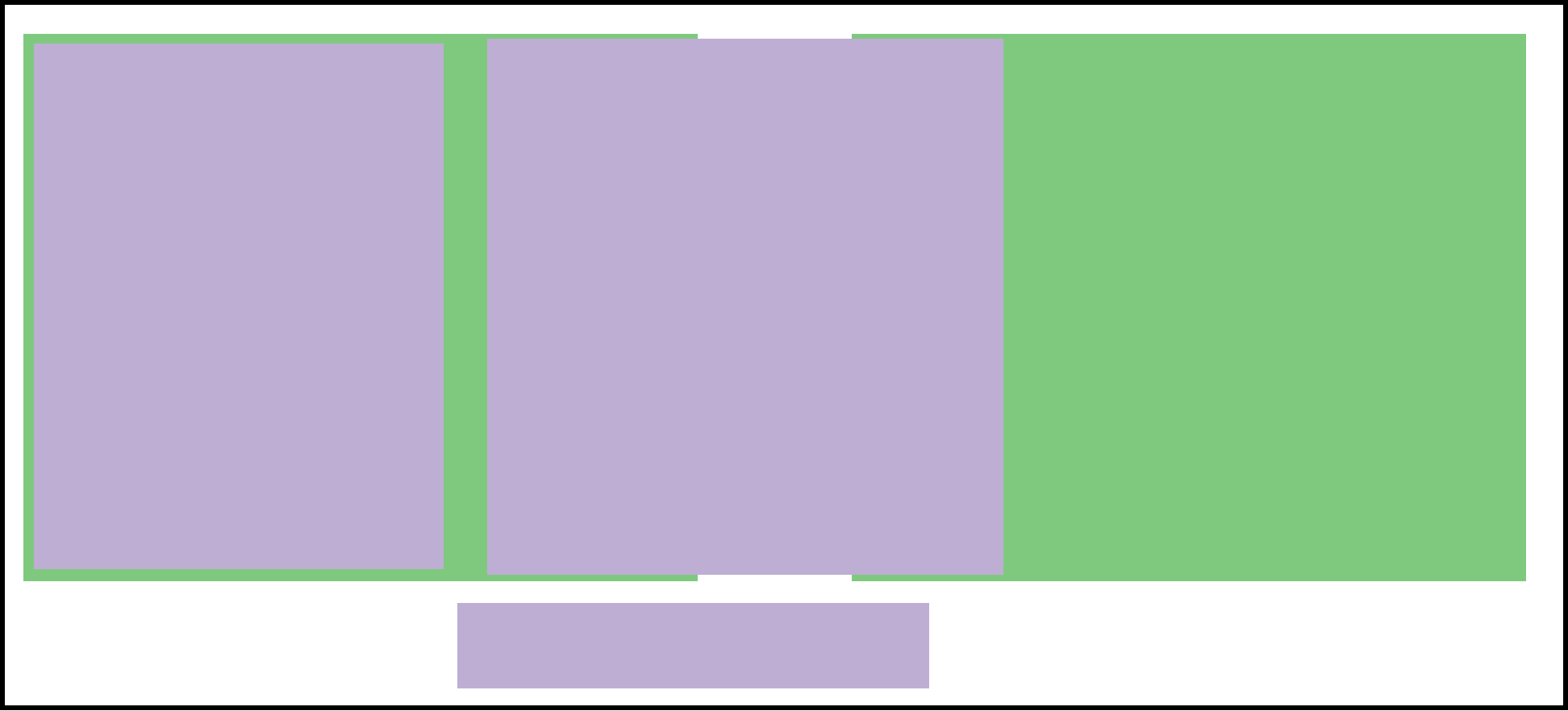}
        \label{fig:toy_l}
    \end{subfigure}
    \begin{subfigure}[b]{0.24\columnwidth}
        \centering
        \caption{}\vspace{-7pt}
        \includegraphics[width=\columnwidth]{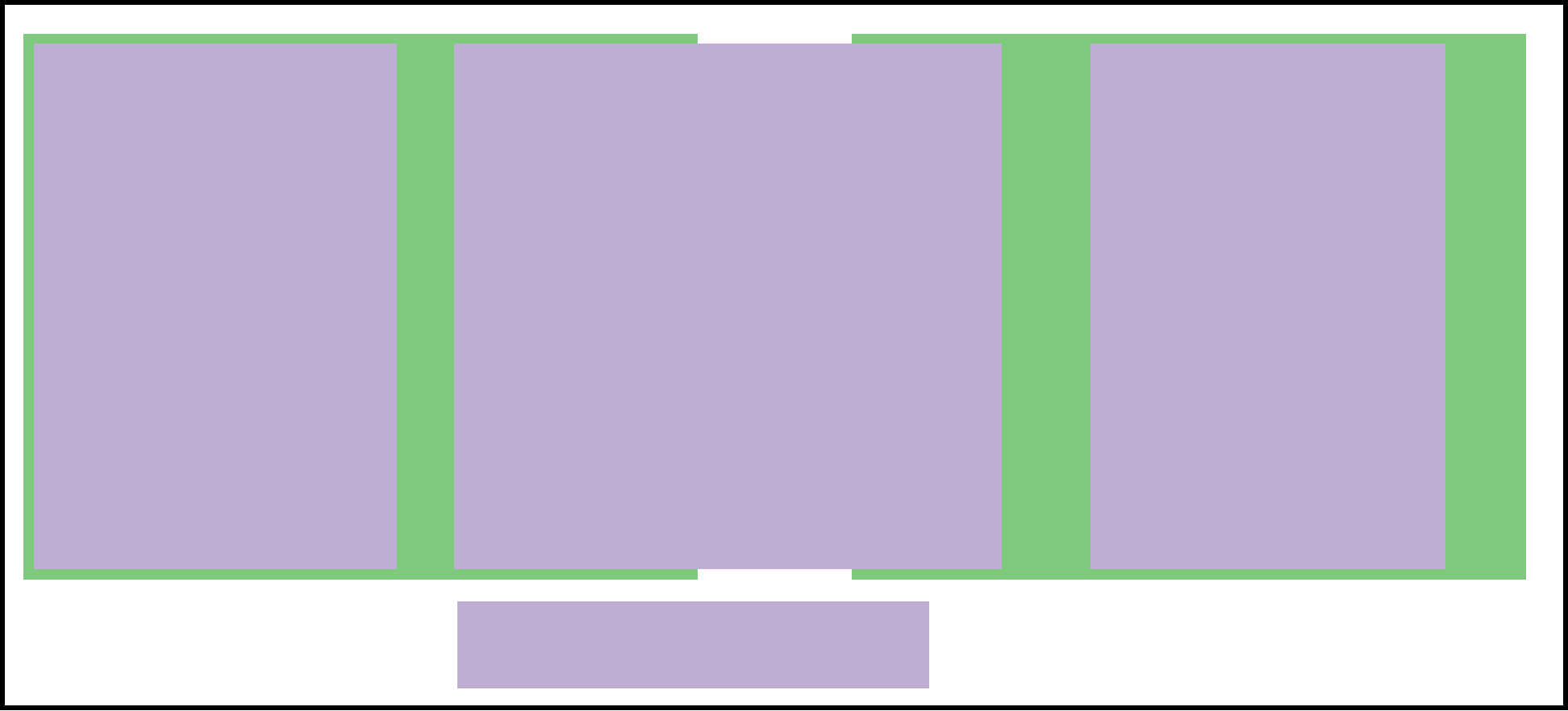}
        \label{fig:toy_m}
    \end{subfigure}
    \begin{subfigure}[b]{0.24\columnwidth}
        \centering
        \caption{ }\vspace{-7pt}
        \includegraphics[width=\columnwidth]{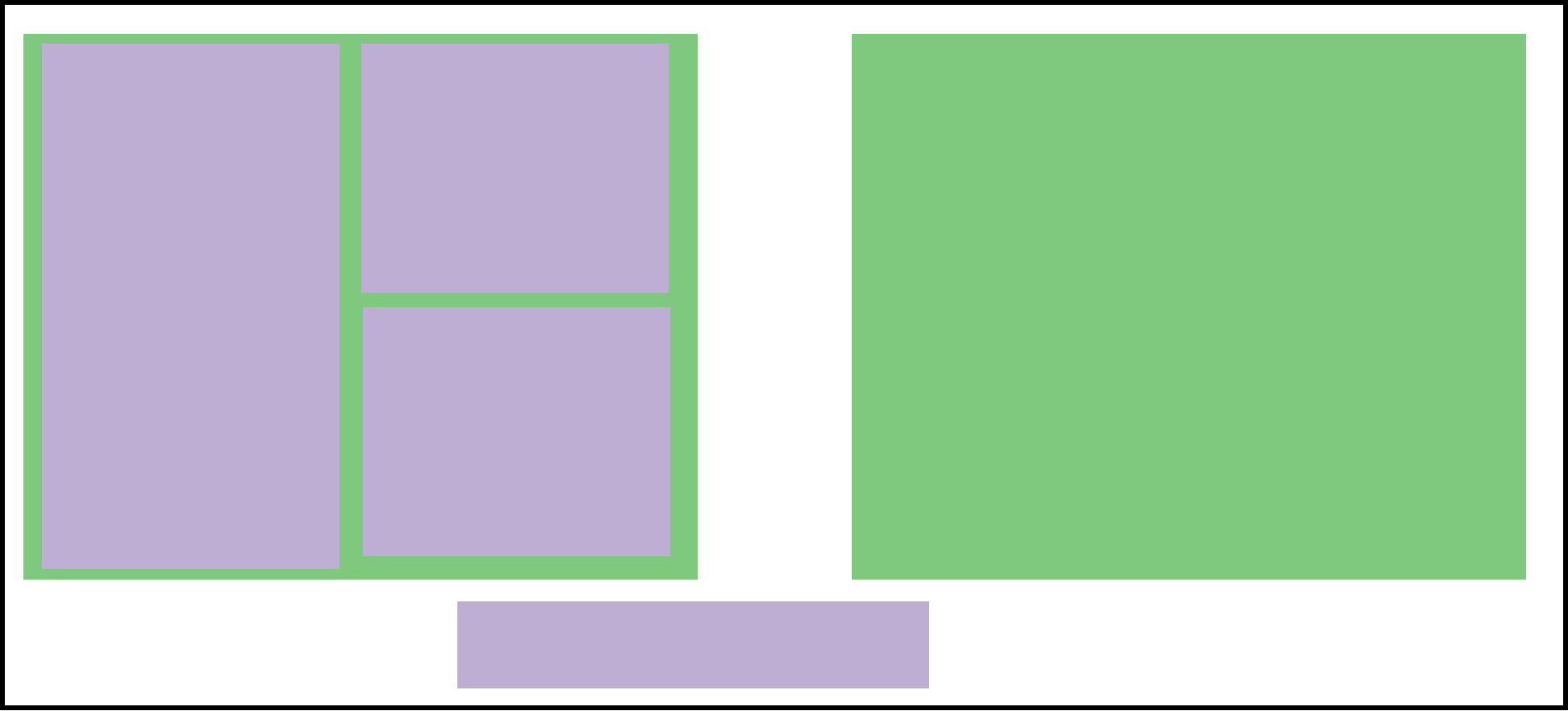}
        \label{fig:toy_n}
    \end{subfigure}
    \begin{subfigure}[b]{0.24\columnwidth}
        \centering
        \caption{ }\vspace{-7pt}
        \includegraphics[width=\columnwidth]{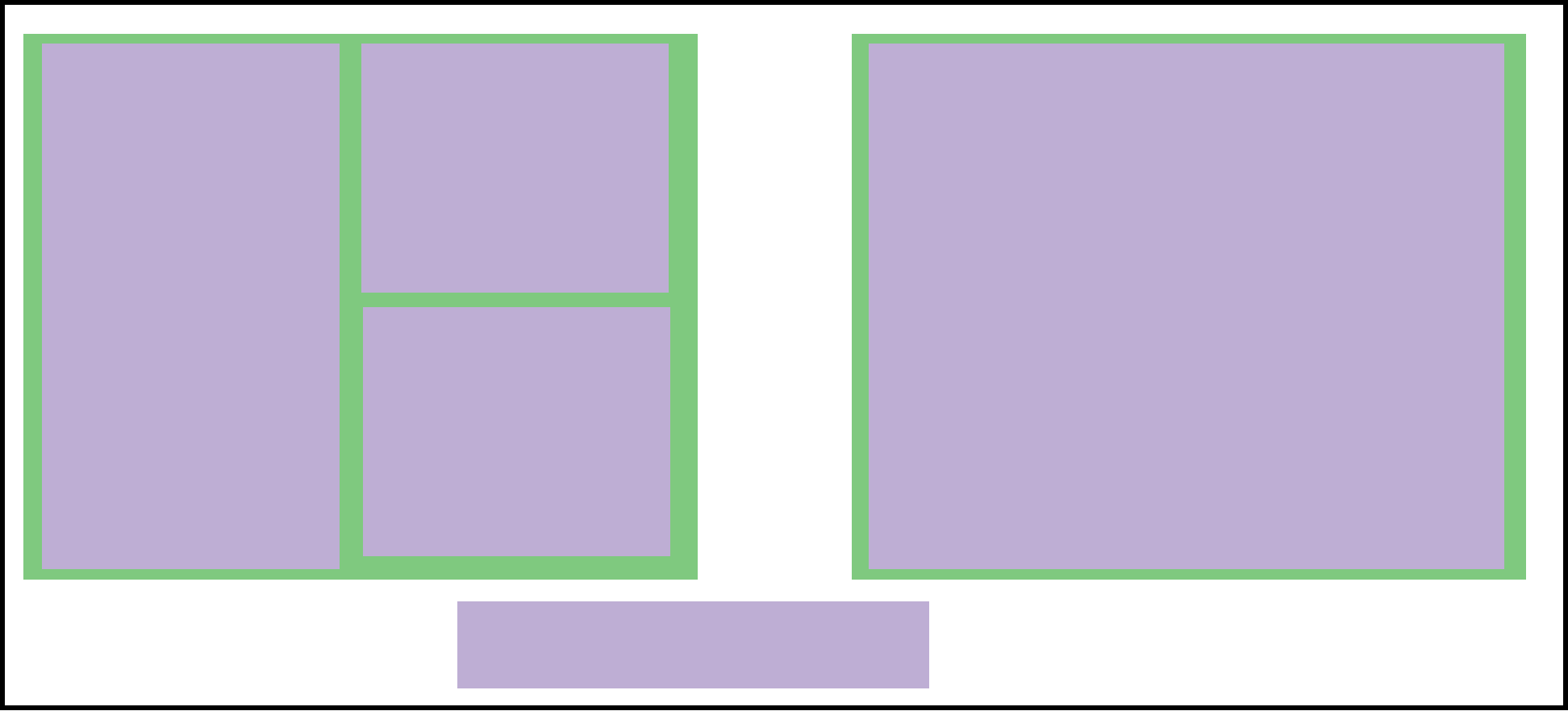}
        \label{fig:toy_o}
    \end{subfigure}
     \begin{subfigure}[b]{0.24\columnwidth}
        \centering          
        \caption{ }\vspace{-7pt}
        \includegraphics[width=\columnwidth]{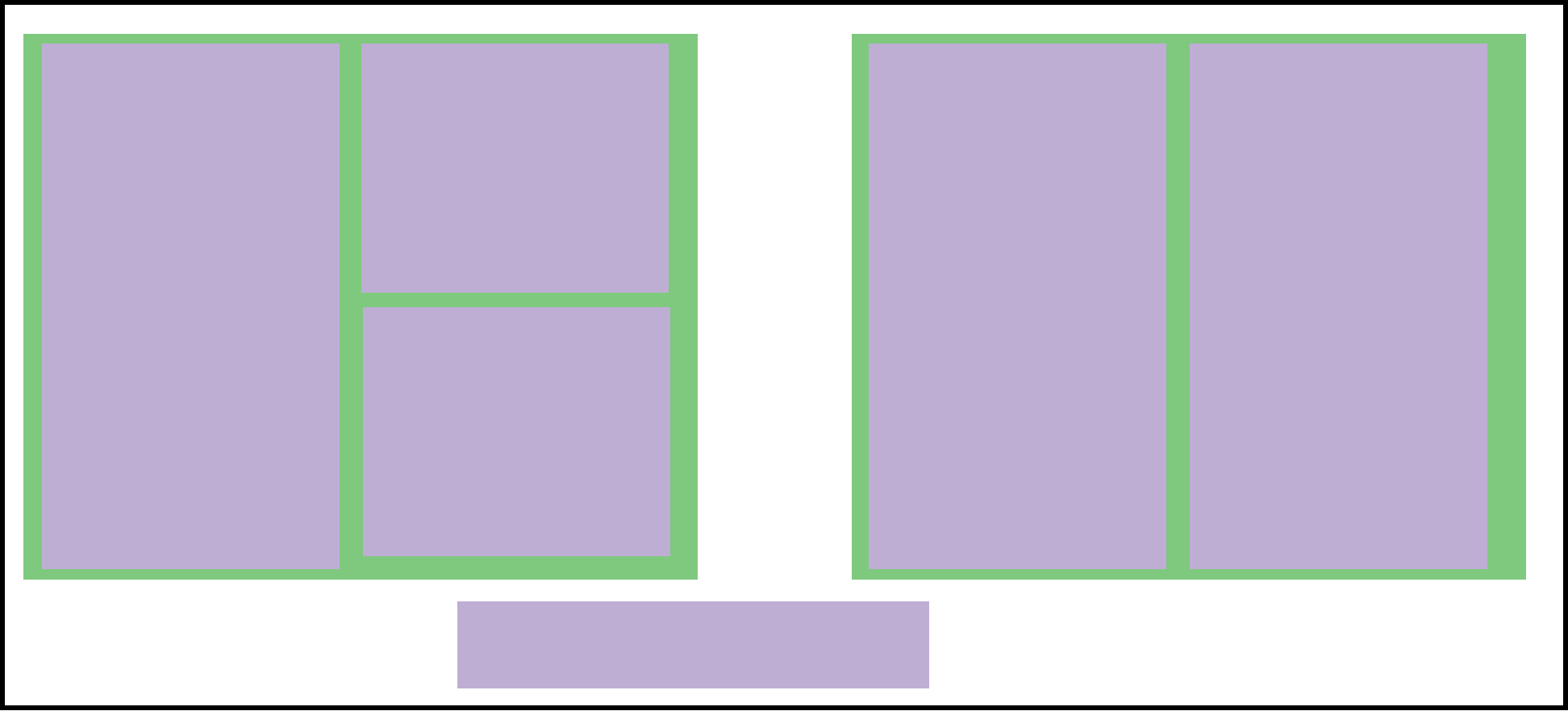}
        \label{fig:toy_p}
    \end{subfigure}
    \vfill
    \begin{subfigure}[b]{1.9\columnwidth}
        \centering
        \resizebox{\columnwidth}{!}{
        \begin{tabular}{l|cccccccccccccccc}
        \toprule
           \textbf{Metric}  &  \textbf{a} & \textbf{b} & \textbf{c} & \textbf{d} & \textbf{e} & \textbf{f} & \textbf{g} & \textbf{h} & \textbf{i} & \textbf{j} & \textbf{k} & \textbf{l} & \textbf{m} & \textbf{n} & \textbf{o} & \textbf{p} \\
           \midrule
           \textbf{IOU Error} & 1.00 & 0.53 & 0.07 & 0.11 & 0.58 & 0.51 & 0.29 & 0.58 & 0.13 & 0.16 & 0.61 & 0.54 & 0.33 & 0.60 & 0.17 & 0.20 \\
            \textbf{GCE}  & 1.00 & 0.27 & 0.07 & 0.12 & 0.34 & 0.38 & 0.29 & 0.26 & 0.12 & 0.14 & 0.40 & 0.42 & 0.33 & 0.32 & 0.17 & 0.20 \\
            \textbf{PE-OS}         & 1.00 & 0.53 & 0.07 & 0.06 & 0.78 & 0.61 & 0.49 & 0.78 & 0.33 & 0.58 & 0.78 & 0.61 & 0.49 & 0.78 & 0.33 & 0.58 \\
            \textbf{1-$AP^{IOU=.50}$}    & 1.00 & 0.00 & 0.00 & 0.33 & 0.50 & 0.50 & 0.33 & 0.67 & 0.50 & 0.60 & 0.67 & 0.67 & 0.50 & 0.75 & 0.60 & 0.67 \\
            \textbf{1-$AP^{IOU=.75}$}    & 1.00 & 0.00 & 0.00 & 0.33 & 1.00 & 1.00 & 1.00 & 1.00 & 0.75 & 1.00 & 1.00 & 1.00 & 1.00 & 1.00 & 0.8 & 1.00 \\
            \textbf{1-PQ }    & 1.00 & 0.67 & 0.51 & 0.64 & 0.76 & 0.75 & 0.71 & 0.82 & 0.69 & 0.74 & 0.84 & 0.82 & 0.78 & 0.86 & 0.76 & 0.79 \\
            \textbf{ROM} (Ours) & 0.00 & 0.00 & 0.00 & 0.00 & 0.46 & 0.46 & 0.96 & 0.76 & 0.64 & 0.99 & 0.32 & 0.32 & 0.91 & 0.64 & 0.54 & 0.98 \\
            \bottomrule
        \end{tabular}
        }
    \end{subfigure}
    \caption{
    Synthetic examples that illustrate and compare metrics. While these primarily address  over-segmentation, they can be generalized to under-segmentation by inverting color interpretation of segmentation and ground-truth.  Each panel represents an image overlay, with ground-truth regions (denoted in \colorbox{ground}{green}) and predicted regions (denoted in \colorbox{pred}{purple}). Panels are designed to represent various situations: (\ref{fig:toy_a}) no positive class prediction at all; (\ref{fig:toy_b}) near perfect prediction for one ground-truth region; (\ref{fig:toy_c}) near perfect prediction; panels (\ref{fig:toy_e} through \ref{fig:toy_j}) different aspects of over-segmentation, with panel (\ref{fig:toy_j}) representing the worst over-segmentation case among those; panels (\ref{fig:toy_k}) through (\ref{fig:toy_p}) correspond to panels (\ref{fig:toy_e}) through (\ref{fig:toy_j}), respectively, but with an additional false-positive predicted region. These metrics include IOU error, GCE~\cite{martin2001database}, Persello's error (PE)~\cite{persello2009novel}, average precision (AP)~\cite{chen2015microsoft}, panoptic quality (PQ) ~\cite{kirillov2019panoptic} and ROM (ours). Note that IOU error is the representative pixel-wise measure, while GCE, PE, AP and ROM are region-wise measures for semantic segmentation; PQ is a metric for the related instance segmentation task, not semantic segmentation.
    }
    \label{fig:simplified_exampls}
\end{figure*}

\vspace{1mm}
\noindent \textbf{IOU and other pixel-wise similarity metrics.}  IOU evaluates pixel-wise similarity irrespective of how pixels are grouped in bounded regions. It is the most common metric for evaluating classification accuracy and is the performance metric of choice for models trained with strong pixel-level supervision. Given a segmentation mask $S_I$ and the ground-truth mask $G_I$ for image $I$, the IOU is calculated as $\frac{S_I \cap G_I}{S_I\cup G_I}$. This metric provides a measure of similarity ranging from 0 (when there is no similar pixel labeling between $G_I$ and $S_I$) to 1 (when there is full concordance between the two pixel assignments $G_I$ and $S_I$). 

IOU gives a measure of pixel-wise assignment similarity that is associated, but not always correlated, with the accuracy of region classification in the single class case. Due to this relationship, the metric is also used as a surrogate for assessing segmentation performance
in various segmentation datasets (e.g., PASCAL VOC ~\cite{everingham2010pascal} and MS-COCO ~\cite{chen2015microsoft}).  
For instance, IOU correctly identifies the difference in overlap between predicted regions and ground-truth in synthetic examples in Figures \ref{fig:toy_c} (lowest IOU error) and \ref{fig:toy_a} (highest IOU error). Additionally, it appropriately penalizes and differentiates models that tend to false-predict regions. However, the correlation between accurate pixel classification and proper segmentation extends only in cases where segmented regions are contiguously uninterrupted, i.e., where it is appropriate to concatenate regions in which geometrically adjacent pixels are similarly labeled. The \textit{horse} leg segmentation (second example in Figure \ref{fig:result_color}) and the \textit{towel} class segmentation (third example in Figure \ref{fig:result_color}) apply exactly in this instance, where IOU does not dovetail with region agreement between segmentation mask and ground-truth. As shown in the synthetic example in Figures \ref{fig:toy_b}, \ref{fig:toy_e}, and \ref{fig:toy_h}, the metric is also not perturbed by significant region-wise refinements.

Other pixel-wise metrics similar to IOU (e.g., pixel accuracy and dice score \cite{long2015fully,taha2015metrics}) also measure pixel-level correspondence between the predicted segmentation mask and ground-truth. These metrics account for the correctness of pixel-based prediction at a global level but again fail to capture region-based agreement between predicted regions and ground-truth. They also fail to robustly capture or differentiate models that provide improvements in region boundary agreement with ground-truth (see Figure \ref{fig:toy_k} and Figure \ref{fig:toy_n}). 

\vspace{1mm}
\noindent \textbf{GCE.} Martin et al. \cite{martin2001database} proposed region-based metrics for evaluating segmentation performance. The global consistency error (GCE) and local consistency error (LCE) were introduced to measure the region-wise inconsistency between a prediction mask, $S_I$, and a ground-truth, $G_I$. 
For a given pixel $x_i$, let the class assignment of $x_i$ in prediction mask $S_I$ be denoted by $C(S_I, x_i)$. Likewise, the class assignment for $x_i$ in the ground-truth, $G_I$, is represented by $C(G_I, x_i)$.
The Local Refinement Error (LRE) is defined at pixel $x_i$ as:
\begin{equation}
    LRE(S_I, G_I, x_i) = \frac{|C(G_I, x_i) \setminus C(S_I,x_i)|}{|C(G_I, x_i)|}
\end{equation}
Since the LRE is one-directional and some inconsistencies arise where the set difference $|C(S_I, x_i) \setminus C(G_I,x_i)|$ accounts for greater divergence between segmentations, the Global Consistency Error (GCE) was defined for an entire image with N pixels as: 
\begin{equation}
    \begin{array}{rl}
        GCE(S_I, G_I) = \frac{1}{N} \text{min} & \left(\sum_{i=1}^{N} LRE(S_I,G_I,x_i) \right.,  \\
         & \left. \sum_{i=1}^{N} LRE(G_I,S_I,x_i) \right)
    \end{array}
\end{equation}
GCE more stringently accounts for consistencies between the two segmentation masks than LCE. GCE, like pixel-based measures, still captures global pixel classification error (similar high and low measures are observed for Figures \ref{fig:toy_a} and \ref{fig:toy_c}, respectively). It also offers some additional penalty for false positive predictions (note difference between GCE and IOU for Figures \ref{fig:toy_e} and \ref{fig:toy_k}). Unlike pixel-wise methods, this metric amplifies disagreement between prediction and ground-truth regions. However, GCE still fails to capture the difference between predictions that refine region segmentation (for example, the GCE for Figure \ref{fig:toy_j} is worse than for \ref{fig:toy_i},  but not by much).

\vspace{1mm}
\noindent \textbf{Partition distance.} In ~\cite{polak2009evaluation}, an error measure was proposed based on partition distance, which counts the number of pixels that must be removed or moved from the prediction for it to match the ground-truth. The partition distance penalizes for under- and over-segmentation to some extent, but partitioning a region in either the prediction or ground-truth into multiple subsets will render the same partition distance. The partition distance tracks much like the GCE for that reason, so we do not calculate it here.

\noindent \textbf{Persello's error.} In~\cite{persello2009novel}, PE was proposed to measure the local error for over-segmentation (PE-OS) and under-segmentation (PE-US) based on the ratio of the area of the largest overlapping prediction region and the area of a ground-truth region. The measure identifies region agreement and penalizes for larger sizes of discrepancy. For each ground-truth region, the error is computed based on the ratio of the area of the largest overlapping predicted region and the area of the ground-truth region. PE-OS increases with over-segmentation (e.g., Figure \ref{fig:toy_b} has  a better score  than either Figure \ref{fig:toy_e} or Figure \ref{fig:toy_h}). However, it provides the same score for Figures \ref{fig:toy_e} and \ref{fig:toy_h} because they have the same-sized largest predicted region for the given ground-truth region.

\vspace{1mm}
\noindent \textbf{Average precision.} AP is the evaluation metric used in the MS-COCO object detection and instance segmentation challenge~\cite{chen2015microsoft}. Instead of directly measuring pixel-wise concordance, AP uses IOU as a threshold and measures region-wise concordance. If a given prediction region's IOU with a ground-truth region exceeds the threshold, it is counted as a true positive (TP); otherwise, it is counted as a false positive (FP). The precision $\left(\frac{TP}{TP+FP}\right)$ is computed and averaged over multiple IOU values, specifically from 0.50 to 0.95 with 0.05 increments. We computed $AP^{IOU=.50}$  (same as PASCAL VOC metric ) and $AP^{IOU=.75}$ (strict metric) for the synthetic examples in Figure~\ref{fig:simplified_exampls}. AP has its advantages over pure pixel-wise measures when assessing error instance-wise, but it will fail when two segmentation masks have similar precision values but suffer from different degrees of over-segmentation. For example, $AP^{IOU=.50}$ gives the same error measure for Figure \ref{fig:toy_h}, \ref{fig:toy_k}, \ref{fig:toy_l} and \ref{fig:toy_p}, but Figure \ref{fig:toy_h} is considered a worse over-segmentation case than \ref{fig:toy_l} and \ref{fig:toy_k}, while \ref{fig:toy_p} represents the worst over-segmentation case among them. In terms of the stricter measure, $AP^{IoU=.75}$ indicates that the majority of these examples have an equal error of 1 without differentiating which one is worse. 

\vspace{1mm}
\noindent \textbf{Panoptic quality.} PQ, which combines the semantic segmentation and instance segmentation tasks~\cite{kirillov2019panoptic}, is a relevant metric but requires having ground-truth annotations for both semantic and instance segmentation. It provides a unified score that measures pixel-wise concordance and penalizes false positive (FP) and false negative (FN) regions. PQ reflects the gross pixel-wise and region-wise error jointly in a predicted segmentation, but it does not differentiate between under- and over-segmentation. Although FP and FN regions indirectly relate to the over- and under-segmentation issues, PQ penalizes these errors in the same direction and may give the same score to a prediction that over-segments as to one that under-segments.

\section{Interpretable Region-based Measures for Over- and Under-segmentation Errors}
\label{Method}
We now introduce two measures that combine the desirable properties of: (1) accounting for model disagreement with ground-truth in over- and under-segmentation cases, and (2) displaying sensitivity for local over- and under-segmentation refinements to model predictions or refinements to ground-truth based on ambiguous semantic segmentation boundaries. These properties are relevant primarily to quantify the consistency of segmentation results (see Section~\ref{qualitative_assessment}). 

\subsection{Region-wise metric estimations }
\label{method_metrics}
We now define metrics that isolate errors due to over- and under-segmentation in region-based tasks like semantic segmentation. 

\vspace{1mm}
\noindent \textbf{Region-wise over-segmentation measure (ROM).} Let $I$ be an RGB image, $G_{I}$ be the ground-truth segmentation mask for $I$, and $S_{I}$ be the model-predicted segmentation mask for $I$ in dataset $\mathcal{D}$ (validation or test set), each with spatial dimensions of $w \times h$, where $w$ and $h$ correspond to width and height, respectively. Assume there are $K$ semantic classes in $\mathcal{D}$. A valid label assignment (also referred to as a segmentation) in $G_{I}$ and $S_{I}$ maps each pixel $x_{r,c}$ to a single integer label $k \in [0,\cdots, K-1]$ representing the class assignment for that pixel. For simplicity, we assume that the background class label is 0. For each non-background class $k \in [1, \cdots, K-1]$, we convert $G_{I}$ and $S_{I}$ to their binary formats, $G_{Ib}$ and $S_{Ib}$ , as follows:
%

\begin{equation}
	G_{Ib}[k, r,c]  = 
	\begin{cases}
		1, & G_I[r, c] == k \\
		0, & \text{Otherwise}
	\end{cases}
\end{equation}
\begin{equation}
	S_{Ib}[k, r,c] = 
	\begin{cases}
		1, & S_I[r, c] == k \\
		0, & \text{Otherwise}
	\end{cases}
\end{equation}
where $r \in [0, h-1]$ and $c \in [0, w-1]$ correspond to row and column indices of a pixel in the image. 

Each spatial plane in $G_{Ib}$, i.e., $G_{Ib}[k]$, consists of $N=\|G_{Ib}[k]\|$ separate contiguous regions, where $\|\cdot\|$ is an operator that counts the number of separate contiguous regions in $G_{Ib}[k]$. Therefore, we can represent each spatial plane $G_{Ib}[k]$ as a set of connected regions $G_{Ib}[k] = \{g_1,g_2,...,g_N\}, k \in [1, K-1]$. We assert that $g_i\cap g_j = \varnothing, \forall i \neq j$. Similarly, we represent each spatial plane in $S_{Ib}$, i.e., $S_{Ib}[k]$ as a set of $M$ connected regions $
S_{Ib}[k] = \{s_1,s_2,...,s_M\}, k \in [1, K-1]$, where $s_i\cap s_j = \varnothing, \forall i \neq j$. We refer to the complete set of binary planes that satisfy these constraints as a valid segmentation ground-truth pair and look for measures of the form $d(S_{Ib},G_{Ib})$.

We begin by evaluating the performance of the prediction mask by making a detailed accounting of over-segmented foreground regions.
A model-predicted region contributes to the over-segmentation count when the prediction region overlaps with a ground-truth region that itself overlaps with more than one model-predicted region. Fig~\ref{fig:oversegexample} shows an example of over-segmentation. 

We denote regions in $S_{Ib}$ contributing to over-segmentation as $S^{I}_O = \{s_i \in S_{Ib}\}$, where $s_i\cap g_l \neq \varnothing \land s_j\cap g_l \neq \varnothing,\ i \in [1,\cdots,M],\ j \in [1,\cdots,M],\ l \in[1, \cdots, N],\ i\neq j$. $S^{I}_O$ marks model-predicted region $s_i \in S_{Ib}$ as included in the over-segmentation count; it must overlap with at least one ground-truth region $g_l \in G_{Ib}$, while the ground-truth region $g_l$ must overlap with at least one other prediction region $s_j \in S_{Ib}$ . The total number of regions that contribute to over-segmentation with respect to ground-truth regions are $\| S^{I}_O \|$. 

Similar accounting identifies ground-truth regions involved in over-segmentation (i.e., overlapping more than one predicted region). We denote regions in  $G_{Ib}$ that are involved in over-segmentation as $G^{I}_O = \{g_i \in G_{Ib}\}$, such that $g_i\cap s_j \neq \varnothing \land g_i\cap s_l \neq \varnothing$, $j\in[1, \cdots, M]$, $l \in [1, \cdots, M]$, and $j\neq l$. This definition asserts that a ground-truth region $g_i \in G_{Ib}$ is counted towards  the over-segmentation count if it overlaps with at least two different model-predicted regions, $s_j \in S_{Ib}$ and $s_l \in S_{Ib}$. $\| G^{I}_O \|$ denotes the total number of ground-truth regions that are involved in over-segmentation.

We are interested in a measure that combines desirable statistical properties with the ability to count disagreements between $S_{Ib}$ and $G_{Ib}$ and accommodate model prediction refinement. 
Specifically, we should not use any thresholds or fixed pixel-percent requirements to determine the overlap between regions in the segmentation mask and ground-truth. This is important in the context of many methods that are used to merge or fuse segmentation results based on geometric proximity or filtering methods \cite{li2015}.

Given the ground-truth labeling, $G_{Ib}$, the probability that a model-predicted region $s_i \in S_{Ib}$ contributes to over-segmentation can be represented as $\frac{\|{S^I_O}\|}{\|S^I_b\|}$, and the probability that a ground-truth region $g_i \in G_{Ib}$ is over-segmented can be represented as $\frac{\|{G^I_O}\|}{\|G^I_b\|}$. Our goal is to express the empirical probability that a predicted segmentation mask is over-segmenting. 
We define this index as the region-wise over-segmentation ratio ($ROR$):  
\begin{equation}
ROR = \frac{\|{G^I_O}\|\|{S^I_O}\|}{\|{G^I_b}\|\|S^I_b\|}.
\end{equation}
This measure takes values between 0 and 1 and indicates the percentage of elements in ${G^I_b}$ and ${S^I_b}$ that relate to over-segmentation. In the worst case, every single model-predicted region and ground-truth region is either contributing to or involved in over-segmentation, giving $ROR$ a value of 1.

$ROR$ may fail to differentiate between cases where any elements in $S^I_O$ are further split by subsets (e.g., Figures \ref{fig:toy_e} and and \ref{fig:toy_h}). Therefore, we seek to penalize the $ROR$ score based on the total number of over-segmenting prediction regions.  Let $S^{I}_{g_i} = \{s_j \in S_{Ib}\}$, such  that $s_j\cap g_i \neq \varnothing$ denotes a set of all predicted regions that overlap with a given $g_i \in G^I_b$. To penalize the $ROR$ score, we compute over-segmentation aggregate  term ($m_o$) from $S^{I}_{g_i}$ for all $g_i \in G_{Ib}$ as:
\begin{equation}
    m_o = \sum_{g_i} max\left(\|S^{I}_{g_i}\|-1, 0.\right) 
\end{equation}
Thus, $m_o$ expresses an aggregate penalty structure accounting for each ground-truth region $g_i \in G_{Ib}$ that is overlapped with at least one model-predicted region. 
To compute the final region-wise over-segmentation measure ($ROM$), we multiply $ROR$ by $m_o$. The resultant value will never be negative because both $ROR$ and $m_o$ are greater than or equal to 0. 
Since the resultant value can be very large, we scale it between 0 and 1 using a $\tanh$ function. 
Doing so confers the property of taking on a wider range of values over $[0,1)$, increasing the sensitivity of the measure to classes that habitually over-segment\footnote{Other scaling functions, such as sigmoid, can be applied. We choose $\tanh{}$ over sigmoid because it is centered at 0.} and allowing us to compare with existing metrics.
\begin{equation}
    ROM = \tanh{(ROR \times m_o)}
\end{equation}
A $ROM$ value of zero indicates that there is no over-segmentation. In this case, $\|{G^I_O}\|\ = \|{S^I_O}\|\ = 0$, indicating that each $g_i \in G^I_b$ overlaps with at most one $s_j \in S^I_b$. A $ROM$ value of 1 indicates that the predicted segmentation is worse and contains abundant over-segmented regions. In this case, $\|{G^I_O}\|\ = \|{G^I_b}\|$, $\|{S^I_O}\|\ = \|{S^I_b}\|$, and $m_o \to \rotatebox{90}{8}$. This means that every single prediction/ground-truth region contributes to over-segmentation, and each ground-truth region overlaps with an infinite number of prediction regions.  

\vspace{1mm}
\noindent \textbf{Region-wise under-segmentation measure (RUM).} A similar argument follows for evaluating $S_I$ from the under-segmentation perspective. A predicted region that contributes to under-segmentation (see example in Fig ~\ref{fig:undersegexample}) is represented as $S^{I}_U = \{s_i \in S_{Ib}\}$, where $\exists k,l \in [1..N],  k\neq l
	s_i\cap g_k \neq \varnothing \land s_i\cap g_l \neq \varnothing $.
The total count of model-segmented regions that contribute to under segmentation is denoted by $\| S^{I}_U \|$. This representation identifies a model-predicted region as under-segmenting when it overlaps with at least two different ground-truth regions $g_k \in G_{Ib}$ and $g_l \in G_{Ib}$. Similarly, a ground-truth region is involved in under-segmentation when it overlaps with a prediction region that in turn overlaps with at least two ground-truth regions. This is represented as $G^{I}_U = 	\{g_i \in G_{Ib}\}$, $	\text{s.t. }\exists j\in[1..N], \ \exists l \in[1..M], \ i\neq j,\ s_l\cap g_i \neq \varnothing \land s_l\cap g_j \neq \varnothing$. 
This representation counts ground-truth regions $g_i \in G_{Ib}$ that overlap with at least one prediction region $s_l \in S_{Ib}$, while the prediction region $s_l$ overlaps with at least one other prediction region $g_j \in G_{Ib}$. The total number of ground-truth regions involved in under-segmentation is denoted by $\| G^{I}_U \|$.

The analog probabilistic terms for the region-wise under-segmentation ratio ($RUR$), under-segmentation multiplier ($m_u$), and region-wise over-segmentation measure ($RUM$) are defined as:
\begin{equation}
    RUR = \frac{\|{G^I_U}\|\|{S^I_U}\|}{\|{G^I_b}\|\|S^I_b\|}    
\end{equation}
\begin{equation}
   m_u = \sum_{s_i}max(\|G^{I}_{s_i}\|-1, 0) 
\end{equation}
\begin{equation}
    RUM = \tanh{(RUR \times m_u)}
    \label{eq:rum}
\end{equation}
 
\subsection{Region-wise confidence measure}
\label{confidence} 
Many segmentation models confer a prediction class for each pixel and an associated probability (confidence) for each pixel prediction. Instead of looking at the confidence of each pixel individually (as was done in the PASCAL VOC object detection evaluation~\cite{everingham2010pascal}), we propose to use the confidence of each predicted contiguous region, when available. We represent the confidence of a predicted region as the numerical mean of the confidence of all pixels enclosed in that region. When evaluating with $ROM$ (or $RUM$), all pixels in a region whose confidence is lower than a certain threshold are mapped to the class \textit{unknown}. We experiment with the effect of different thresholds in Section~\ref{sec:experiments}. 

\subsection{Qualitative assessment}

\label{qualitative_assessment}
Here, we qualitatively demonstrate that the ROM/RUM metric can: (1) differentiate between models that provide fewer or more numerous over-/under-segmentation inconsistencies, and (2) adequately quantify improvements to model predictions while providing robust tolerance to small perturbations in ground-truth regions.

We highlight that $ROM$ (or $RUM$) cannot be used as measures of classification accuracy that account for gross error. For example, panels in Figure \ref{fig:toy_a} through Figure \ref{fig:toy_d} are equivalent from the perspective of over-segmentation error ($ROM$ is 0 for all), indicating the lack of over-segmentation errors in all of these examples (whether due to mis-classification or not).
As demonstrated in Section \ref{sec:related_work}, widely available pixel-wise metrics can be used in conjunction with $ROM$ and $RUM$ to address these concerns. Future work might consider how to combine these metrics in a principled way. 

The advantage of the $ROM$ metric is that it isolates disagreement and correlates with discrepancies between model and ground-truth due to region-wise over-segmentation. The analog is true for $RUM$ with respect to under-segmentation. Specifically, accounting only for over-segmentation, panels in Figure \ref{fig:toy_e} and Figure \ref{fig:toy_f} are equivalent. Moreover, the metric is useful in accounting for over-segmentation in the average model-predicted region, therefore indicating higher over-segmentation in panels in Figure \ref{fig:toy_e} and Figure \ref{fig:toy_f} versus panels  in Figure \ref{fig:toy_k} and Figure \ref{fig:toy_l}. This demonstrates the ability to quantify model differences that provide fewer or more numerous over-segmentation inconsistencies. Finally, as $ROM$ error increases in panels in Figure \ref{fig:toy_b}, \ref{fig:toy_e} and \ref{fig:toy_h}, we see that the $ROM$ differentiates and grows monotonically with over-segmentation errors in model predictions.

Notably, our approach yields a supervised objective evaluation and highlights discrepancies in overall region segmentation results, comparing it to a ground-truth. 
Calculations for $ROM$ and $RUM$ specifically avoid penalizing pixel-wise non-concordance in $g_i\cap s_j = \varnothing$ because such information has been handled by other metrics (like IOU). $ROM$ and $RUM$ help solely with evaluation from the over-and under-segmentation perspective, respectively.

\section{Experimental Set-up}
\label{sec:setup}
To demonstrate the effectiveness of our proposed metrics, we evaluate and compare them with existing metrics across different semantic segmentation datasets and different segmentation models. 

\vspace{1mm}
\noindent \textbf{Baseline metrics:} We use the following metrics for comparison: (1) mIOU error, (2) Dice error, (3) pixel error, (4) global consistency error (GCE), (5) Parsello's error (PE-OS and PE-US for over- and under-segmentation), and (6) Average precision error ($1- AP^{IOU=.50:.05:.95}$). These metrics report a similarity score between 0 and 1, where 0 means that predicted mask $S_I$ and ground-truth mask $G_I$ for image $I$ are the same, while 1 means they are not similar. 


\begin{figure*}[htbp!]
    \centering
    \begin{subfigure}[b]{2\columnwidth}
        \begin{subfigure}[b]{0.32\columnwidth}
            \centering
            \includegraphics[width=\columnwidth]{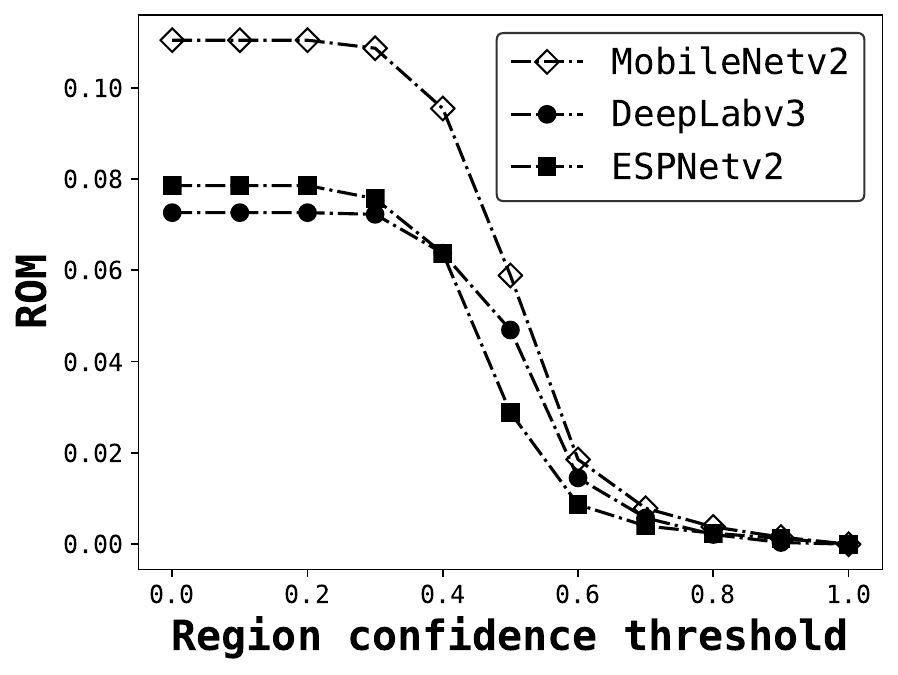}
        \end{subfigure}
        \hfill
        \begin{subfigure}[b]{0.32\columnwidth}
            \centering
            \includegraphics[width=\columnwidth]{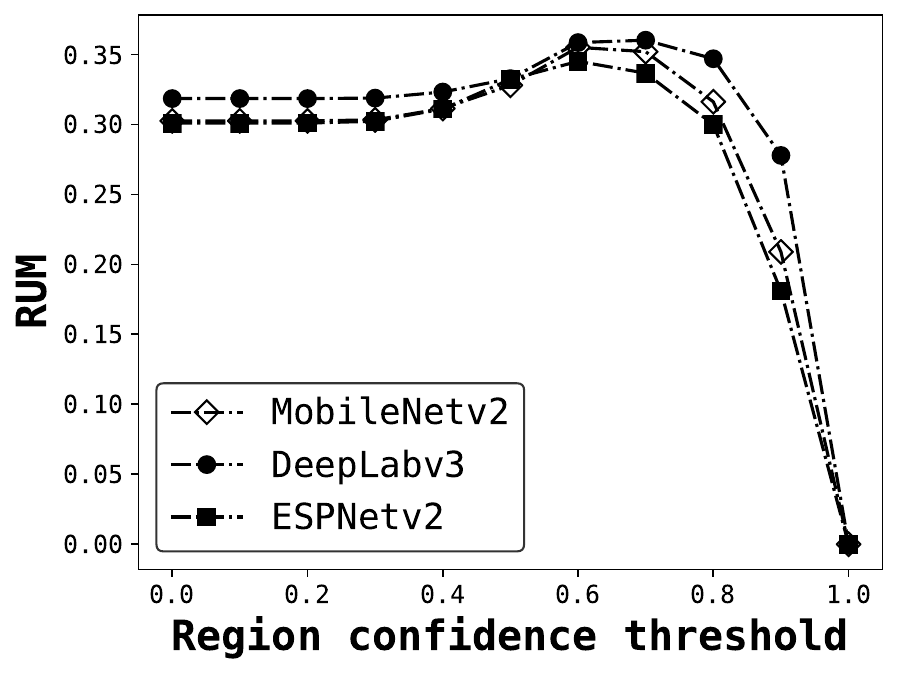}
        \end{subfigure}
        \hfill
        \begin{subfigure}[b]{0.32\columnwidth}
            \centering
            \includegraphics[width=\columnwidth]{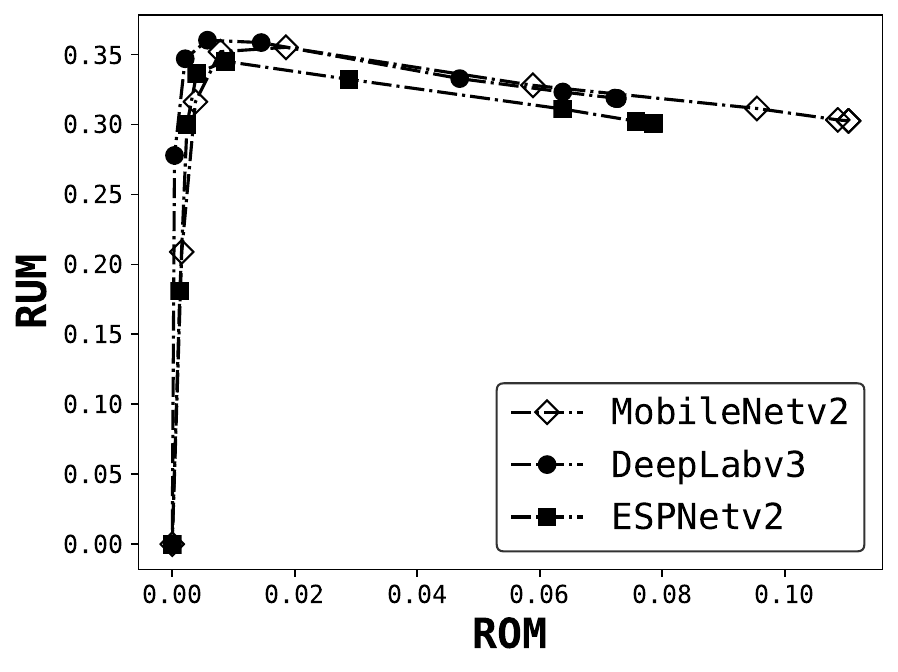}
        \end{subfigure}
        \caption{The PASCAL VOC dataset}
        \label{fig:roc_pascal}
    \end{subfigure}
    \vfill
    \begin{subfigure}[b]{2\columnwidth}
        \begin{subfigure}[b]{0.32\columnwidth}
            \centering
            \includegraphics[width=\columnwidth]{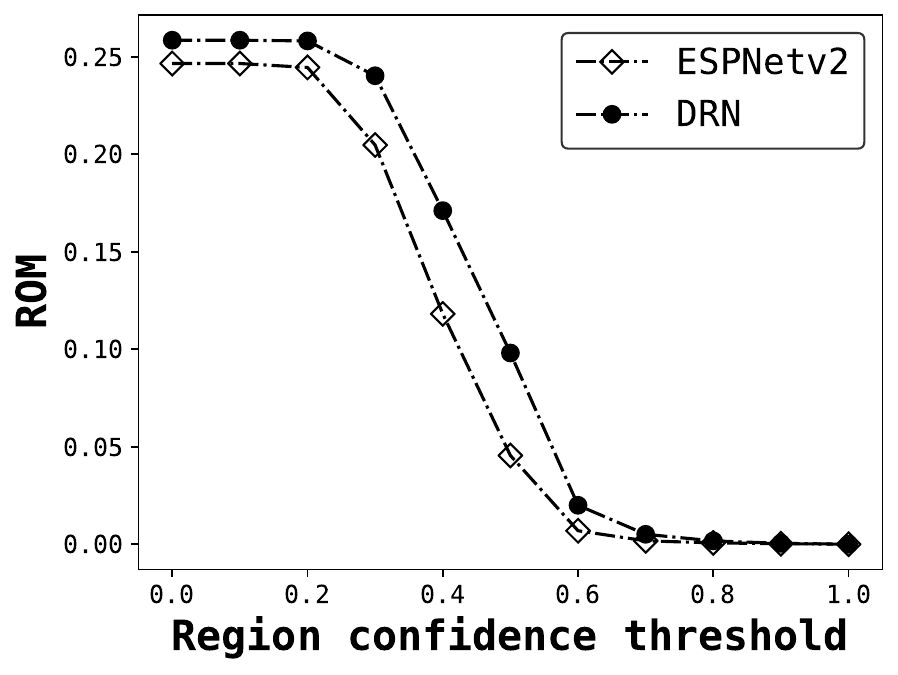}
        \end{subfigure}
        \hfill
        \begin{subfigure}[b]{0.32\columnwidth}
            \centering
            \includegraphics[width=\columnwidth]{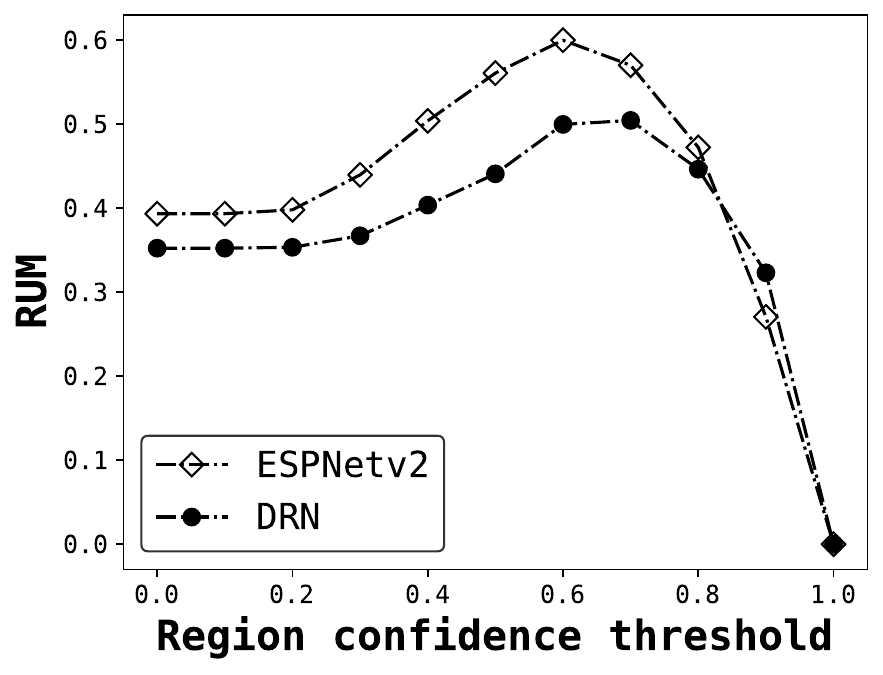}
        \end{subfigure}
        \hfill
        \begin{subfigure}[b]{0.32\columnwidth}
            \centering
            \includegraphics[width=\columnwidth]{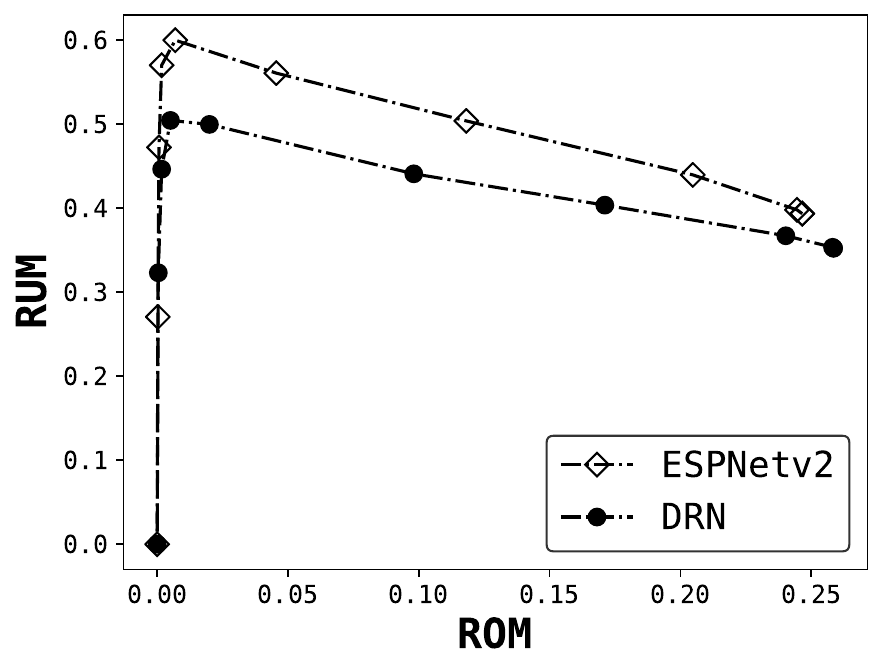}
        \end{subfigure}
        \caption{The Cityscapes dataset}
        \label{fig:roc_city}
    \end{subfigure}
    \vfill
    \begin{subfigure}[b]{2\columnwidth}
        \begin{subfigure}[b]{0.32\columnwidth}
            \centering
            \includegraphics[width=\columnwidth]{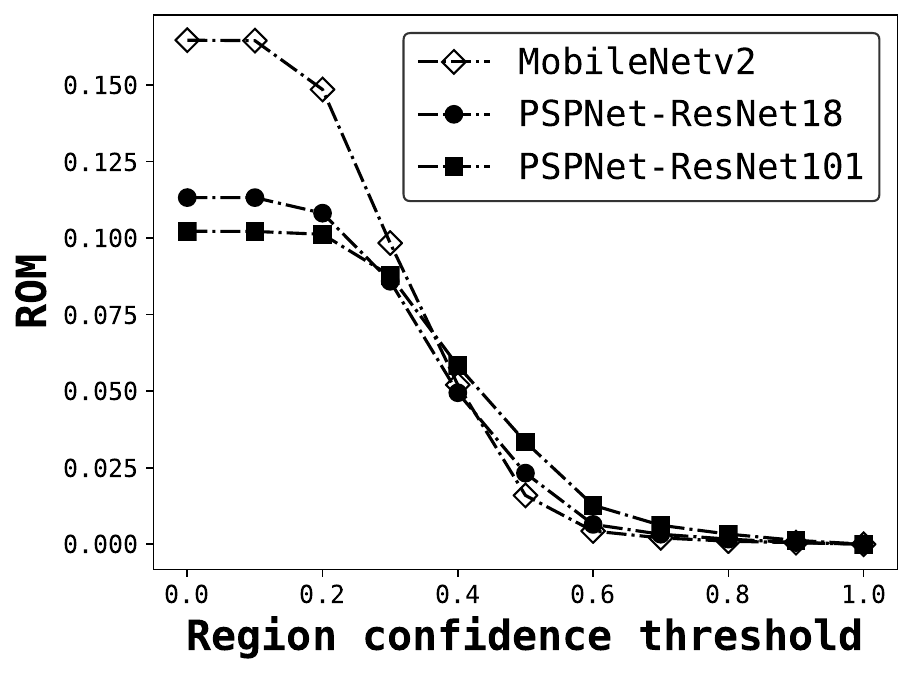}
        \end{subfigure}
        \hfill
        \begin{subfigure}[b]{0.32\columnwidth}
            \centering
            \includegraphics[width=\columnwidth]{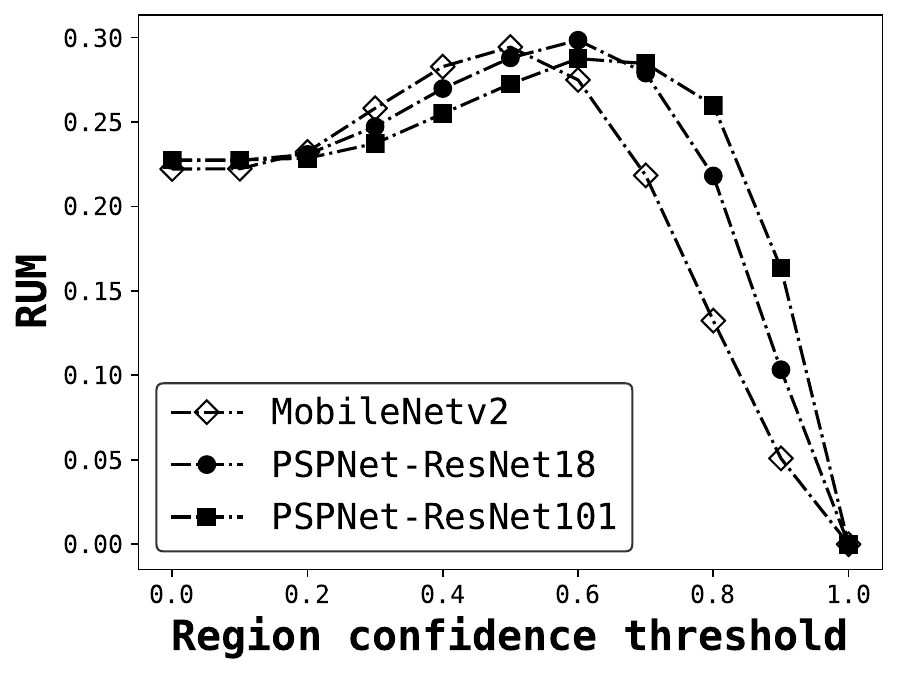}
        \end{subfigure}
        \hfill
        \begin{subfigure}[b]{0.32\columnwidth}
            \centering
            \includegraphics[width=\columnwidth]{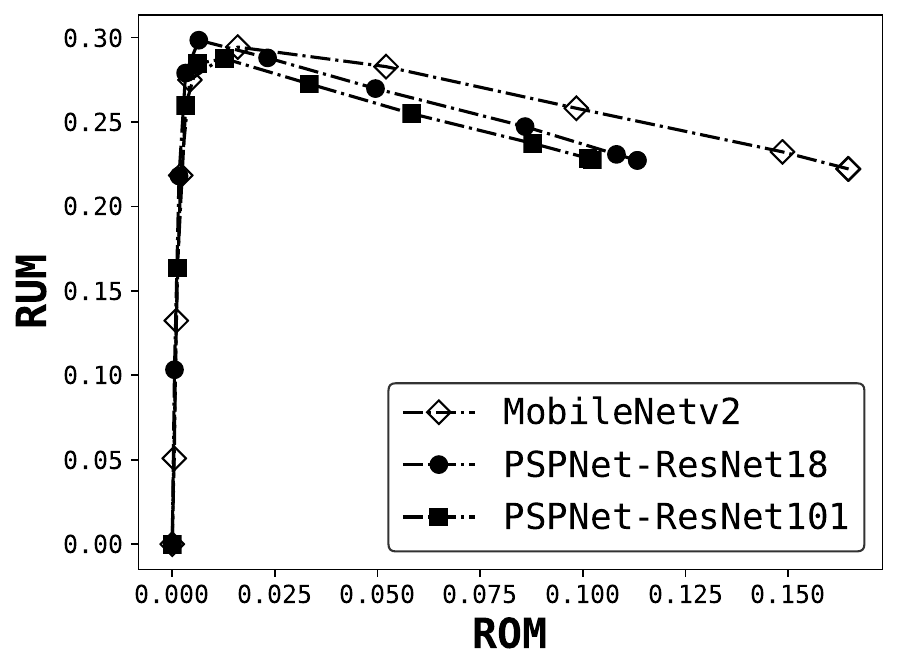}
        \end{subfigure}
        \caption{The ADE20K dataset}
        \label{fig:roc_ade20k}
    \end{subfigure}
    \caption{ROC curves for over- (ROM) and under-segmentation (RUM) on different datasets. Note the rightmost plot in each sub-figure is between ROM and RUM errors. Therefore, a lower area under ROC curve means better performance.}
    \label{fig:main_results}
\end{figure*}

\begin{figure*}[htbp!]
    \centering

    \begin{subfigure}[b]{2\columnwidth}
        \centering
        \begin{subfigure}[b]{0.23\columnwidth}
            \centering
            \caption*{\bf RGB Image}
        \end{subfigure}
        \begin{subfigure}[b]{0.23\columnwidth}
            \centering
            \caption*{\bf Ground truth}
        \end{subfigure}
        \begin{subfigure}[b]{0.23\columnwidth}
            \centering
            \caption*{\bf Prediction}
        \end{subfigure}
        \begin{subfigure}[b]{0.28\columnwidth}
            \centering
            \caption*{\bf Error metrics}
        \end{subfigure}
    \end{subfigure}
    
    \begin{subfigure}[b]{2\columnwidth}
        \centering
        \begin{subfigure}[b]{0.23\columnwidth}
            \centering
            \includegraphics[width=\columnwidth]{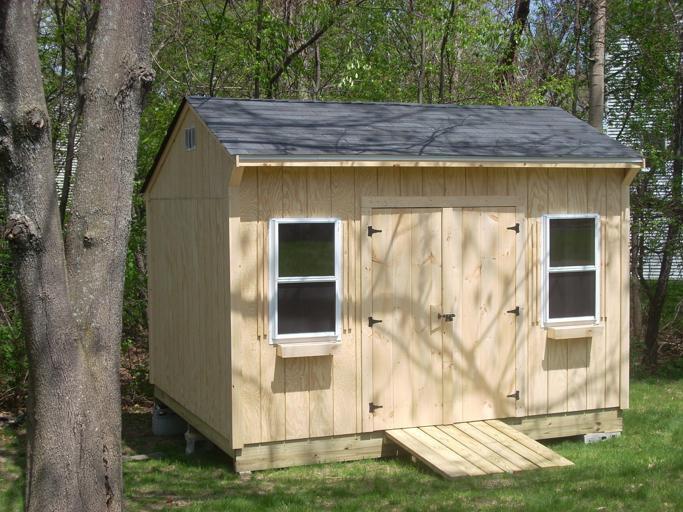}   
        \end{subfigure}
        \begin{subfigure}[b]{0.23\columnwidth}
            \centering

            \includegraphics[width=\columnwidth]{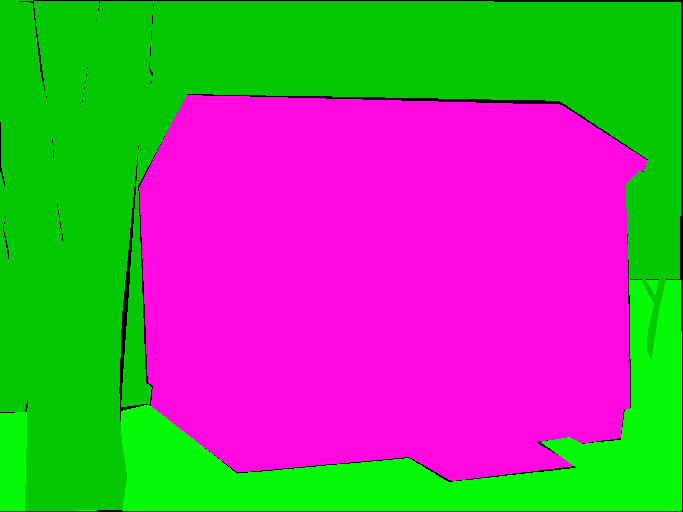}
        \end{subfigure}
        \begin{subfigure}[b]{0.23\columnwidth}
            \centering

             \includegraphics[width=\columnwidth]{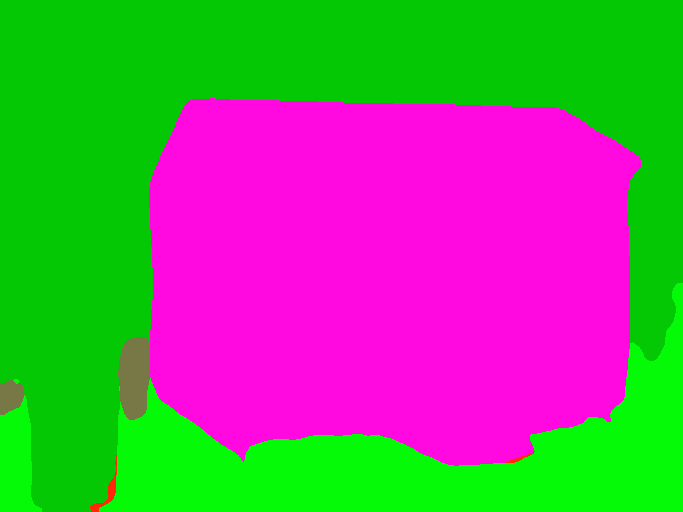}
        \end{subfigure}
        \begin{subfigure}[b]{0.28\columnwidth}
            \centering

            \includegraphics[width=\columnwidth]{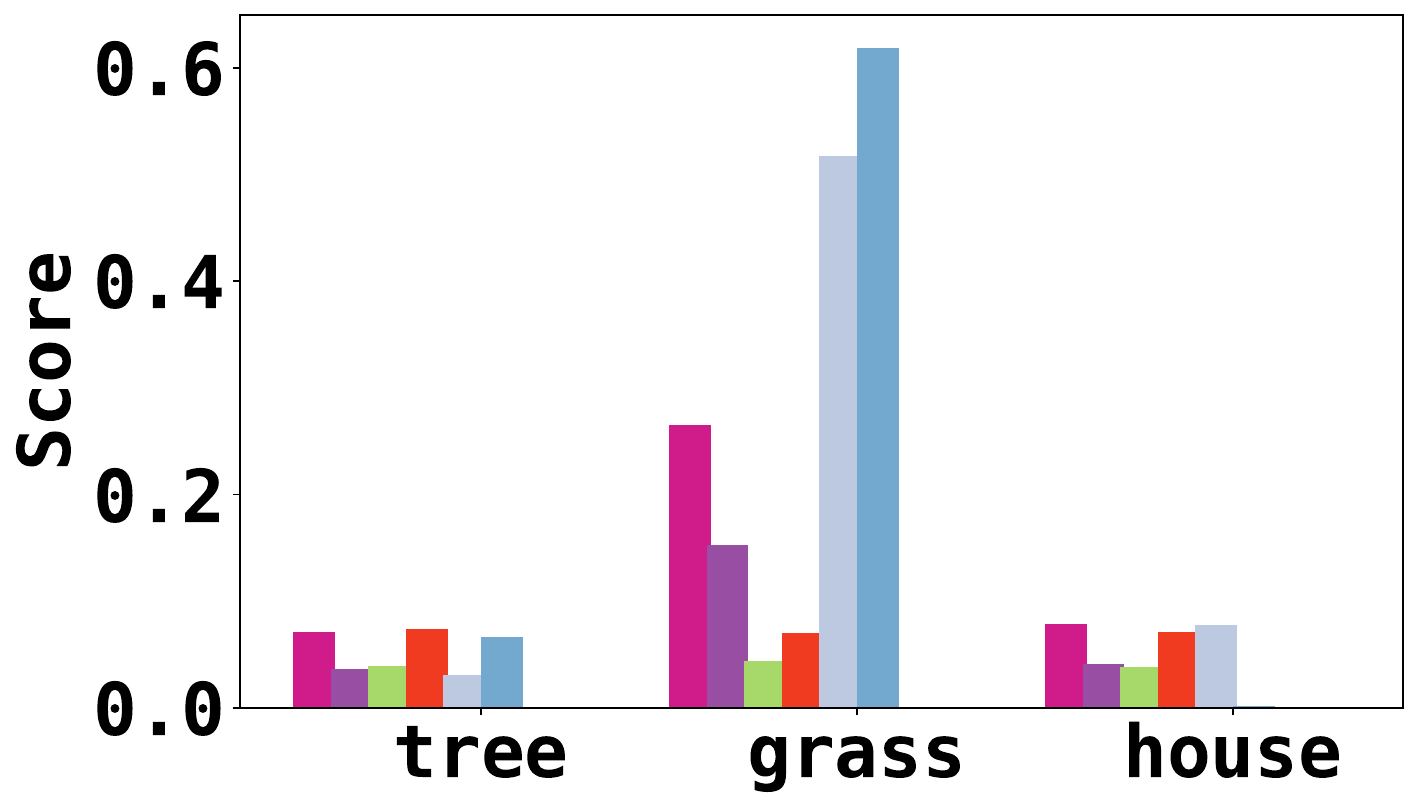}
        \end{subfigure}
         \vspace{2mm}%
    \end{subfigure}

    \begin{subfigure}[b]{2\columnwidth}
        \centering
        \begin{subfigure}[b]{0.23\columnwidth}
            \centering
            \includegraphics[width=\columnwidth]{results_pascal/rgb/2007_008802.jpg}
        \end{subfigure}
        \begin{subfigure}[b]{0.23\columnwidth}
            \centering
            \includegraphics[width=\columnwidth]{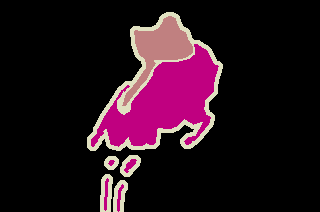}
        \end{subfigure}
        \begin{subfigure}[b]{0.23\columnwidth}
            \centering
            \includegraphics[width=\columnwidth]{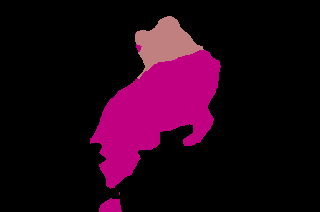}
        \end{subfigure}
        \begin{subfigure}[b]{0.28\columnwidth}
            \centering
            \includegraphics[width=\columnwidth]{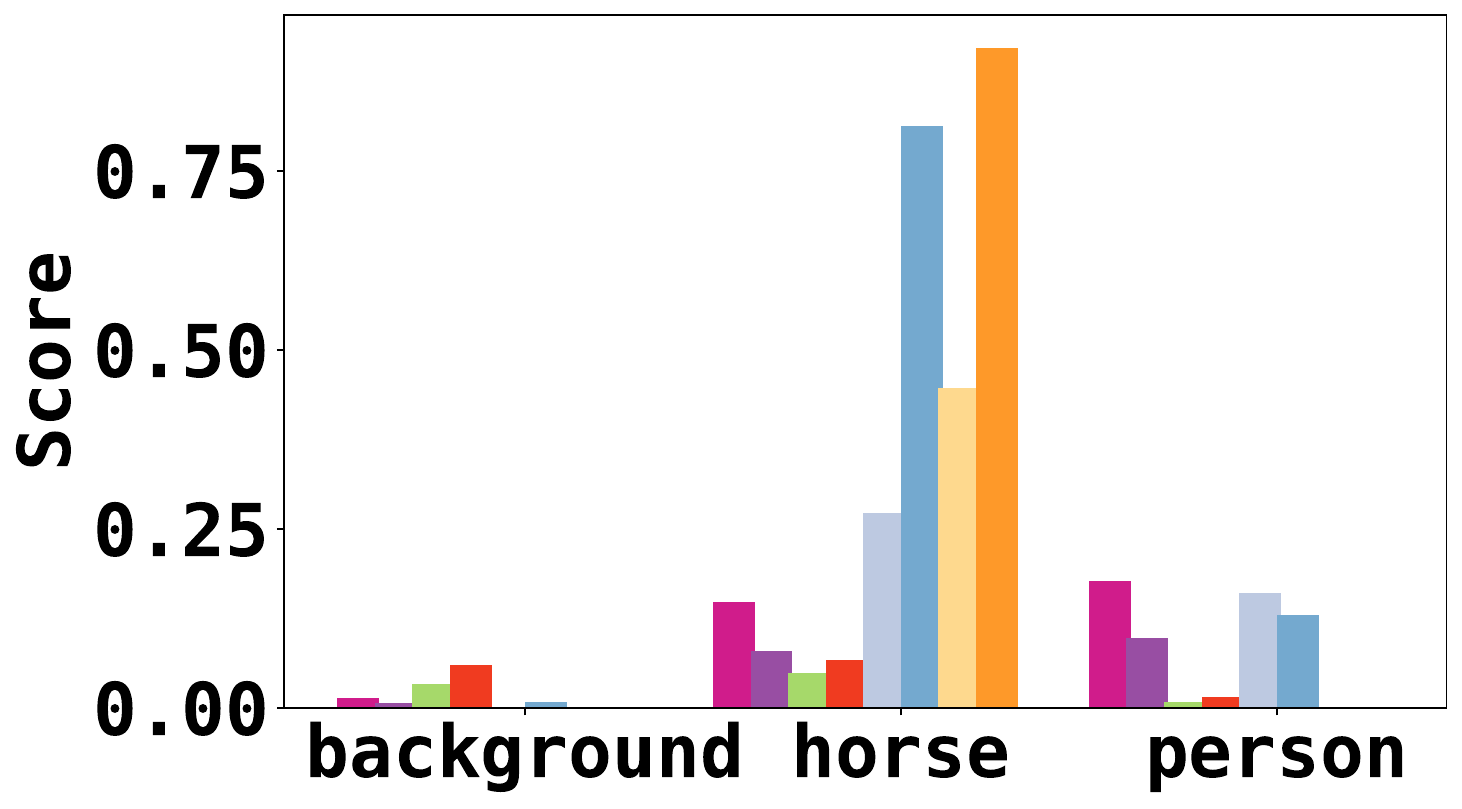}
        \end{subfigure}
        \vspace{2mm}%
    \end{subfigure}
    
    \begin{subfigure}[b]{2\columnwidth}
        \centering
        \begin{subfigure}[b]{0.23\columnwidth}
            \centering
            \includegraphics[width=\columnwidth]{results_ade/rgb/ADE_val_00001103.jpg}
        \end{subfigure}
        \begin{subfigure}[b]{0.23\columnwidth}
            \centering
            \includegraphics[width=\columnwidth]{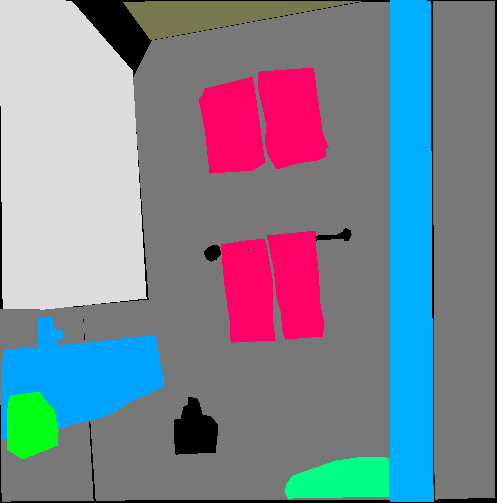}
        \end{subfigure}
        \begin{subfigure}[b]{0.23\columnwidth}
            \centering
            \includegraphics[width=\columnwidth]{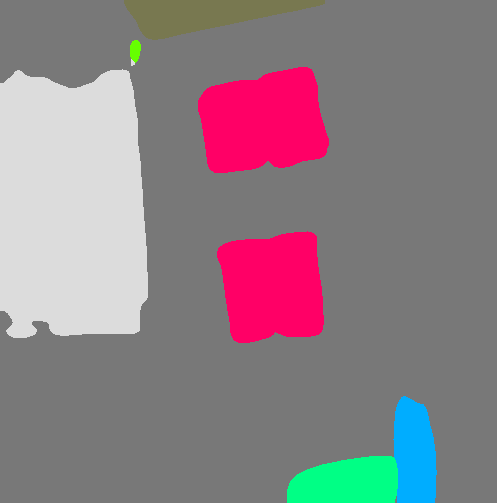}
        \end{subfigure}
        \begin{subfigure}[b]{0.28\columnwidth}
            \centering
            \includegraphics[width=\columnwidth]{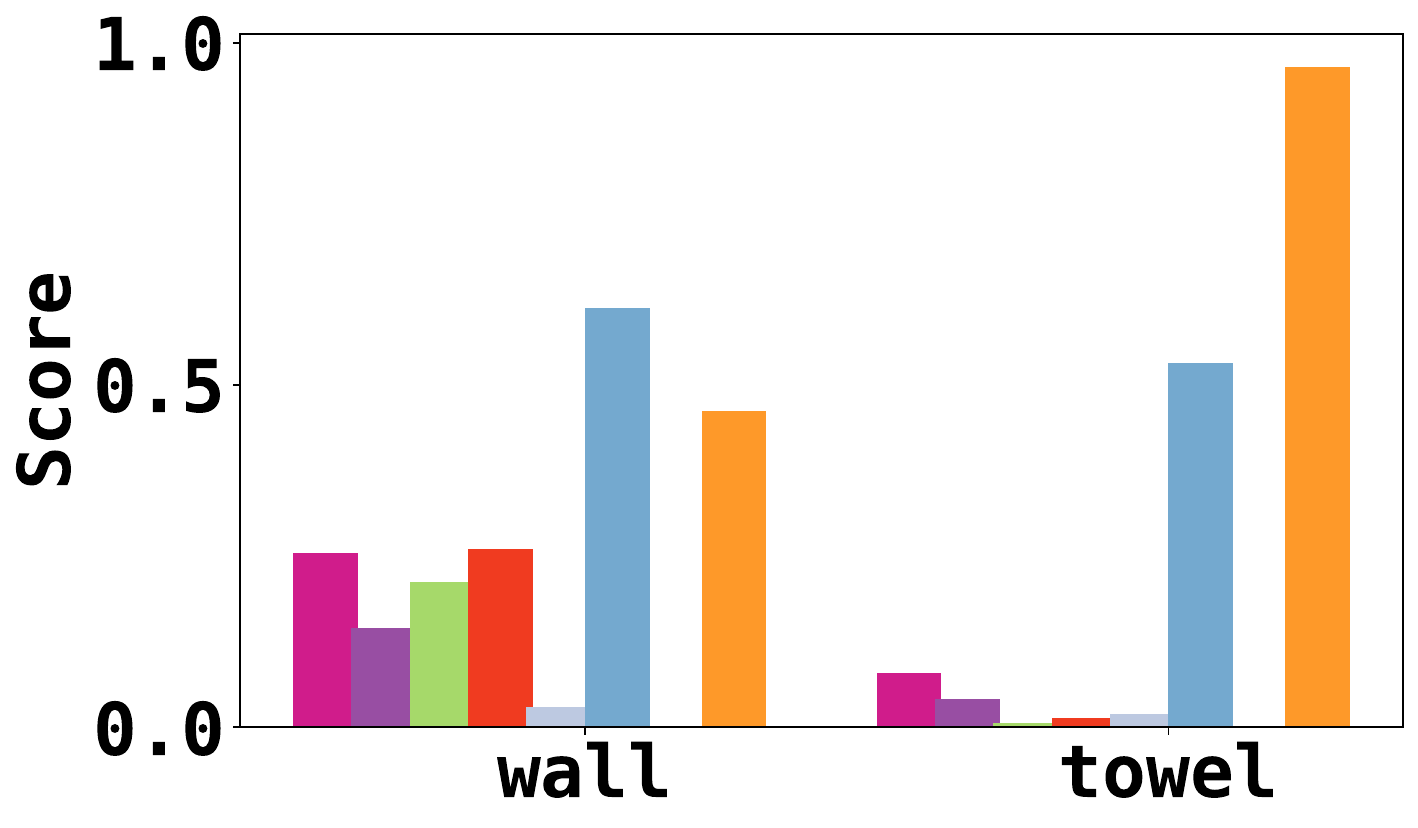}
        \end{subfigure}
        \vspace{2mm}%
    \end{subfigure}
    
    \begin{subfigure}[b]{2\columnwidth}
        \centering
        \begin{subfigure}[b]{0.23\columnwidth}
            \centering
            \includegraphics[width=\columnwidth, height=75px]{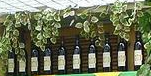}
        \end{subfigure}
        \begin{subfigure}[b]{0.23\columnwidth}
            \centering
            \includegraphics[width=\columnwidth, height=75px]{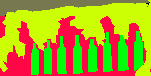}
        \end{subfigure}
        \begin{subfigure}[b]{0.23\columnwidth}
            \centering
            \includegraphics[width=\columnwidth, height=75px]{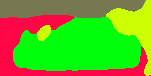}
        \end{subfigure}
        \begin{subfigure}[b]{0.28\columnwidth}
            \centering
            \includegraphics[width=\columnwidth]{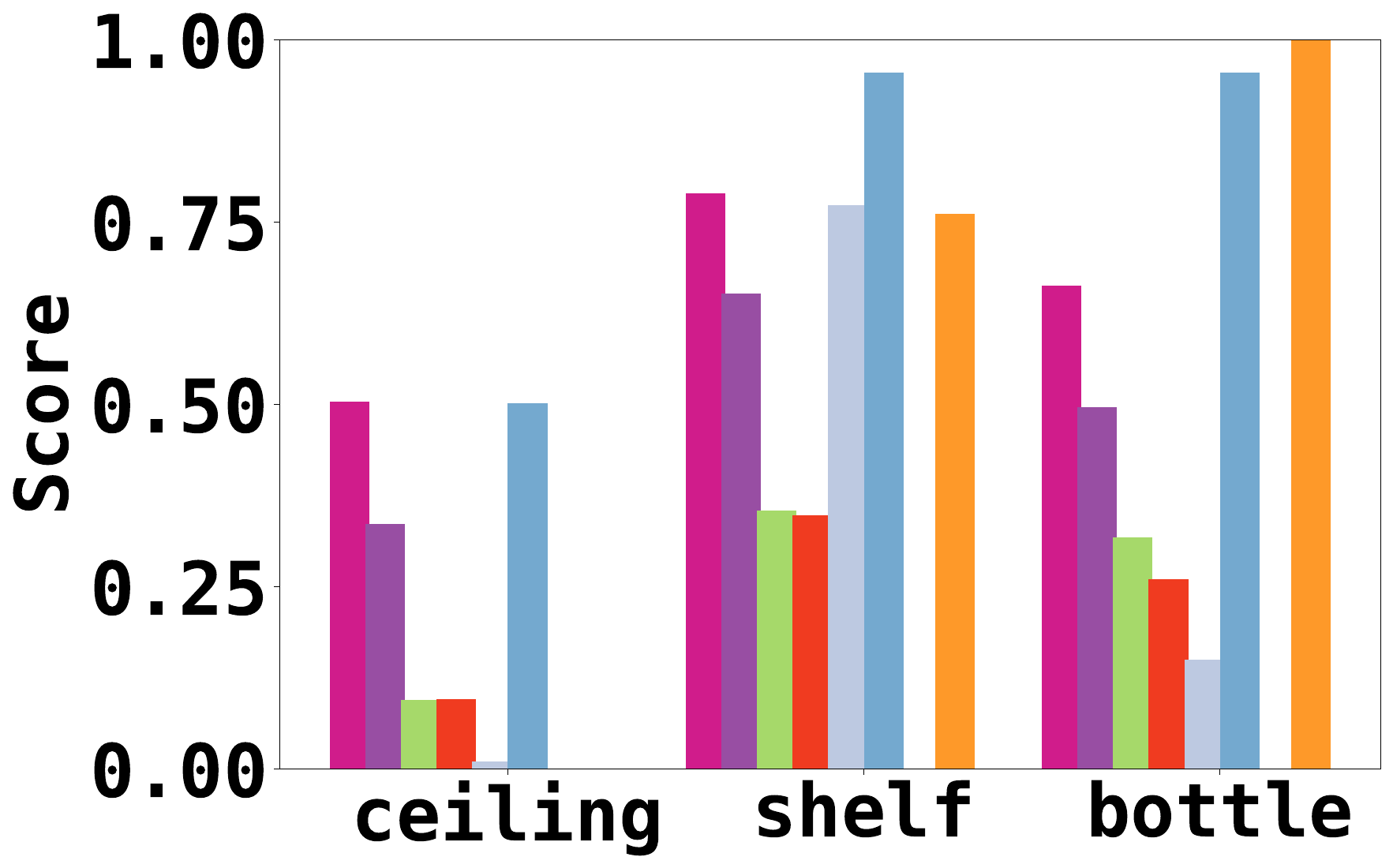}
        \end{subfigure}
        \vspace{2mm}%
    \end{subfigure}

    \begin{subfigure}[b]{2\columnwidth}
        \centering
        \begin{subfigure}[b]{0.23\columnwidth}
            \centering
            \includegraphics[width=\columnwidth]{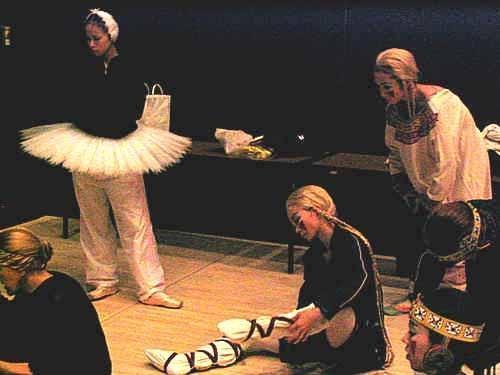}
        \end{subfigure}
        \begin{subfigure}[b]{0.23\columnwidth}
            \centering
            \includegraphics[width=\columnwidth]{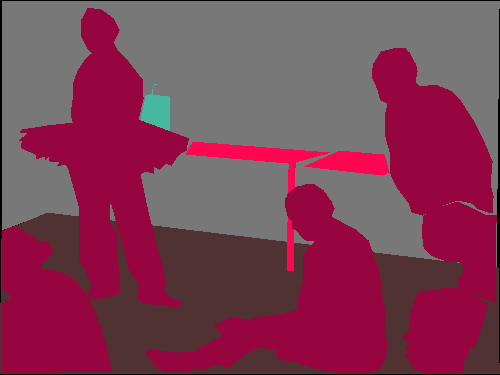}
        \end{subfigure}
        \begin{subfigure}[b]{0.23\columnwidth}
            \centering
            \includegraphics[width=\columnwidth]{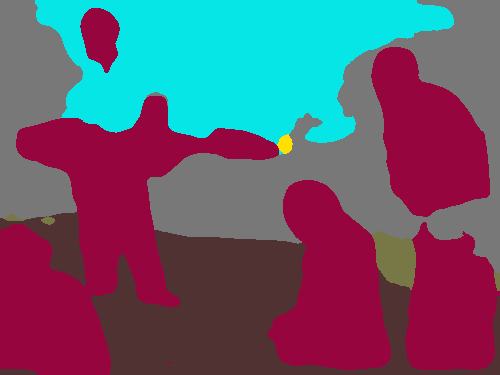}
        \end{subfigure}
        \begin{subfigure}[b]{0.28\columnwidth}
            \centering
            \includegraphics[width=\columnwidth]{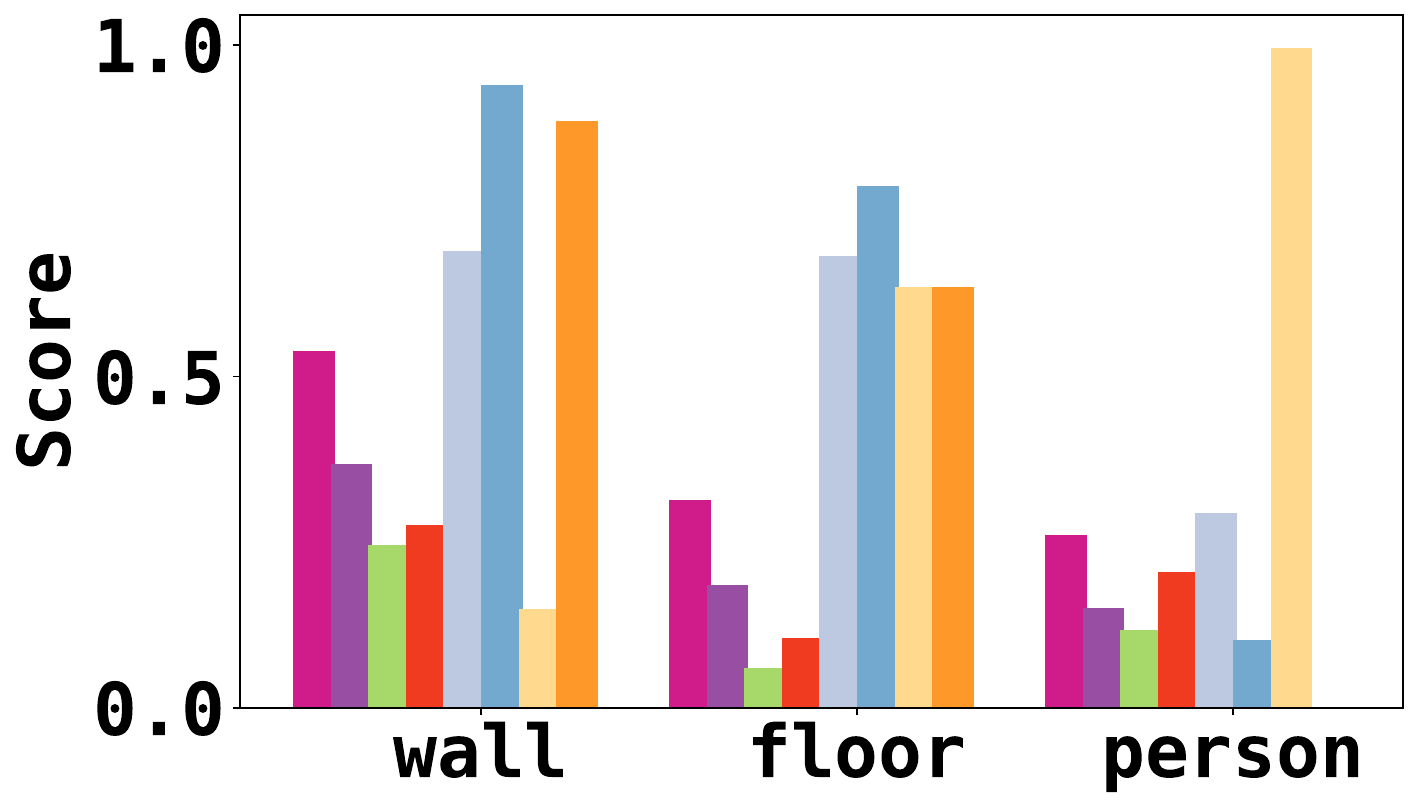}
        \end{subfigure}
        \vspace{2mm}%
    \end{subfigure}
     
    \begin{subfigure}[b]{2\columnwidth}
        \centering
        \begin{subfigure}[b]{0.23\columnwidth}
            \centering
            \includegraphics[width=\columnwidth]{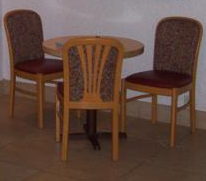}
        \end{subfigure}
        \begin{subfigure}[b]{0.23\columnwidth}
            \centering
            \includegraphics[width=\columnwidth]{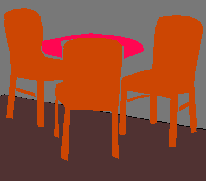}
        \end{subfigure}
        \begin{subfigure}[b]{0.23\columnwidth}
            \centering
            \includegraphics[width=\columnwidth]{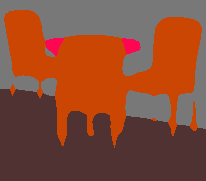}
        \end{subfigure}
        \begin{subfigure}[b]{0.28\columnwidth}
            \centering
            \includegraphics[width=\columnwidth]{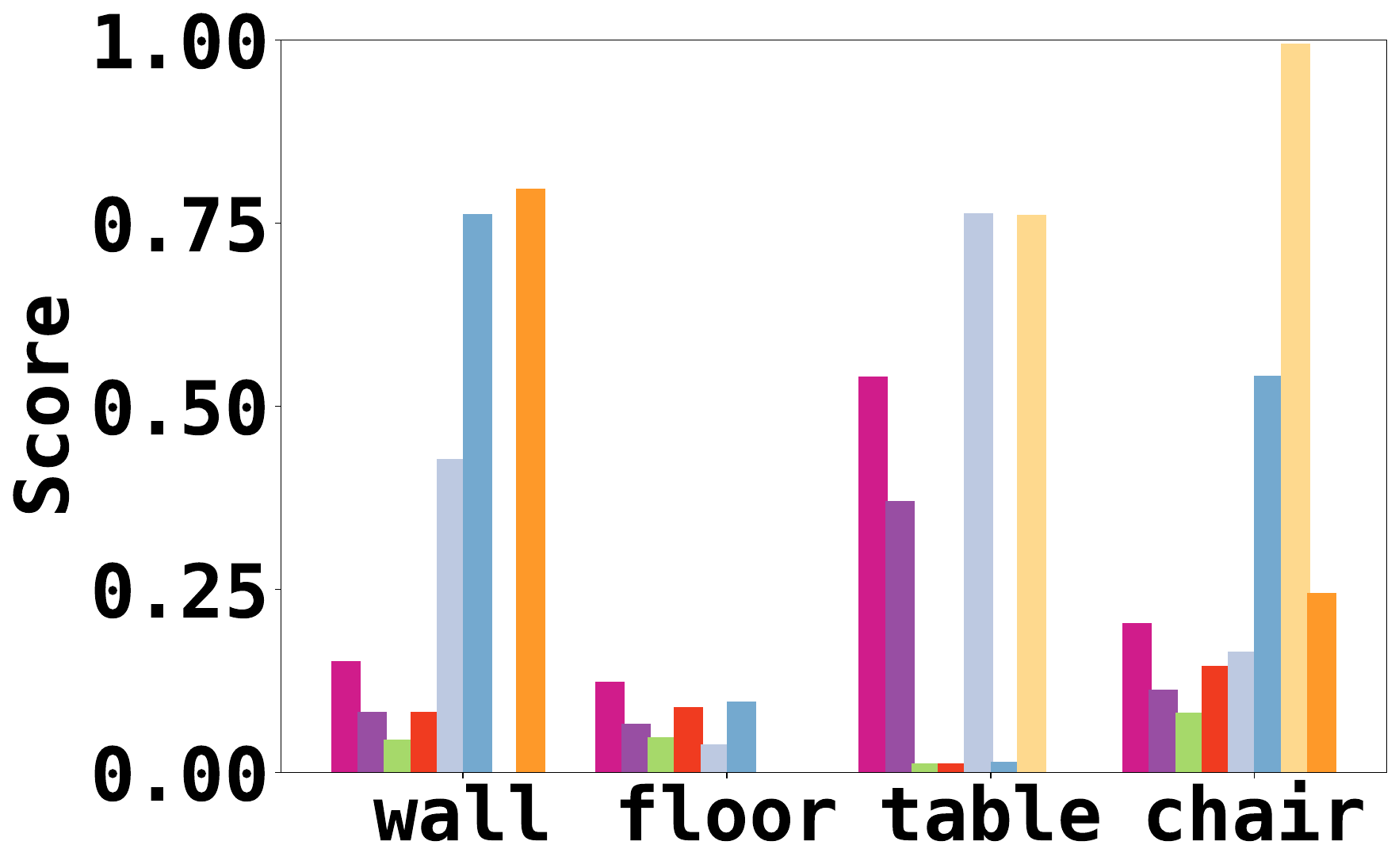}
        \end{subfigure}
        \vspace{2mm}%
    \end{subfigure}
    \begin{subfigure}[b]{2\columnwidth}
        \centering
        \includegraphics[width=0.99\columnwidth]{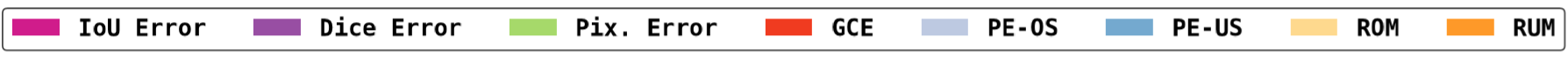}
    \end{subfigure}
    \caption{Qualitative results illustrating over- and under-segmentation. The example in the top row is a near-perfect segmentation. The second row shows an under-segmentation example on the \textit{horse} class. The third row shows an under-segmentation example on the \textit{towel} class. The fourth row shows an under-segmentation example on the \textit{bottle} class. The fifth row shows an over-segmentation example on the \textit{person} class. The last row shows an over-segmentation example on the \textit{chair} class. The error metric legend is shown in the bar graphs (bottom). \textbf{See Supplementary Material for more examples and detailed discussions.}}
    \label{fig:result_color}
\end{figure*}

\begin{table}[b!]
\centering
\caption{Performance Comparison Using Different Metrics. Our Metrics Quantify Over- and Under-segmentation Errors, Thus Explaining The Source of Error in Semantic Segmentation Performance.}
\label{tab:comparison_error_metrics}
    \begin{subtable}[b]{1.1\columnwidth}
        \centering
        \caption{PASCAL VOC 2012}
       \label{tab:comparison_error_metrics_pascal}
        \resizebox{0.8\columnwidth}{!}{
            \begin{tabular}{lccc}
                \toprule[1.5pt]
                \textbf{Error metrics}  & \textbf{DeepLabv3} & \textbf{ESPNetv2} &  \textbf{MobileNetv2}  \\
                \midrule[1.0pt]
                \textbf{mIOU Error} & 0.18 & 0.25 & 0.24 \\
                \textbf{Dice Error} & 0.13& 0.20 & 0.18 \\
                \textbf{Pixel Error} & 0.11 & 0.13 &  0.13 \\
                \textbf{GCE} & 0.07 & 0.09 &  0.10 \\
                \textbf{PE-OS} & 0.31 &  0.37 &  0.35 \\
                \textbf{PE-US} & 0.41 &  0.45 & 0.44 \\
                \textbf{AP Error} & 0.67 &  0.70 & 0.70 \\
                \textbf{ROM} (Ours) & 0.07 &  0.08 &  0.11 \\
                \textbf{RUM} (Ours) & 0.32 &  0.30 &  0.30 \\
                \bottomrule[1.5pt]
            \end{tabular}
        }
        \vspace{2mm}
    \end{subtable}
    \vfill
    \begin{subtable}[b]{0.8\columnwidth}
        \centering
        \caption{Cityscapes}
        \label{tab:comparison_error_metrics_city}
        \resizebox{0.85\columnwidth}{!}{
            \begin{tabular}{lcccc}
                \toprule[1.5pt]
                \textbf{Error metrics} &  & \textbf{DRN} & & \textbf{ESPNetv2}\\
                \midrule[1.0pt]
                \textbf{mIOU Error} &  & 0.30 &  & 0.38\\
                \textbf{Dice Error} &  & 0.22 & & 0.30\\
                \textbf{Pixel Error} &  & 0.17 &  & 0.18 \\
                \textbf{GCE} &  & 0.13 &  & 0.16\\
                \textbf{PE-OS} &  & 0.46 & &  0.55\\
                \textbf{PE-US} &  & 0.50 & &  0.62\\
                \textbf{AP Error} &  & 0.81 & &  0.86\\
                \textbf{ROM} (Ours) &  & 0.26 &&  0.25\\
                \textbf{RUM} (Ours) &  & 0.35 &&  0.40\\
                \bottomrule[1.5pt]
            \end{tabular}
        }
        \vspace{2mm}
    \end{subtable}
    \vfill
    \begin{subtable}[b]{\columnwidth}
        \centering
        \caption{ADE20K}
        \label{tab:comparison_error_metrics_ade20k}
        \resizebox{0.8\columnwidth}{!}{
            \begin{tabular}{lccc}
                \toprule[1.5pt]
                \multicolumn{1}{l}{\multirow{2}{*}{\textbf{Error metrics}}} & \multicolumn{2}{c}{\textbf{PSPNet}} &  \multirow{1}{*}{\textbf{MobileNetv2}} \\
                \cmidrule{2-3}
                & \textbf{ResNet-18} & \textbf{ResNet-101} &\\
                \midrule[1.0pt]
                 \textbf{mIOU Error} & 0.62 &  0.58 &  0.65 \\
                \textbf{Dice Error}& 0.43 &  0.40 &  0.46 \\
                \textbf{Pixel Error}& 0.21 &  0.19 &  0.24 \\
                \textbf{GCE} & 0.19 &  0.17 &  0.21 \\
                \textbf{PE-OS} & 0.53 & 0.50 &  0.56 \\
                \textbf{PE-US} & 0.52 &  0.49 &  0.54 \\
                \textbf{AP Error} & 0.73 &  0.69 &  0.76 \\
                \textbf{ROM} (Ours) &  0.11 &  0.10 &  0.16 \\
                \textbf{RUM} (Ours) &  0.23 &  0.23 &  0.22 \\
                \bottomrule[1.5pt]
            \end{tabular}
        }
    \end{subtable}
\end{table}

\vspace{1mm}
\noindent \textbf{Semantic segmentation models and datasets.} We study our metric using state-of-the-art semantic segmentation models on three standard datasets (PASCAL VOC 2012 \cite{everingham2010pascal}, ADE20K \cite{zhou2019semantic}, and Cityscapes \cite{cordts2016cityscapes}). These models were selected based on network design decisions (e.g., light- vs. heavy-weight), public availability, and segmentation performance (IOU), while the datasets were selected based on different visual recognition tasks (PASCAL VOC 2012: foreground-background segmentation, ADE20K: scene parsing, and Cityscapes: urban scene understanding). In this paper, we focus on semantic segmentation datasets because of their  wide applicability across different domains, including medical imaging (e.g., \cite{menze2014multimodal}). However, our metric is generic and can be easily applied to other segmentation tasks (e.g., panoptic segmentation) to assess over- and under-segmentation performance.

For the PASCAL VOC 2012, we chose Deeplabv3 \cite{chen2017rethinking} as our heavy-weight model and two recent efficient networks (ESPNetv2 \cite{mehta2019espnetv2} and MobileNetv2 \cite{sandler2018mobilenetv2}) as light-weight models. For ADE20K, we chose PSPNet \cite{zhao2017pyramid} (with ResNet-18 and ResNet-101 \cite{he2016deep}) as heavy-weight networks and MobileNetv2 \cite{sandler2018mobilenetv2} as the light-weight network. For Cityscapes, we chose DRN \cite{yu2017dilated} and ESPNetv2 \cite{mehta2019espnetv2} as heavy- and light-weight networks, respectively.

\section{Results and Discussion}
\label{sec:experiments}

We describe the utility of our approach through extensive use of examples. Figure~\ref{fig:main_results}, Figure \ref{fig:result_color}, and Table \ref{tab:comparison_error_metrics} show results for different datasets and segmentation models. Qualitatively, our metric can quantify over- and under-segmentation issues (Figure \ref{fig:result_color}). For instance, the example in the first row of Figure~\ref{fig:result_color} shows a three-class task. The model prediction was almost perfect for this example, i.e., it correctly segmented each object. Our metrics correctly assigned $ROM=0$ and $RUM=0$, meaning there was no over- or under-segmentation in this example. However, all other metrics indicated that non-negligible errors occurred in this example. The fifth row of Figure~\ref{fig:result_color} shows an example of over-segmentation for the \textit{person} class. The torso of the leftmost person was over-segmented into two regions that were far apart, delivering a false prediction that two people were at the same location. A high value for $ROM$ reflected this discrepancy. The third row of Figure~\ref{fig:result_color} shows under-segmentation for the towel class, where accurate pixel classification and proper region segmentation disagreed. Pixel-wise measures (IOU, Dice score, and pixel error) indicated little error for the \textit{towel} class. However, the number of model-predicted regions were only half the number of ground-truth regions. A high value for $RUM$ reflected this disagreement.

Quantitatively, all models performed well on the PASCAL VOC dataset, as reflected by the low error scores in Table~\ref{tab:comparison_error_metrics_pascal}. The light-weight models ESPNetv2 and MobileNetv2 had similar mIOU errors, but ROM shows that ESPNetv2 made fewer mistakes in terms of over-segmentation. The Receiver operating characteristic (ROC) curves in Figure~\ref{fig:roc_pascal} indicate that all models made fewer mistakes when region confidence thresholds increased. At the threshold of 0.6, MobileNetv2 showed a reduction in over-segmentation error (achieving similarity to the other two models) without worsening under-segmentation error.

In the context of quotidian goals of weighing trade-offs between heavy- and light-weight models for particular segmentation tasks, we can make several observations about $ROM$/$RUM$. 
Table \ref{tab:comparison_error_metrics_city} compares overall performance of different models on the Cityscapes validation set. Comparing the heavy-weight (DRN) and light-weight (ESPNetv2) models, we expected to see ESPNetv2 generally perform worse, which is reflected by an 8 point difference in IOU, Dice score, and pixel accuracy. In addition to these pixel-wise measures, our metrics help explain the source of performance discrepancy. DRN and ESPNetv2 had the same value for $ROM$ but a significant difference of 5 points in $RUM$. This indicates that performance degradation in ESPNetv2 in this dataset mainly emanated from under-, not over-, segmentation. Similarly, in Table \ref{tab:comparison_error_metrics_ade20k}, we can explain that the lower performance of MobileNetv2 (a light-weight model) compared to PSPNet with ResNet-101 (a heavy-weight model) on the ADE20K dataset is primarily due to over-segmentation issues. A researcher may elect to tolerate the increase in under- or over-segmentation in exchange for computational efficiency, depending on downstream uses. Conversely, for navigation tasks, under-segmenting light poles may undercut downstream wayfinding and mapping uses, making it an undesirable outcome~\cite{zhang2019stereo}.


\section{Conclusion}
This work reviewed measures of similarity popular in computer vision for assessing semantic segmentation fidelity to a ground-truth and discussed their shortcomings in accounting for over- and under-segmentation. While IOU and similar evaluation metrics can be applied to measure pixel-wise correspondence between both model-predicted segmentation and ground-truth, the evaluation can suffer from inconsistencies due to different region boundaries and notions of significance with respect to a particular class of objects. We proposed an approach to segmentation metrics that decomposes to explainable terms and is sensitive to over- and under-segmentation errors. This new approach confers additional desirable properties, like robustness to boundary changes (since both annotations and natural-image semantic boundaries are non-deterministic). We contextualized the application of our metrics in current model selection problems that arise in practice when attempting to match the context of use to region-based segmentation performance in supervised datasets.

\section*{Acknowledgment}

This work was sponsored by the State of Washington Department of  Transportation (WSDOT) research grant T1461-47, in cooperation with
the Taskar Center for Accessible Technology at the Paul G. Allen School at the University of Washington.

\bibliographystyle{IEEEtran}
\bibliography{arXiv}

\end{document}


\title{Supplementary Material: \\Rethinking Semantic Segmentation Evaluation for Explainability and Model Selection}

\author{Yuxiang~Zhang,
        Sachin~Mehta,
        and~Anat~Caspi 
\thanks{Y. Zhang, S. Mehta, and A. Caspi are with the University of Washington, Seattle,
 WA, 98115 USA. E-mail: \{yz325, sacmehta, caspian\}@cs.washington.edu.}
}

\maketitle

\section{Outline}
\label{sec:outline}
This document accompanies the paper “Rethinking Semantic Segmentation Evaluation for Explainability and Model Selection”  We illustrate our evaluation analysis of the ROM/RUM metrics with results using additional image examples.
 
In section~\ref{sec:examples}, we provide additional examples for comparing different metrics in various scenarios. Extended discussions based on the additional examples are made in section~\ref{sec:extend_discussion}.

\section{Conjoint use of RUM/ROM with mIOU to inform model selection and evaluation}
\label{sec:examples}

Our paper proposes that region-based error metrics are not necessarily correlated with pixel-wise error metrics. As alternative sources of information about the model's performance in segmentation settings, and in particular, specific class performance, it is important to understand how pixel-based and region-based metrics might interact to better inform model selection or focus on improvements in any particular class.

\begin{figure*}[tbh!]
\centering
\includegraphics[width=2\columnwidth]{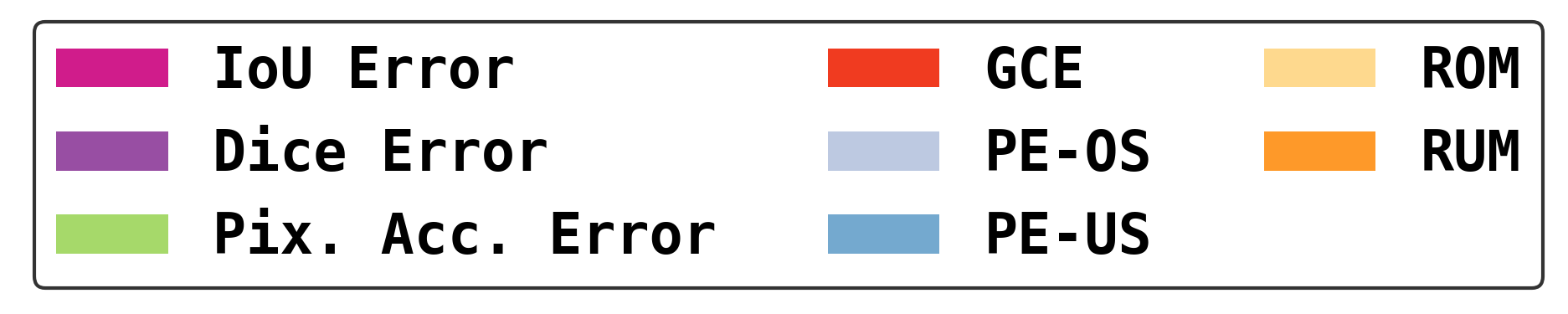}
\caption{Legend for different metrics}
\label{fig:legend}
\end{figure*}

In this section, we demonstrate our suggested metrics are orthogonal to traditional segmentation metrics and the manner in which the metrics can be used conjointly to inform model selection and evaluation. We integrate additional segmentation examples, with their class-wise score and RUM/ROM score. The examples are drawn from the ADE20K dataset and the Cityscapes dataset. The ADE20K dataset provides a variety of indoor/outdoor scenes with 150 object classes. The wide range of scenes in this dataset allows us to locate examples for different segmentation scenarios. The Cityscapes dataset contains urban scenes captured from a car dashboard's perspective, with 19 object classes commonly seen in the urban road environment. This dataset is important to tasks involving autonomous vehicles and outdoor environment mapping. The examples drawn from these two datasets are assorted into different categories as follows. The color code representing different metrics is shown in Figure~\ref{fig:legend}.

\subsection{Detailed Interrogation Into Low Pixel-wise Error and Region-based Error Metrics}
\label{sec:Low Pixel-wise Error}

Pixel-wise measures (IOU, dice score, pixel accuracy) are indeed informative when addressing questions surrounding pixel classification. However correlated these measures are, they are not surrogate metrics for near-perfect semantic segmentation. Importantly, pixel-wise measurements do not capture any region-wise information, nor indicating if there are any over-/under-segmentation issues in the prediction.  The following examples show that despite similar pixel-wise error metrics, the quality of semantic segmentation in the prediction will diverge.

\begin{figure*}[tbh!]
\begin{subfigure}[b]{2\columnwidth}
    \centering
    \begin{tabular}{>{\centering\arraybackslash}p{0.22\columnwidth}>{\centering\arraybackslash}p{0.22\columnwidth}>{\centering\arraybackslash}p{0.22\columnwidth}>{\centering\arraybackslash}p{0.22\columnwidth}}
        \toprule
        \textbf{RGB Image} & \textbf{Ground-Truth} & \textbf{Prediction}& \textbf{Error metrics}\\
        \midrule
        \end{tabular}        
    \end{subfigure}
        
     \begin{subfigure}[b]{2\columnwidth}
         \centering
        \includegraphics[width=0.24\columnwidth]{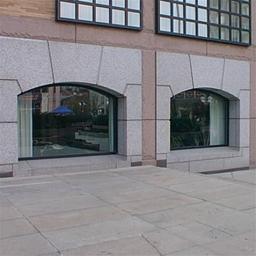}  
        \includegraphics[width=0.24\columnwidth]{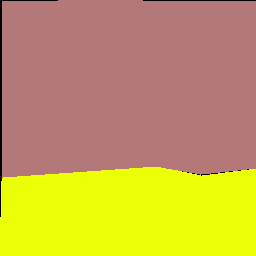} 
        \includegraphics[width=0.24\columnwidth]{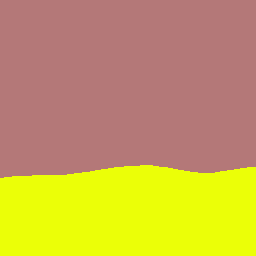} 
        \includegraphics[width=0.24\columnwidth]{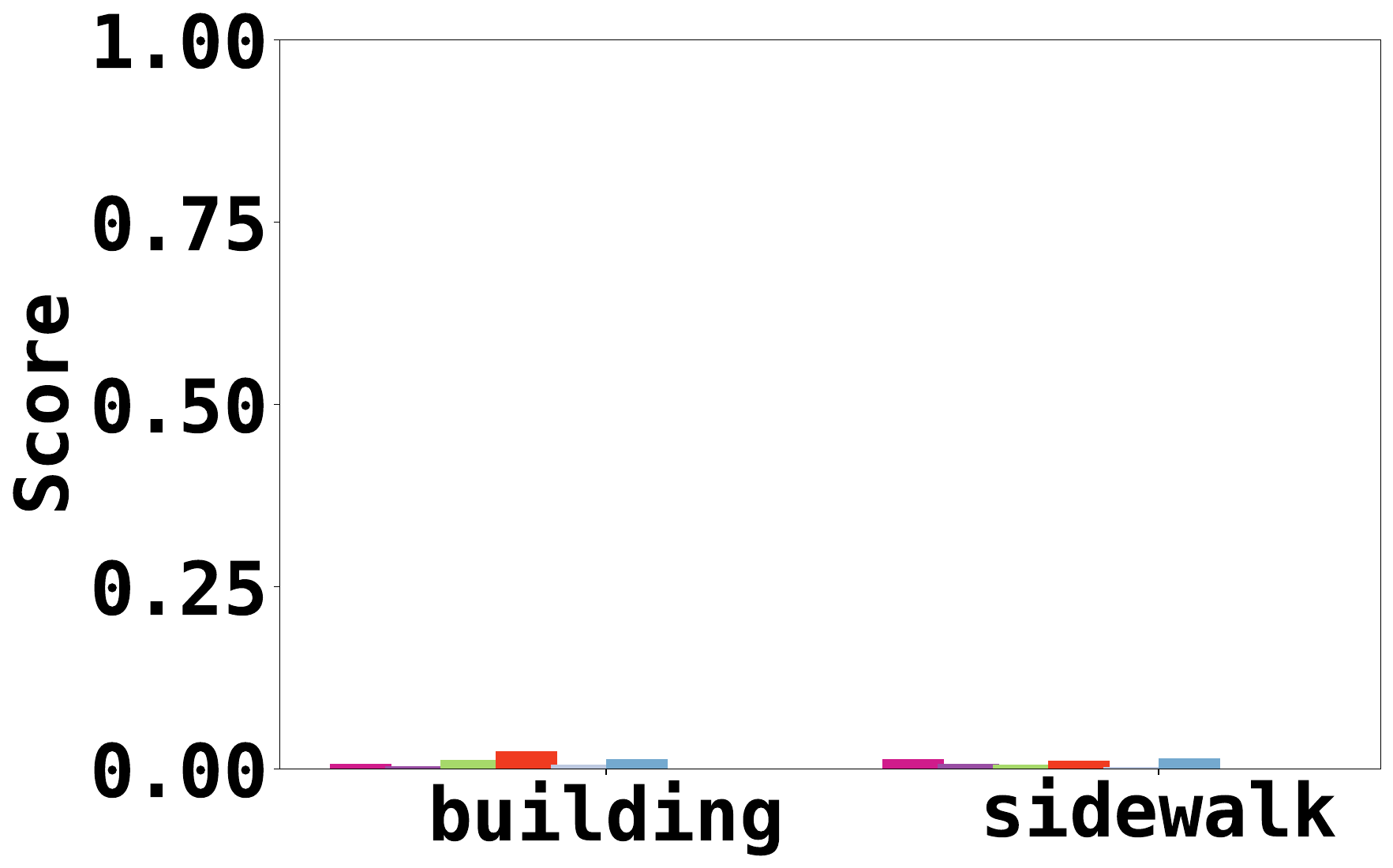} \\
    \end{subfigure}
    \caption{Low IOU Error, Low ROM/RUM}
    \label{fig:low iou low rom low rum}

\end{figure*}

\noindent \textbf{Low ROM, Low RUM}: Exploring the category of joint low pixel-wise errors, low over-segmentation error (ROM), and low under-segmentation error (RUM), we anticipate high model-prediction fidelity to the ground truth. Looking at these metrics conjointly, we expect that the model predicts a near-perfect segmentation. As an example, in Figure~\ref{fig:low iou low rom low rum}, an excellent prediction on a simple 2-class image with low pixel-wise error metrics, and no error along with ROM/RUM. Without the need to inspect the results by eye, we can expect few or no region-wise issues in this example and others like it.\\

\begin{figure*}[tbh!]

\begin{subfigure}[b]{2\columnwidth}

    \centering
    \begin{tabular}{>{\centering\arraybackslash}p{0.22\columnwidth}>{\centering\arraybackslash}p{0.22\columnwidth}>{\centering\arraybackslash}p{0.22\columnwidth}>{\centering\arraybackslash}p{0.22\columnwidth}}
        \toprule
        \textbf{RGB Image} & \textbf{Ground-Truth} & \textbf{Prediction}& \textbf{Error metrics}\\
        \midrule
        \end{tabular}    
        \end{subfigure}
        
     \begin{subfigure}[b]{2\columnwidth}
        \includegraphics[width=0.24\columnwidth]{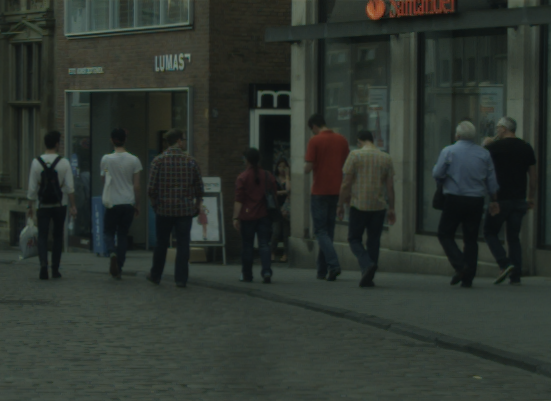}  
        \includegraphics[width=0.24\columnwidth]{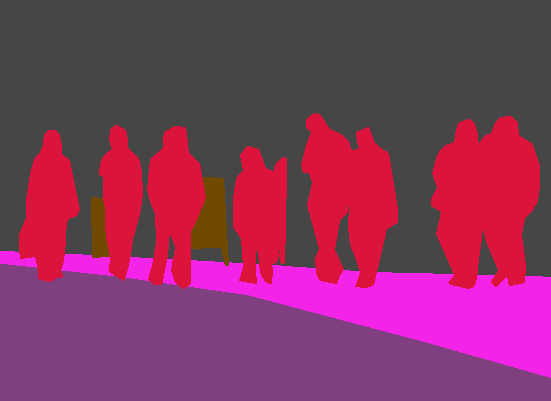} 
        \includegraphics[width=0.24\columnwidth]{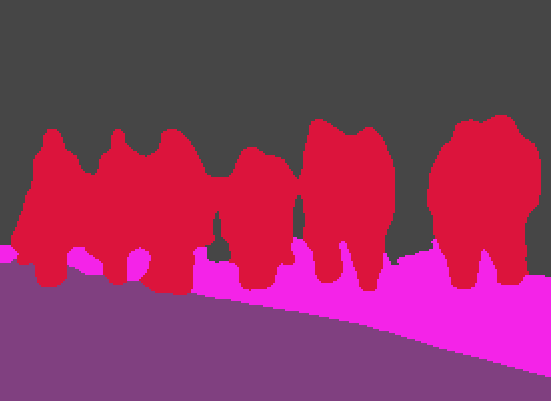} 
        \includegraphics[width=0.24\columnwidth]{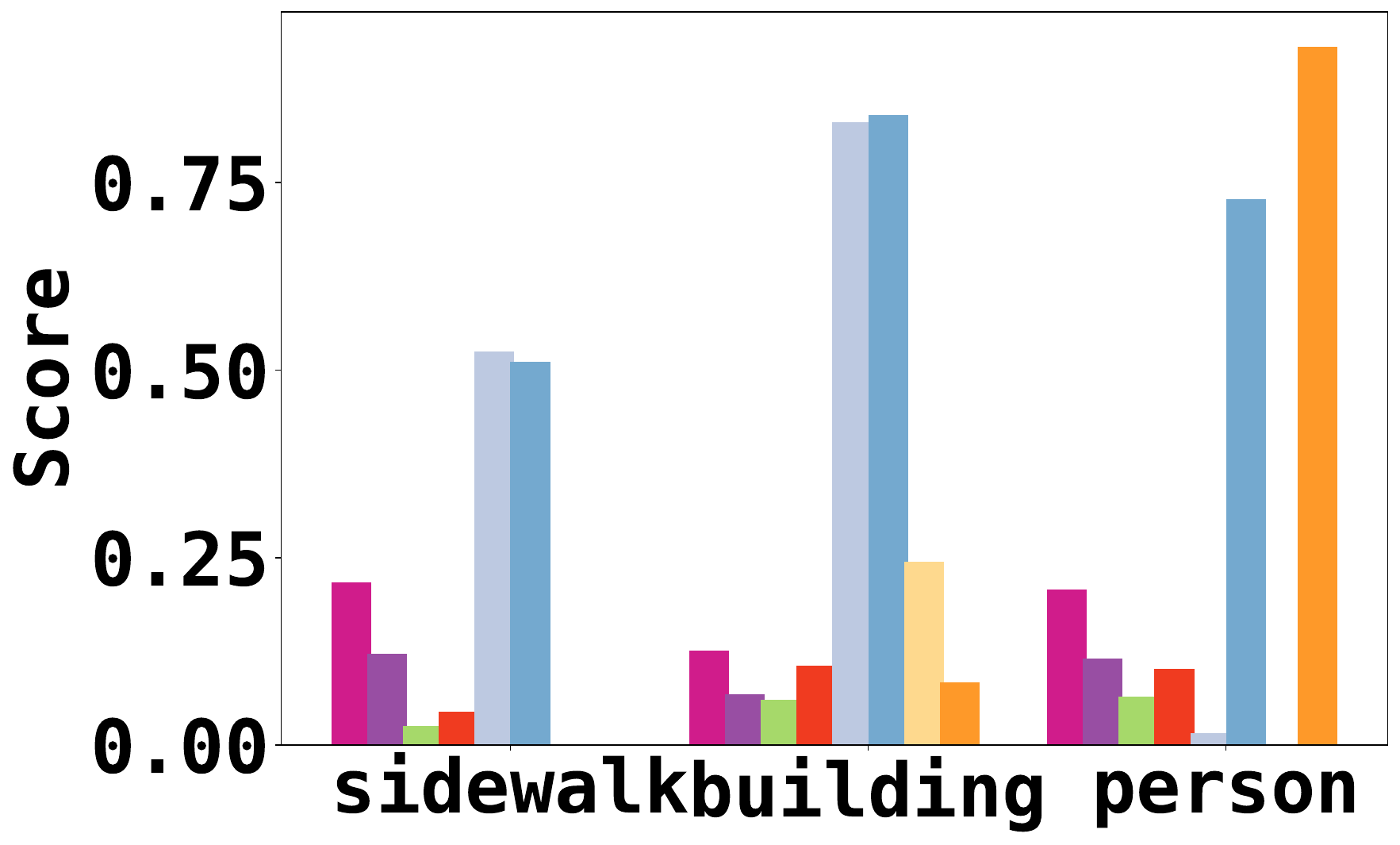} \\
    \end{subfigure}

    \caption{Low IOU Error, High RUM (person)}
    \label{fig:low iou high rum}
\end{figure*}

\noindent \textbf{High RUM}: Demonstrating that pixel-wise and region-wise metrics are not always correlated, it is important to note cases in which a result yields low pixel-wise errors, but high RUM, indicating that the number of regions predicted for a particular class by the model is less than the number of regions in the ground truth. As an example, in Figure~\ref{fig:low iou high rum}, the model only predicts 2 regions of people out of 6 in the ground truth. The pixel distance between regions classified as people is small, and therefore region under-segmentation does not significantly impact pixel-wise error. Nevertheless, but the region-wise discrepancy is reflected by the high RUM. By far, this is the most prevalent type of conjoint error we see in the Cityscapes dataset with the ESPNetv2 model predictions.\\

\begin{figure*}[tbh!]
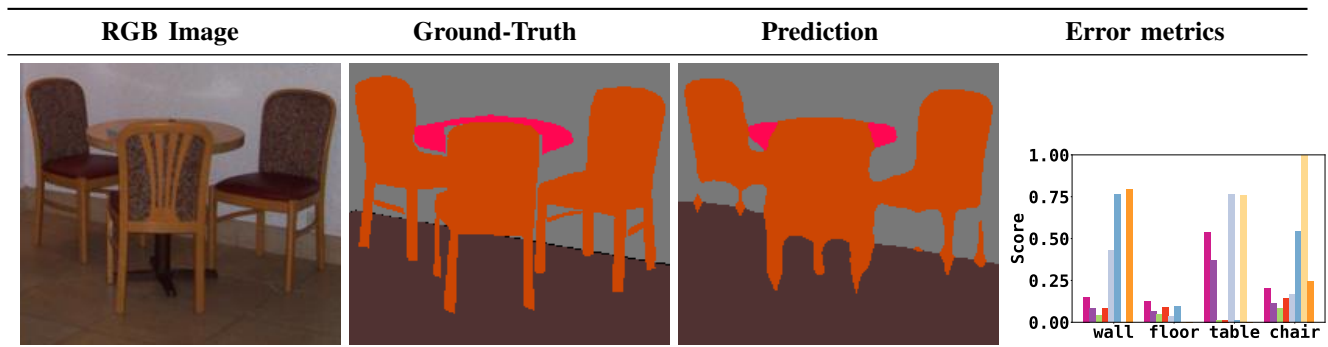

    \centering    
    \begin{subfigure}[b]{2\columnwidth}
    \begin{tabular}{>{\centering\arraybackslash}p{0.22\columnwidth}>{\centering\arraybackslash}p{0.22\columnwidth}>{\centering\arraybackslash}p{0.22\columnwidth}>{\centering\arraybackslash}p{0.22\columnwidth}}
        \toprule
        \textbf{RGB Image} & \textbf{Ground-Truth} & \textbf{Prediction}& \textbf{Error metrics}\\
        \midrule
        \end{tabular}  
    \end{subfigure}
        
    \begin{subfigure}[b]{2\columnwidth}
        \centering
        \includegraphics[width=0.24\columnwidth]{supplementary/ade_examples/ADE_val_00000429_rgb.png}  \includegraphics[width=0.24\columnwidth]{supplementary/ade_examples/ADE_val_00000429_gt} 
        \includegraphics[width=0.24\columnwidth]{supplementary/ade_examples/ADE_val_00000429_color.png} 
        \includegraphics[width=0.24\columnwidth]{supplementary/ade_examples/ADE_val_00000429.pdf} \\
    \end{subfigure}

    \caption{Low IOU Error, High ROM (chair)}
    \label{fig:low iou high rom}
\end{figure*}

\noindent \textbf{High ROM}: High ROM in conjunction with low pixel-wise errors may also occur, indicative of model-predictions that demonstrate low per-pixel classification for an object class, despite segmenting more than one distinct regions per single class object in the ground truth. For example in Figure~\ref{fig:low iou high rom}, the chair legs are segmented into multiple separate regions, creating over-segmentation issues and delivering false inference that there are multiple chairs at the same location. Note that this is an interesting example of the interplay between different class segmentation: when one chair is over-segmented in multiple disconnected regions, the result impacts another class' region segmentation, i.e., the chair background is then connected into one large region for the wall class. The wall is then under-segmented as a result of the chair over-segmentation. This demonstrates how models might be interrogated for intricate nuanced relationships among classes. In this case, looking at the entire model (ResNet101+PSPNet) performance on this dataset (ADE20k), overall mIOU error is 0.58, whereas for the chair class (mIOU error is 0.54, RUM is 0.09, and ROM is 0.14 ), interplayed with wall class (mIOU error is 0.32, RUM is 0.31, and ROM is 0.10). \\

\subsection{Detailed Interrogation Into High Pixel-wise Error and Region-based Error Metrics}
\label{sec:High Pixel-wise Error}
There is no question that significant pixel-wise classification errors cannot yield very good segmentation. However, the examples we looked at are encouraging in terms of the potential uses of light-weight models (with occasionally higher pixel-wise error) in certain semantic segmentation scenarios. 

We looked at images that tended to have high pixel-wise errors in conjunction with ROM/RUM. Identifying images with specific region-wise metric attributes can provide insight on region-wise segmentation qualities and offer clues about the types of images or scenes in which pixel-wise predictions may fail. \\

\begin{figure*}[tbh!]
    \centering
     \begin{subfigure}[b]{2\columnwidth}
    \begin{tabular}{>{\centering\arraybackslash}p{0.22\columnwidth}>{\centering\arraybackslash}p{0.22\columnwidth}>{\centering\arraybackslash}p{0.22\columnwidth}>{\centering\arraybackslash}p{0.22\columnwidth}}
        \toprule
        \textbf{RGB Image} & \textbf{Ground-Truth} & \textbf{Prediction}& \textbf{Error metrics}\\
        \midrule
        \end{tabular}    
    \end{subfigure}
        
     \begin{subfigure}[b]{2\columnwidth}
         \centering
        \includegraphics[width=0.24\columnwidth]{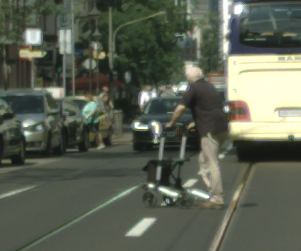}  
        \includegraphics[width=0.24\columnwidth]{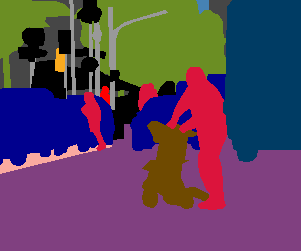} 
        \includegraphics[width=0.24\columnwidth]{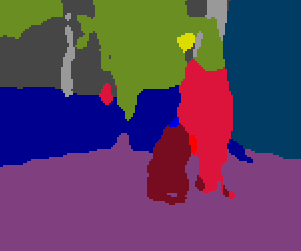} 
        \includegraphics[width=0.24\columnwidth]{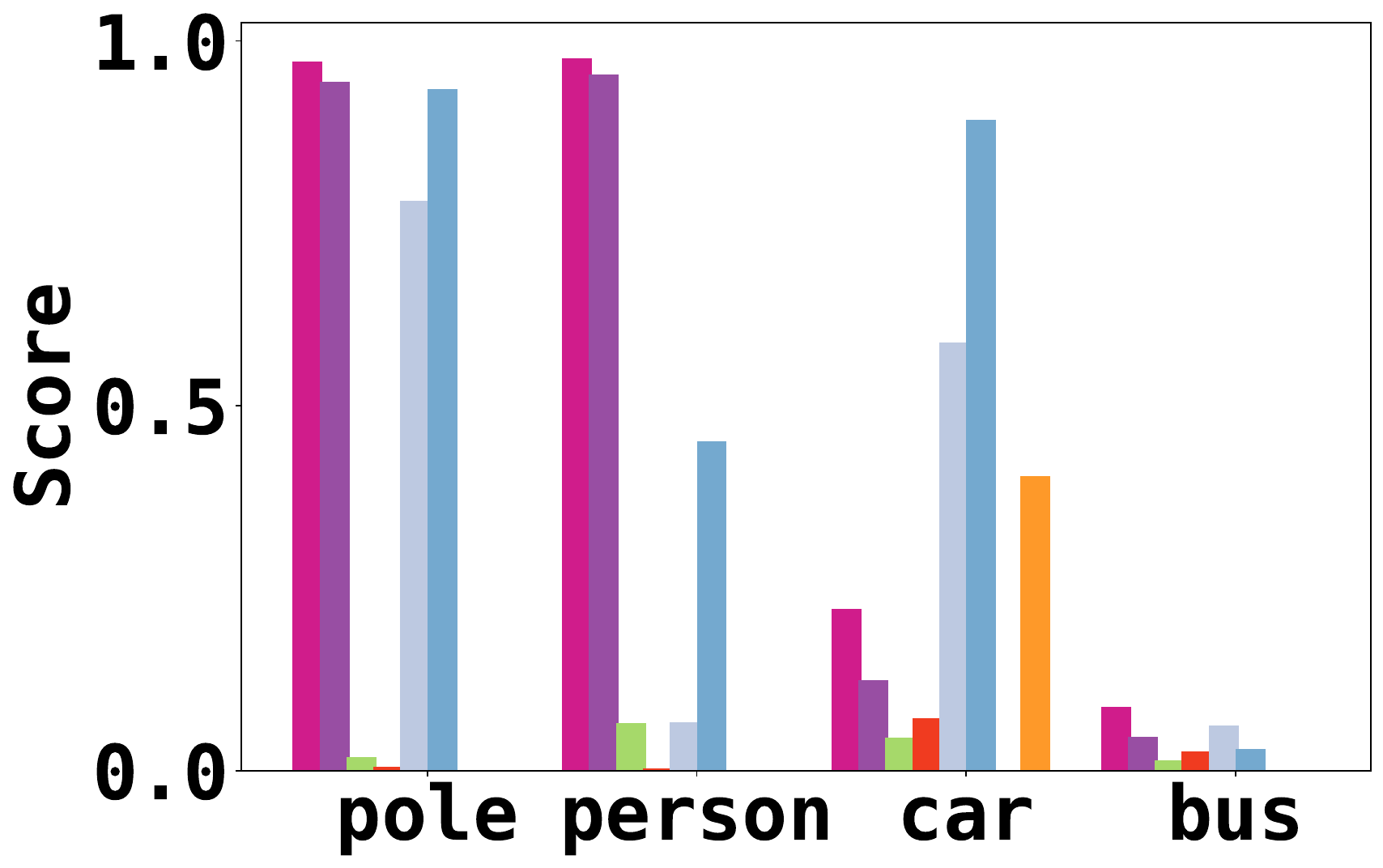} \\
    \end{subfigure}
    \caption{High IOU Error, Low RUM/ROM (person, pole)}
    \label{fig:high iou low rom low rum}

\end{figure*}

\noindent \textbf{Low ROM, Low RUM}: In this class of images, the predicted pixel classification may create false-negative/false-positive regions that degrade pixel-wise accuracy. However, those classes yielding regions that are correctly predicted by the model correspond well with ground-truth, and reflected by the low ROM/RUM metrics. For example, in Figure~\ref{fig:high iou low rom low rum}, although the model fails to predict some persons and poles, and consequently makes few false-positive predictions, those regions that are predicted correctly are matched to the corresponding objects in the ground truth. As no over-/under-segmentation occurred in those particular predicted regions, they are assigned a 0 value ROM/RUM. Again, this notion can be used to interrogate model predictions in a more nuanced manner to understand how the interplay between different classes (with potentially different backgrounds) may account for gross pixel-wise classification errors with certain models. \\

\begin{figure*}[tbh!]
    \centering
    \begin{subfigure}[b]{2\columnwidth}
    \begin{tabular}{>{\centering\arraybackslash}p{0.22\columnwidth}>{\centering\arraybackslash}p{0.22\columnwidth}>{\centering\arraybackslash}p{0.22\columnwidth}>{\centering\arraybackslash}p{0.22\columnwidth}}
        \toprule
        \textbf{RGB Image} & \textbf{Ground-Truth} & \textbf{Prediction}& \textbf{Error metrics}\\
        \midrule
        \end{tabular}    
    \end{subfigure}
        
        \begin{subfigure}[b]{2\columnwidth}
        \includegraphics[width=0.24\columnwidth]{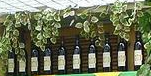} 
        \includegraphics[width=0.24\columnwidth]{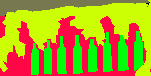} 
        \includegraphics[width=0.24\columnwidth]{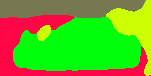} 
        \includegraphics[width=0.24\columnwidth]{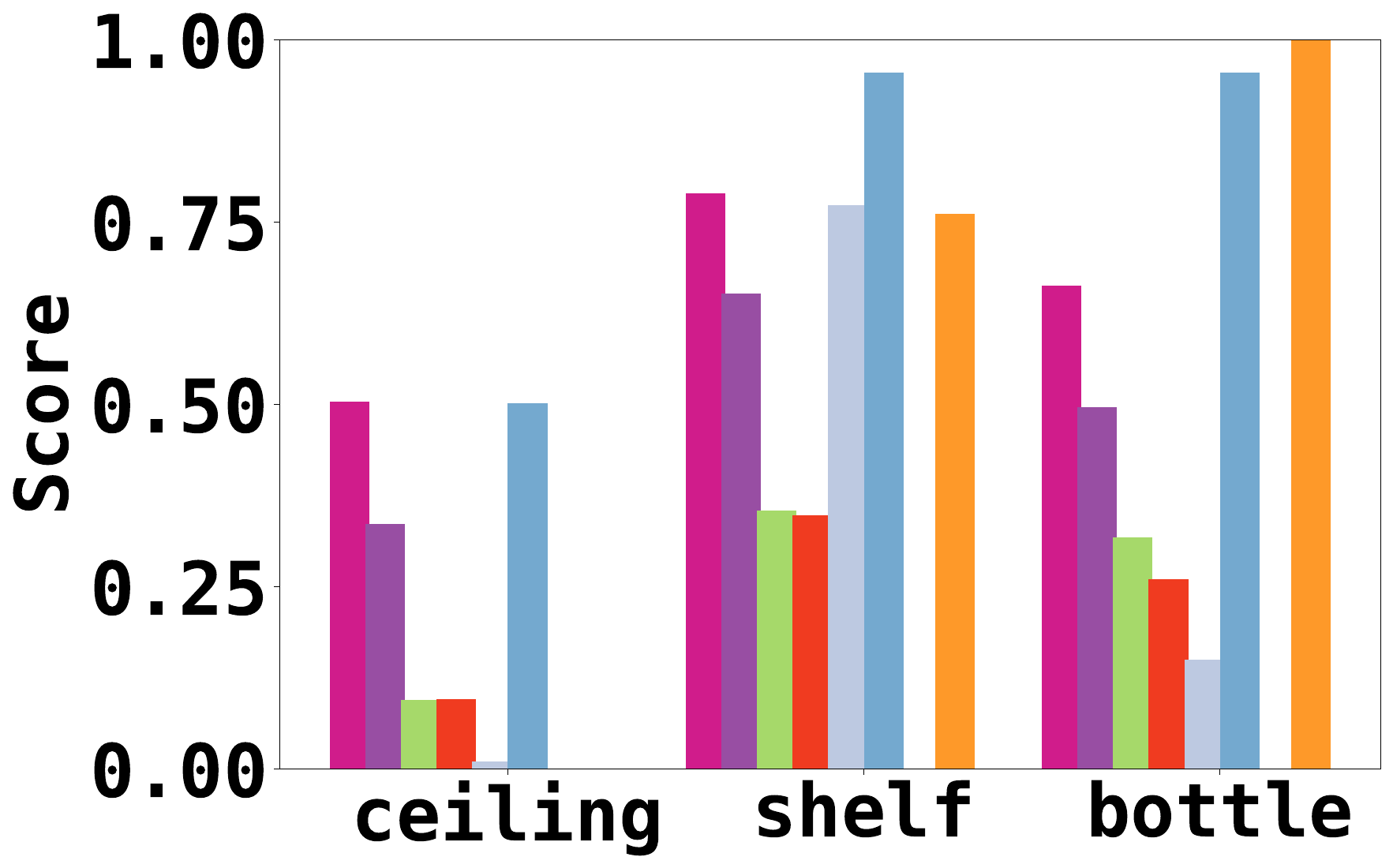} \\
    \end{subfigure}
        \caption{High IOU Error, High RUM (bottle)}
        \label{fig:high iou high rum}
\end{figure*}

\noindent \textbf{High RUM}: In this category, the pixel-wise accuracy errors are high, and a high RUM value indicates that these errors mainly come from under-segmentation. As shown in Figure~\ref{fig:high iou high rum}, the bottles on the shelf are falsely grouped into one large object. Meanwhile, the pixels in between each bottle are falsely predicted into the bottle class which impairs the overall pixel-wise accuracy. \\

\begin{figure*}[tbhp!]
    \centering
    \begin{subfigure}[b]{2\columnwidth}
    \begin{tabular}{>{\centering\arraybackslash}p{0.22\columnwidth}>{\centering\arraybackslash}p{0.22\columnwidth}>{\centering\arraybackslash}p{0.22\columnwidth}>{\centering\arraybackslash}p{0.22\columnwidth}}
        \toprule
        \textbf{RGB Image} & \textbf{Ground-Truth} & \textbf{Prediction}& \textbf{Error metrics}\\
        \midrule
        \end{tabular}    
        \end{subfigure}
        
    \begin{subfigure}[b]{2\columnwidth}
        \centering
        \includegraphics[width=0.24\columnwidth]{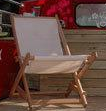}  
        \includegraphics[width=0.24\columnwidth]{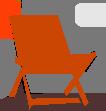} 
        \includegraphics[width=0.24\columnwidth]{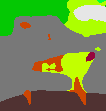} 
        \includegraphics[width=0.24\columnwidth]{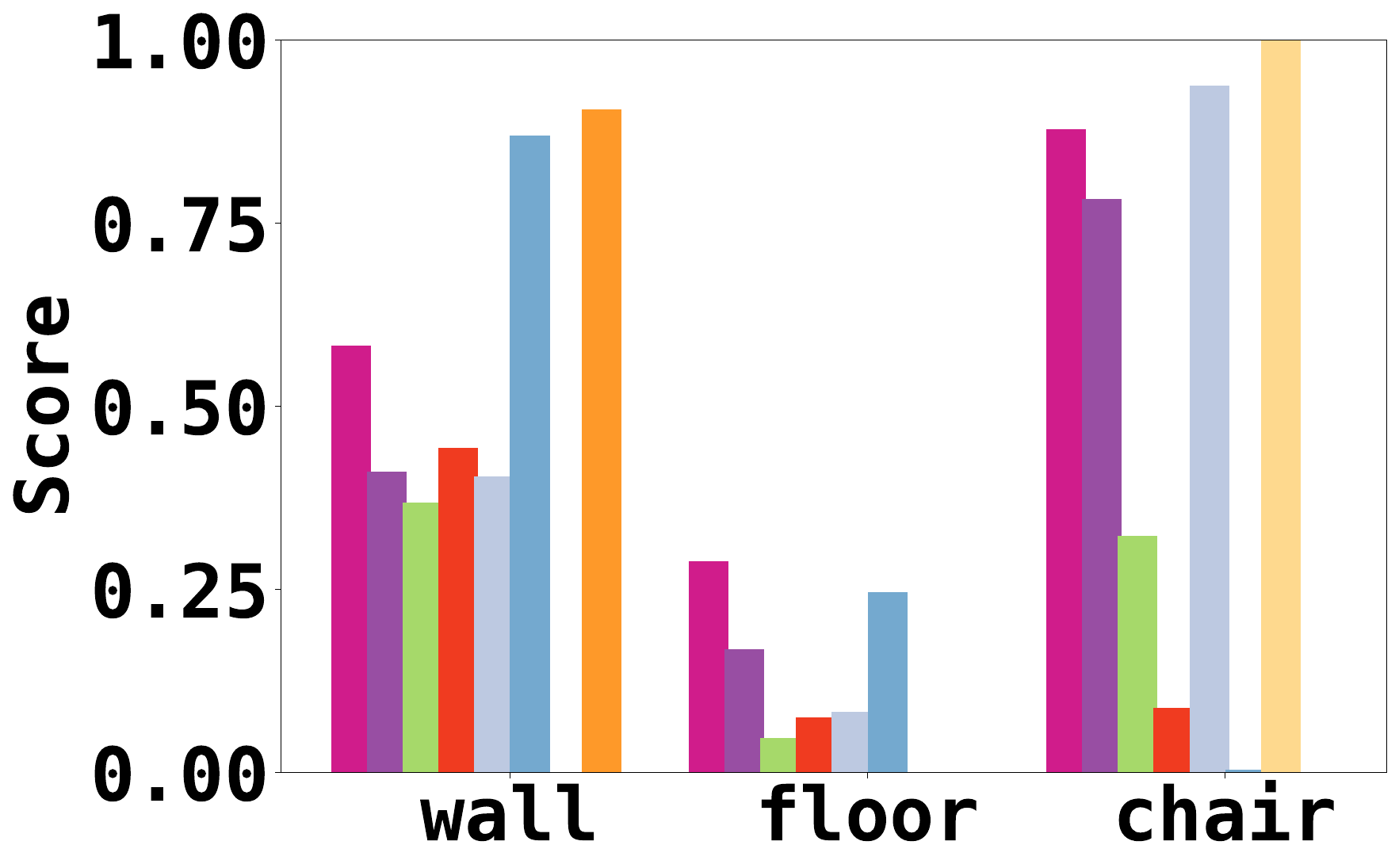} \\
    \end{subfigure}
    \caption{High IOU Error, High ROM (chair)}
    \label{fig:high iou high rom}

\end{figure*}

\noindent \textbf{High ROM}: In this category, the pixel-wise accuracy errors are high, and a high ROM value explains that these errors occur when a single object in the ground truth is subdivided into several regions in the prediction. For example, in Figure~\ref{fig:high iou high rom}, a chair is falsely segmented into multiple far apart regions. This is concurrent with the majority of pixels in the middle of the chair being predicted as other classes, resulting in low pixel-wise accuracy.\\

\begin{figure*}[tbhp!]
    \centering
    \begin{subfigure}[b]{2\columnwidth}
    \begin{tabular}{>{\centering\arraybackslash}p{0.22\columnwidth}>{\centering\arraybackslash}p{0.22\columnwidth}>{\centering\arraybackslash}p{0.22\columnwidth}>{\centering\arraybackslash}p{0.22\columnwidth}}
        \toprule
        \textbf{RGB Image} & \textbf{Ground-Truth} & \textbf{Prediction}& \textbf{Error metrics}\\
        \midrule
        \end{tabular}   
    \end{subfigure}
        
    \begin{subfigure}[b]{2\columnwidth}
        \includegraphics[width=0.24\columnwidth]{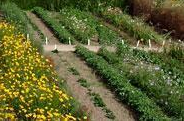}  
        \includegraphics[width=0.24\columnwidth]{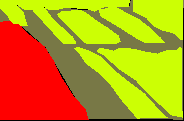} 
        \includegraphics[width=0.24\columnwidth]{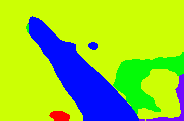} 
        \includegraphics[width=0.24\columnwidth]{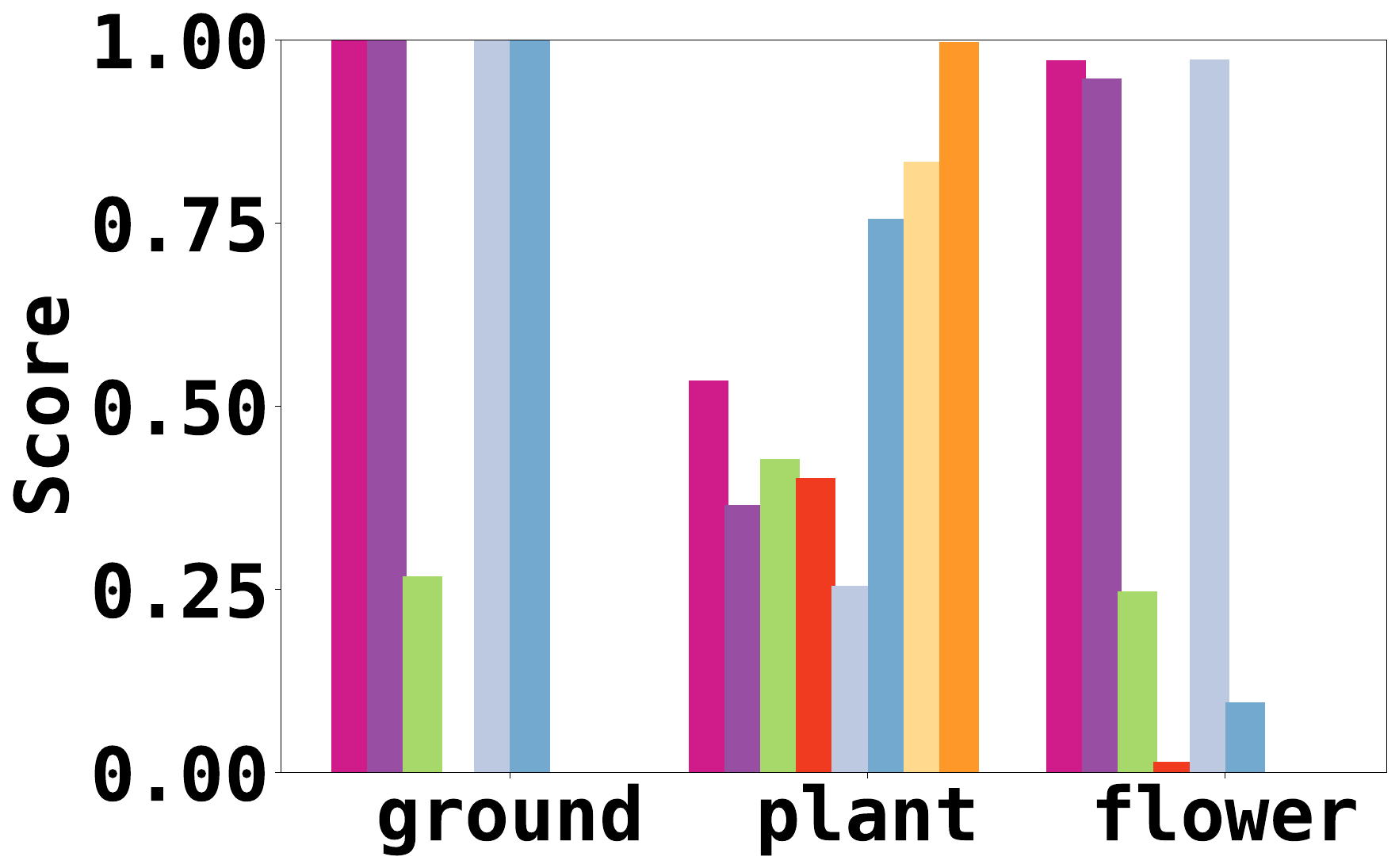} \\
    \end{subfigure}
    \caption{High IOU Error, High ROM/RUM (plant)}
    \label{fig:high iou high rom high rum}
\end{figure*}

\noindent \textbf{High ROM, High RUM}: Rarely, there are cases where both ROM and RUM are high, e.g. the plant class in Figure~\ref{fig:high iou high rom high rum}. This happens when one or more regions in the ground truth are over-segmented, while there are other regions are under-segmented. \\

\section{Discussion}
\label{sec:extend_discussion}
From the examples shown in Section~\ref{sec:examples}, we demonstrate that ROM and RUM have the following attributes and advantages.

\begin{enumerate}
    \item ROM/RUM accounts for model disagreement with ground-truth in over- and under-segmentation cases. Revealing over-/under-segmentation issues that are not reflected in other metrics. For example, as shown in Figure~\ref{fig:low iou high rum}, the prediction may have high pixel-wise accuracy, while suffering from severe under-segmentation issues. In downstream applications where under-segmentation is the primary concern, RUM needs to be considered together with other pixel-wise metrics for model selection.
    
    \item ROM/RUM can differentiate among different degrees of over-segmentation and under-segmentation issues. This is important because it allows researchers to quantify the severity of the over-/under-segmentation issues in a particular model as it pertains to a specific class or classes of objects. For example, in Figure~\ref{fig:low iou high rom}, both the table class and the chair class are over-segmented, but the chair class receives a higher ROM because it has more over-segmented regions per ground truth region. 
    
    \item ROM/RUM offers additional information for evaluating and selecting a segmentation model alongside the pixel-wise metrics.
    
    When pixel-wise accuracy is high, ROM/RUM can assist in validating the model prediction in region-wise performance. All examples in Figure~\ref{fig:low iou low rom low rum},~\ref{fig:low iou high rum},~\ref{fig:low iou high rom} have relatively low pixel-wise errors for certain object classes, but the pixel-wise metrics alone do not lend for a proper nor complete interpretation of the model's region-wise performance.
    As illustrated in Figure~\ref{fig:low iou low rom low rum}, a prediction should receive 0 values for pixel-wise error, ROM, and RUM if and only if it predicts perfect region-wise segmentation. A non-zero value of ROM/RUM indicates there are over-/under-segmentation issues within the prediction even is the pixel-wise errors are low. 
    
    \item When pixel-wise accuracy is low, ROM/RUM can explain the source of performance degradation and provide useful information for model selection. As illustrated in Figure~\ref{fig:high iou high rum} and Figure~\ref{fig:high iou high rom}, the pixel-wise accuracy of predictions can be penalized to a similar degree while creating contrasting region-wise segmentation qualities-- one may have major under-segmentation issues, while the other creates over-segmentation issues. 
    
\end{enumerate}

Overall, we demonstrated certain common model interrogation scenarios in which combining ROM/RUM with other pixel-wise measures allows researchers to select the most appropriate model for specific datasets or applications. Furthermore, interrogating models while using metrics conjointly may give rise to a more nuanced understanding of model performance for certain classes alone, or for certain models in multi-class correlated scenarios. 
In future work, we intend to explore how ROM/RUM can be used during the learning process of a model, in order to tailor a model for a specific use case. We intend to further study the interrelation between ROM and RUM, and combined ROM/RUM with pixel-wise metrics applied in multi-class correlated settings. A principled approach in this direction can allow for comprehensive, simultaneous evaluation and interpretation of model performance in semantic segmentation.

\clearpage

\bibliographystyle{IEEEtran}